\documentclass[11pt]{article}

\usepackage{acl}

\usepackage{mathptmx}
\usepackage{latexsym}

\usepackage{geometry}
\usepackage{changepage}
\usepackage[utf8]{inputenc} 
\usepackage[T1]{fontenc}    
\usepackage{graphicx}
\usepackage{amsmath}
\usepackage{amsfonts}       
\usepackage{hyperref}       
\usepackage{cleveref}
\usepackage{url}            
\usepackage{booktabs}       
\usepackage{nicefrac}       
\usepackage{microtype}      
\usepackage{xcolor}         
\usepackage{tikz}
\usetikzlibrary{calc}
\usepackage{float}
\usepackage{afterpage}
\usepackage{pdflscape}
\usepackage{subcaption} 
\newcommand{\model}{GPT-NeoX-20B}

\usepackage{todonotes}
\usepackage{tokenization}
\usepackage{float}

\DeclareMathOperator{\softmax}{softmax}

\DeclareMathOperator{\Attn}{Attn}
\DeclareMathOperator{\LN}{LN}
\DeclareMathOperator{\FF}{FF}

\title{\model: An Open-Source Autoregressive Language Model}

\author{Sid Black\thanks{Lead authors. Authors after the first three are listed in alphabetical order. See \Cref{app:contrib} for individual contribution details. Correspondence can be sent to \texttt{\{sid, stella, contact\}@eleuther.ai}} \And Stella Biderman\protect\footnotemark[1] \And Eric Hallahan\protect\footnotemark[1] \AND Quentin Anthony \And Leo Gao \And Laurence Golding \And Horace He \AND Connor Leahy \And Kyle McDonell \And Jason Phang \And Michael Pieler \AND USVSN Sai Prashanth \And Shivanshu Purohit \And Laria Reynolds \AND Jonathan Tow \And Ben Wang \And Samuel Weinbach}
\date{\today}
\begin{document}

\maketitle

\begin{abstract}

    We introduce \model{}, a 20 billion parameter autoregressive language model trained on the Pile, whose weights will be made freely and openly available to the public through a permissive license. It is, to the best of our knowledge, the largest dense autoregressive model that has publicly available weights at the time of submission. In this work, we describe \model{}'s architecture and training and evaluate its performance on a range of language-understanding, mathematics, and knowledge-based tasks. We find that GPT-NeoX-20B is a particularly powerful few-shot reasoner and gains far more in performance when evaluated five-shot than similarly sized GPT-3 and FairSeq models. We open-source the training and evaluation code, as well as the model weights, at \url{https://github.com/EleutherAI/gpt-neox}.
  
\end{abstract}

\section{Introduction}
\label{sec:intro}

Over the past several years, there has been an explosion in research surrounding large language models (LLMs) for natural language processing, catalyzed largely by the impressive performance of Transformer-based language models such as BERT \citep{devlin2019bert}, GPT-2 \citep{radford2019language}, GPT-3 \citep{brown2020language}, and T5 \citep{raffel2020exploring}. One of the most impactful outcomes of this research has been the discovery that the performance of LLMs scales predictably as a power law with the number of parameters, with architectural details such as width/depth ratio having a minimal impact on performance within a wide range \citep{kaplan2020scaling}. A consequence of this has been an abundance of research focusing on scaling Transformer models up to ever-larger scales, resulting in dense models that surpass 500B parameters \citep{megatron-530b,chowdhery2022palm}, a milestone that would have been almost unthinkable just a few years prior.

Today, there are dozens of publicly acknowledged LLMs in existence, the largest having more than two orders of magnitude more parameters than GPT-2, and even at that scale there are nearly a dozen different models. However, these models are almost universally the protected intellectual property of large organizations, and are gated behind a commercial API, available only upon request, or not available for outsider use at all. To our knowledge, the only freely and publicly available dense autoregressive language models larger than GPT-2 are GPT-Neo (2.7B parameters) \citep{black2021gpt}, GPT-J-6B \citep{gpt-j}, Megatron-11B\footnote{\href{https://github.com/pytorch/fairseq/tree/main/examples/megatron\_11b}{This model} does not work using the provided codebase, and we have been told it under-performs GPT-J.}, Pangu-$\alpha$-13B \citep{zeng2021pangualpha}, and the recently released FairSeq models (2.7B, 6.7B, and 13B parameters) \citep{fairseq-13B}.

In this paper we introduce \model{}, a 20 billion parameter open-source autoregressive language model. We make the models weights freely and openly available to the public through a permissive license, motivated by the belief that open access to LLMs is critical to advancing research in a wide range of areas---particularly in AI safety, mechanistic interpretability, and the study of how LLM capabilities scale. Many of the most interesting capabilities of LLMs only emerge above a certain number of parameters, and they have many properties that simply cannot be studied in smaller models. Although safety is often cited as a justification for keeping model weights private, we believe this is insufficient to prevent misuse, and is largely a limitation on the ability to probe and study LLMs for researchers not based at the small number of organizations that have access to state of the art language models. In addition, we make partially trained checkpoints avaliable at evenly spaced 1000 step intervals throughout the whole of training. We hope that by making a wide range of checkpoints throughout training freely available, we will facilitate research on the training dynamics of LLMs, as well as the aforementioned areas of AI safety and interpretability.

In studying \model{}, we find several noteworthy phenomena at odds with the established literature. We train on a dataset that contains duplicated data for more than one epoch but see no evidence of performance loss. While \citep{hendrycks2020measuring} claims that few-shot prompting doesn't improve performance on their task, we find that this is actually a phenomenon unique to GPT-3 and doesn't apply to either \model{} or FairSeq models. Finally, we find that \model{} is a powerful few-shot learner, recieving a much larger performance boost from few-shot examples than comparable sized GPT-3 and FairSeq models. As we see the same with GPT-J-6B \citep{gpt-j}, we hypothesize that this may be due to the shared choice of training data.

In the following sections, we give a broad overview of \model{}'s architecture and training hyperparameters, detail the hardware and software setup used for training and evaluation, and elaborate on the choices made when designing the training dataset and tokenization. We also address of some of the difficulties and unknowns we encountered in training such a large model. We place significant importance on the broader impacts of the release \model{}, and provide a lengthy discussion of why we believe its release is a net benefit. We also document issues of training cost and carbon emissions in as much detail as much as possible.

\section{Model Design and Implementation}
\label{sec:model-details}

\model{} is an autoregressive transformer decoder model whose architecture largely follows that of GPT-3 \citep{brown2020language}, with a few notable deviations described below. Our model has 20 billion parameters, of which 19.9 billion are ``non-embedding'' parameters that \citet{kaplan2020scaling} identify as the proper number to use for scaling laws analysis. Our model has 44 layers, a hidden dimension size of 6144, and 64 heads.

\subsection{Model Architecture}
\label{subsec:model-arch}

Although our architecture is largely similar to GPT-3, there are some notable differences. In this section we give a high-level overview of those differences, but ask the reader to refer to \cite{brown2020language} for full details of the model architecture. Our model architecture is almost identical to that of GPT-J \citep{gpt-j}\footnote{The sole difference is due to an oversight discussed in \Cref{subsubsec:parallel-ff}}, however we choose to use GPT-3 as the point of reference because there is no canonical published reference on the design of GPT-J.

\subsubsection{Rotary Positional Embeddings}
\label{subsubsec:rope} 
We use rotary embeddings \citep{su2021roformer} instead of the learned positional embeddings used in GPT models \citep{radford2018improving}, based on our positive prior experiences using it in training LLMs. 
Rotary embeddings are a form of static relative positional embeddings. In brief, they twist the embedding space such that the attention of a token at position $m$ to token at position $n$ is linearly dependent on $m-n$. More formally, they modify the standard multiheaded attention equations from \[\softmax\left(\frac{1}{\sqrt{d}}\sum_{n,m}\mathbf{x}_m^T\mathbf{W}^T_q\mathbf{W}_k\mathbf{x}_n\right),\]
where $\mathbf{x}_m$, $\mathbf{x}_n$ are (batched) embeddings of tokens at position $m$ and $n$ respectively and $\mathbf{W}^T_q$, $\mathbf{W}_k$ are the query and key weights respectively to
\[\softmax\left(\frac{1}{\sqrt{d}}\sum_{n,m}\mathbf{x}_m^T\mathbf{W}^T_qR^{d}_{\Theta,(n-m)}\mathbf{W}_k\mathbf{x}_n\right),\]
where $R^d_{\Theta,x}$ is a $d\times d$ block diagonal matrix with the block of index $i$ being a $2$D rotation by $x\theta_i$ for hyperparameters $\Theta = \{\theta_i = 10000^{-2i/d}\;\vert\;i\in\{0,1,2,\ldots, (d-1)/2\}\}$. 

\begin{figure}[th]
    \centering
    \includegraphics[width=\linewidth]{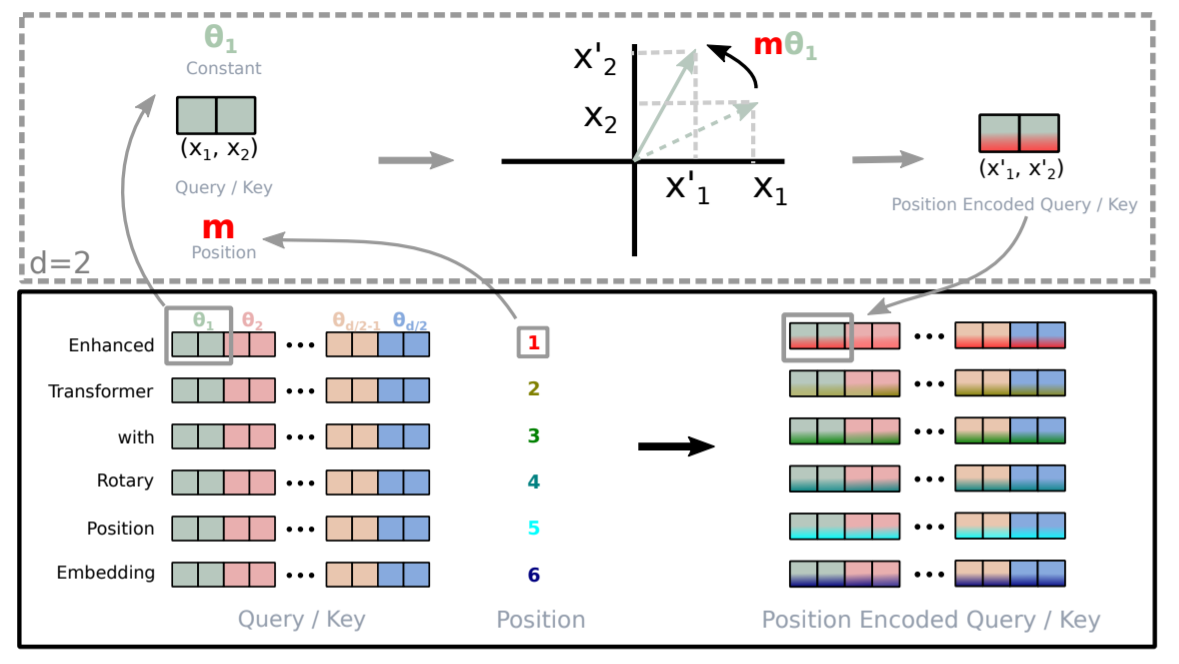}
    \caption{A pictorial representation of rotary embeddings, from \citet{su2021roformer}.}
    \label{fig:rotary}
\end{figure}

While \citet{su2021roformer} apply rotary embeddings to every embedding vector, we follow \citet{gpt-j} and instead apply it only to the first $25\%$ of embedding vector dimensions. Our initial experiments indicate that this strikes the best balance of performance and computational efficiency.\footnote{See the Weights \& Biases reports \href{https://wandb.ai/eleutherai/neox/reports/Partial-Rotary-Tests-v2--Vmlldzo2MjE4MTQ}{here} and \href{https://wandb.ai/lucidrains/x-transformers-experiments/reports/partial-rotary--Vmlldzo2MjAyODY?accessToken=f657029yibaln2vseli6egxxwykhpeuedeuvcnmmdgn4i6d5b1r30it3mp0gw0k5}{here} for further details.}

\subsubsection{Parallel Attention + FF Layers}
\label{subsubsec:parallel-ff}
We compute the Attention and Feed-Forward (FF) layers in parallel\footnote{See \href{https://github.com/EleutherAI/gpt-neox/blob/ac3d8087f1762213880523893a52329d66d2d1a9/megatron/model/transformer.py\#L593}{GitHub} for implementation details.} and sum the results, rather than running them in series. This is primarily for efficiency purposes, as each residual addition with op-sharding requires one all-reduce in the forward pass and one in the backwards pass \citep{shoeybi2020megatronlm}. By computing the Attention and FFs in parallel, the results can be reduced locally before performing a single all-reduce. In Mesh Transformer JAX \citep{mesh-transformer-jax}, this led to a 15\% throughput increase, while having comparable loss curves with running them in series during early training.

Due to an oversight in our code, we unintentionally apply two independent Layer Norms instead of using a tied layer norm the way \citet{gpt-j} does. Instead of computing \[x + \Attn(\LN_1(x)) + \FF(\LN_1(x))\] as intended, our codebase unties the layer norms:
\[x + \Attn(\LN_1(x)) + \FF(\LN_2(x)).\] Unfortunately, this was only noticed after we were much too far into training to restart. Subsequent experiments at small scales indicated that the untied layer norm makes no difference in performance, but we nevertheless wish to highlight this in the interest of transparency.

\subsubsection{Initialization}
\label{subsubsec:init}

For the Feed-Forward output layers before the residuals, we used the initialization scheme introduced in \citet{mesh-transformer-jax}, $\frac{2}{L\sqrt{d}}$. This prevents activations from growing with increasing depth and width, with the factor of 2 compensating for the fact that the parallel and feed-forward layers are organized in parallel. For all other layers, we use the \textit{small init} scheme from \citet{transformers-without-tears}, $\sqrt{\frac{2}{d+4d}}$

\subsubsection{All Dense Layers}
\label{subsubsec:dense-layers}
While GPT-3 uses alternating dense and sparse layers using the technique introduced in \citet{child2019generating}, we instead opt to exclusively use dense layers to reduce implementation complexity.

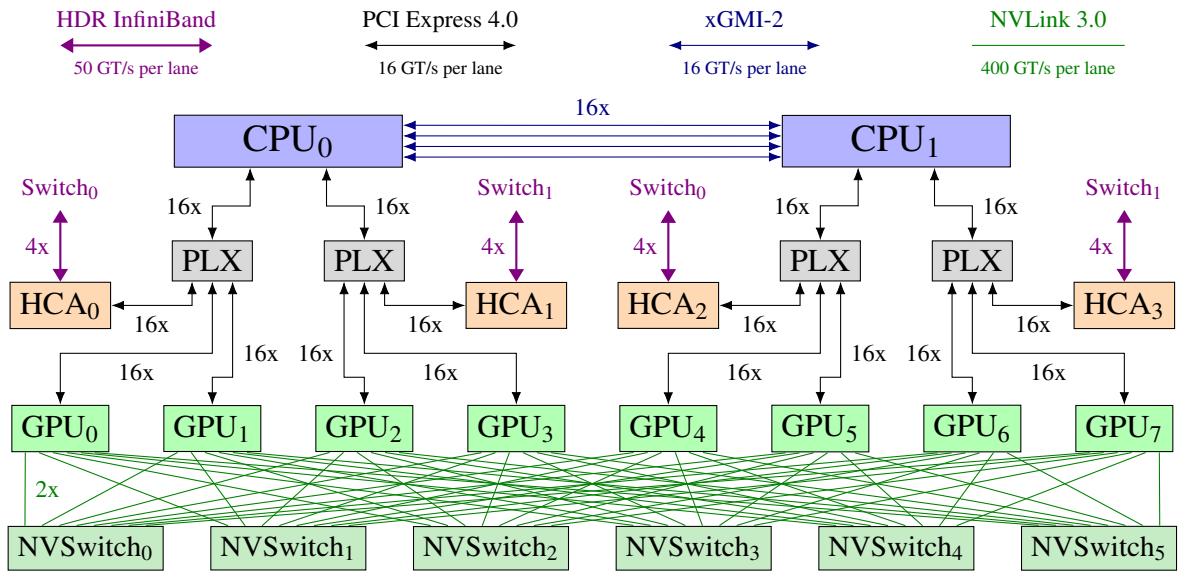
\begin{figure*}[hbt]
    \centering
    \usetikzlibrary{graphs,arrows.meta,calc}
\begin{tikzpicture}[new set=cpus,new set=plxs,new set=hcas,new set=gpus,new set=nvls]
	\begin{scope}[nodes=draw,align=center,shape=rectangle]
		\draw (0,0) foreach \cpui/\poscpu in {0/left,1/right} {
			+(\poscpu:0.25*\textwidth) node[set=cpus,minimum width=03/16*\textwidth,fill=blue!30](cpu\cpui) {\Large CPU{$_\cpui$}} 
			child[edge from parent/.style={draw=none},level distance=0.625in,sibling distance=0.125*\textwidth] foreach \plxi in {0,1} {
				node[set=plxs,fill=gray!30](plx\cpui\plxi) {\large PLX}
			}
		};
		\draw foreach \cpui/\plxi/\pos/\hcai in {0/0/left/0,0/1/right/1,1/0/left/2,1/1/right/3} {
			({plx\cpui\plxi}.south) ++(0,-0.125in) ++(\pos:0.125*\textwidth) node[set=hcas,fill=orange!30](hca\cpui\plxi) {\large HCA{$_\hcai$}}
		};
		\draw (-0.4375*\textwidth,-1.5in) foreach \gpui in {0,...,7} {
			+(\gpui*0.125*\textwidth,0)node[set=gpus,fill=green!30](gpu\gpui) {\large GPU{$_\gpui$}}
		};
		\foreach \gpui/\alias in {0/gpu000,1/gpu001,2/gpu011,3/gpu010,4/gpu100,5/gpu101,6/gpu111,7/gpu110}
			\pgfnodealias{\alias}{gpu\gpui};
		\draw (-0.416666667*\textwidth,-2.125in) foreach \nvli in {0,...,5} {
			+(\nvli*0.166666667*\textwidth,0) node[set=nvls,fill=green!50!gray!30](nvl\nvli) {NVSwitch{$_\nvli$}}
		};
	\end{scope}
	\begin{scope}[infiniband/.style={color=violet,arrows={{Triangle[]}-{Triangle[]}},thick,auto},pciexpress/.style={arrows={{Latex[]}-{Latex[]}},auto},xgmi2/.style={color=blue!50!black,arrows={{Latex[]}-{Latex[]}},auto},nvlink/.style={color=green!50!black,line cap=round}]
		\foreach \gpui[count=\gpuj] in {0,...,7}
			\foreach \nvli[count=\nvlj] in {0,...,5}
				\draw[nvlink] ($ ({gpu\gpui}.south west)!\nvlj/7!({gpu\gpui}.south east) + (0,-0.2pt)$) to node(nvlink\gpui\nvli) {} ($ ({nvl\nvli}.north west)!\gpuj/9!({nvl\nvli}.north east) + (0,+0.2pt)$);
		\draw[nvlink,right] (nvlink00) node[right](nvlinktext) {\footnotesize 2x};
		\begin{scope}[pciexpress]
			\foreach \cpui in {0,1}
				\foreach \plxi/\pos/\posi in {0/west/east,1/east/west}
					\draw ($ ({cpu\cpui}.south)!1/3!({cpu\cpui}.south \pos)$) --  +(0,-0.2in) -| node[anchor=\posi]{\footnotesize 16x} ({plx\cpui\plxi}.north);
			\foreach \cpui/\plxi/\pos in {0/0/east,0/1/west,1/0/east,1/1/west}
				\draw ($ ({plx\cpui\plxi}.south)!-1/2!({plx\cpui\plxi}.south \pos)$) |- node[ near end, below]{\footnotesize 16x} ({hca\cpui\plxi}.{\pos});
			\foreach \cpui/\plxi/\pos in {0/0/left,0/1/right,1/0/left,1/1/right}
				\draw[infiniband] ({hca\cpui\plxi}.north) -- node(ib\cpui\plxi){\footnotesize 4x} ++(0,0.375in) node[anchor=south](switch\cpui\plxi){\footnotesize Switch{$_\plxi$}};
			\foreach \cpui in {0,1}
				\foreach \plxi/\pos in {0/east,1/west}
					\draw ({plx\cpui\plxi}.south) -- ++(0,-0.375in) -| node[ near start, below]{\footnotesize 16x} ({gpu\cpui\plxi0}.north);
			\foreach \cpui in {0,1}
				\foreach \plxi/\pos/\posi in {0/east/west,1/west/east}
					\draw ($ ({plx\cpui\plxi}.south)!1/2!({plx\cpui\plxi}.south \pos)$) -- node[near end, anchor=\posi]{\footnotesize 16x} ++(0,-0.5in) -| ({gpu\cpui\plxi1}.north) ;
			\draw[xgmi2] ($ (cpu0.north east)!1/5!(cpu0.south east) $) -- node(xgmi) {\footnotesize 16x} ($ (cpu1.north west)!1/5!(cpu1.south west)$);%
			\foreach \xgmii in {2,...,4} 
			    \draw[xgmi2] ($ (cpu0.north east)!\xgmii/5!(cpu0.south east) $) -- node(xgmi\xgmii){} ($ (cpu1.north west)!\xgmii/5!(cpu1.south west) $);
		\end{scope}
		\draw[infiniband] (-7/16*\textwidth,0.5in) -- +(1/16*\textwidth,0) node[yshift=0.125in](ib) {\small\strut HDR InfiniBand} node[yshift=-0.125in](ibspeed) {\scriptsize\strut 50 GT/s per lane} -- +(1/8*\textwidth,0);
		\draw[pciexpress] (-3/16*\textwidth,0.5in) -- +(1/16*\textwidth,0) node[centered,yshift=0.125in](pcie) {\small\strut PCI Express 4.0} node[yshift=-0.125in](pciespeed) {\scriptsize\strut 16 GT/s per lane} -- +(1/8*\textwidth,0);
		\draw[xgmi2] (1/16*\textwidth,0.5in) -- +(1/16*\textwidth,0) node[yshift=0.125in](xgmi) {\small\strut xGMI-2} node[yshift=-0.125in](xgmispeed) {\scriptsize\strut 16 GT/s per lane}
		-- +(1/8*\textwidth,0);%
		\draw[nvlink] (5/16*\textwidth,0.5in) -- +(1/16*\textwidth,0) node[yshift=0.125in](nvlink) {\small\strut NVLink 3.0} node[yshift=-0.125in](nvlinkspeed) {\scriptsize\strut 400 GT/s per lane}
		-- +(1/8*\textwidth,0);
	\end{scope}
\end{tikzpicture}
    \caption{Architecture diagram of a single training node.}
    \label{fig:training-node}
\end{figure*}

\subsection{Software Libraries}
\label{subsec:software}

Our model is trained using a codebase that builds on Megatron \citep{shoeybi2020megatronlm} and DeepSpeed \citep{rasley2020deepspeed} to facilitate efficient and straightforward training of large language models with tens of billions of parameters. We use the official PyTorch v1.10.0 release binary package compiled with CUDA 11.1. This package is bundled with NCCL 2.10.3 for distributed communications.

\subsection{Hardware}
\label{subsec:hardware}

We trained \model{} on twelve Supermicro AS-4124GO-NART servers, each with eight NVIDIA A100-SXM4-40GB GPUs and configured with two AMD EPYC 7532 CPUs. All GPUs can directly access the InfiniBand switched fabric through one of four ConnectX-6 HCAs for GPUDirect RDMA. Two NVIDIA MQM8700-HS2R switches---connected by 16 links---compose the spine of this InfiniBand network, with one link per node CPU socket connected to each switch. \Cref{fig:training-node} shows a simplified overview of a node as configured for training.

\section{Training}

Due to the intractability of performing a hyperparameter sweep for a 20 billion parameter model, we opted to use the values from \citet{brown2020language} to guide our choice of hyperparameters. As \citet{brown2020language} did not train a model at our exact scale, we interpolate between the learning rates of their 13B and 175B models to arrive at a learning rate of $0.97\mathrm{\textsc{e}}{-5}$. Based on the results of smaller scale experiments, we select a weight decay of 0.01. To achieve a higher training throughput, we opt to use the same batch size as OpenAI's 175B model--approximately 3.15M tokens, or 1538 contexts of 2048 tokens each, and train for a total of $150,000$ steps, decaying the learning rate with a cosine schedule to $10\%$ of its original value at the end of training.

We use the AdamW \citep{loshchilov2019decoupled} optimizer, with beta values of $0.9$ and $0.95$ respectively, and an epsilon of $1.0\mathrm{\textsc{e}}{-8}$. We extend AdamW with the \textit{ZeRO} optimizer \citep{rajbhandari2020zero} to reduce memory consumption by distributing optimizer states across ranks. Since the weights and optimizer states of a model at this scale do not fit on a single GPU, we use the tensor parallelism scheme introduced in \citet{shoeybi2020megatronlm} in combination with pipeline parallelism \citep{harlap2018pipedream} to distribute the model across GPUs. To train \model{}, we found that the most efficient way to distribute the model given our hardware setup was to set a tensor parallel size of 2, and a pipeline parallel size of 4. This allows for the most communication intensive processes, tensor and pipeline parallelism, to occur within a node, and data parallel communication to occur across node boundaries. In this fashion, we were able to achieve and maintain an efficiency of 117 teraFLOPS per GPU.

\subsection{Training Data}
\label{sec:training-data}

\model{} was trained on the Pile \citep{gao2020pile}, a massive curated dataset designed specifically for training large language models. It consists of data from 22 data sources, coarsely broken down into 5 categories:

\begin{itemize}
    \item \textbf{Academic Writing}: Pubmed Abstracts and PubMed Central, arXiv, FreeLaw,\footnote{\url{https://www.courtlistener.com/}} USPTO Backgrounds,\footnote{\url{https://bulkdata.uspto.gov/}} PhilPapers,\footnote{\url{https://philpapers.org/}} NIH Exporter\footnote{\url{https://exporter.nih.gov/}}
    \item \textbf{Web-scrapes and Internet Resources}: CommonCrawl, OpenWebText2, StackExchange,\footnote{\url{https://archive.org/details/stackexchange}} Wikipedia (English)
    \item \textbf{Prose}: BookCorpus2, Bibliotik, Project Gutenberg \citep[PG-19;][]{rae2019compressive}
    \item \textbf{Dialogue}: Youtube subtitles, Ubuntu IRC,\footnote{\url{https://irclogs.ubuntu.com/}} OpenSubtitles \citep{lison-tiedemann-2016-opensubtitles2016}, Hacker News,\footnote{\url{https://news.ycombinator.com/}} EuroParl \citep{koehn-2005-europarl}
    \item \textbf{Miscellaneous}: GitHub, the DeepMind Mathematics dataset \citep{saxton2019analysing}, Enron Emails \citep{10.1007/978-3-540-30115-8_22}
\end{itemize}

In aggregate, the Pile consists of over 825~GiB of raw text data. The diversity of data sources reflects our desire for a general-purpose language model. Certain components are up-sampled to obtain a more balanced data distribution. In contrast, GPT-3's training data consists of web-scrapes, books datasets, and Wikipedia. When comparing results in this work to GPT-3, the training data is almost certainly the biggest known unknown factor. Full details of the Pile can be found in the technical report \citep{gao2020pile} and the associated datasheet \citep{biderman2022datasheet}.

It is particularly notable that the Pile contains a scrape of StackExchange preprocessed into a Q/A form. There is a significant and growing body of work on the influence of the syntactic structure of finetuning data on downstream performance \citep{zhong2021adapting,tan2021msp,sanh2021multitask,wei2021finetuned}. While so far there has been no systematic work that focuses on \textit{prompted pretraining}, recent work \citep{biderman2022neural} observed that the formulation of the StackExchange component of the Pile appears to heavily influence code generation. 

\subsection{Tokenization}
\label{sec:tokenization}

\begin{figure}[t]
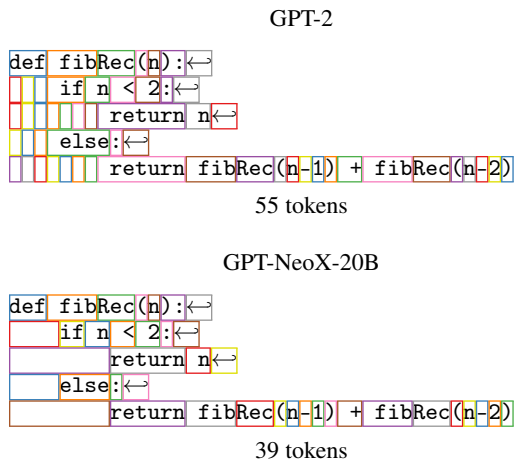

	\footnotesize
	\centering
	\begin{minipage}{\linewidth}
		\subcaption*{GPT-2}
		\begin{tokenized}
		\parbox{\textwidth}{
			\tokenbox{def}\tokenbox{~fib}\tokenbox{Rec}\tokenbox{(}\tokenbox{n}\tokenbox{):}\tokenbox{$\hookleftarrow$}\newline
			\tokenbox{~}\tokenbox{~}\tokenbox{~}\tokenbox{~if}\tokenbox{~n}\tokenbox{~<}\tokenbox{~2}\tokenbox{:}\tokenbox{$\hookleftarrow$}\newline
			\tokenbox{~}\tokenbox{~}\tokenbox{~}\tokenbox{~}\tokenbox{~}\tokenbox{~}\tokenbox{~}\tokenbox{~return}\tokenbox{~n}\tokenbox{$\hookleftarrow$}\newline
			\tokenbox{~}\tokenbox{~}\tokenbox{~}\tokenbox{~else}\tokenbox{:}\tokenbox{$\hookleftarrow$}\newline
			\tokenbox{~}\tokenbox{~}\tokenbox{~}\tokenbox{~}\tokenbox{~}\tokenbox{~}\tokenbox{~}\tokenbox{~return}\tokenbox{~fib}\tokenbox{Rec}\tokenbox{(}\tokenbox{n}\tokenbox{-}\tokenbox{1}\tokenbox{)}\tokenbox{~+}\tokenbox{~fib}\tokenbox{Rec}\tokenbox{(}\tokenbox{n}\tokenbox{-}\tokenbox{2}\tokenbox{)}
	}\end{tokenized}
	\vskip0.5em\centering{55 tokens}
	\end{minipage}
	\vskip1.5em
	\begin{minipage}{\linewidth}
		\subcaption*{\model{}}
		\begin{tokenized}
		\parbox{\textwidth}{
			\tokenbox{def}\tokenbox{~fib}\tokenbox{Rec}\tokenbox{(}\tokenbox{n}\tokenbox{):}\tokenbox{$\hookleftarrow$}\newline
			\tokenbox{~~~~}\tokenbox{if}\tokenbox{~n}\tokenbox{~<}\tokenbox{~2}\tokenbox{:}\tokenbox{$\hookleftarrow$}\newline
			\tokenbox{~~~~~~~~}\tokenbox{return}\tokenbox{~n}\tokenbox{$\hookleftarrow$}\newline
			\tokenbox{~~~~}\tokenbox{else}\tokenbox{:}\tokenbox{$\hookleftarrow$}\newline
			\tokenbox{~~~~~~~~}\tokenbox{return}\tokenbox{~fib}\tokenbox{Rec}\tokenbox{(}\tokenbox{n}\tokenbox{-}\tokenbox{1}\tokenbox{)}\tokenbox{~+}\tokenbox{~fib}\tokenbox{Rec}\tokenbox{(}\tokenbox{n}\tokenbox{-}\tokenbox{2}\tokenbox{)}
		}\end{tokenized}
	\vskip0.5em\centering{39 tokens}
	\end{minipage}
	\caption{GPT-2 tokenization vs. \model{} tokenization. \model{} tokenization handles whitespace better, which is particularly useful for text such as source code. For more examples, see \Cref{app:tokenization}.}
    \label{fig:tok_example}
\end{figure}

For \model{}, we use a BPE-based tokenizer similar to that used in GPT-2, with the same total vocabulary size of 50257, with three major changes to the tokenizer. First, we train a new BPE tokenizer based on the Pile, taking advantage of its diverse text sources to construct a more general-purpose tokenizer. Second, in contrast to the GPT-2 tokenizer which treats tokenization at the start of a string as a non-space-delimited token, the \model{} tokenizer applies consistent space delimitation regardless. This resolves an inconsistency regarding the presence of prefix spaces to a tokenization input.\footnote{\url{https://discuss.huggingface.co/t/bpe-tokenizers-and-spaces-before-words/475/2}}. An example can be seen in \Cref{fig:tok_example}.
Third, our tokenizer contains tokens for repeated space tokens (all positive  integer amounts of repeated spaces up to and including 24). This allows the \model{} tokenizer to tokenize text with large amounts of whitespace using fewer tokens; for instance, program source code or arXiv \LaTeX{} source files. See \Cref{sec:tokenizer-analysis} for an analysis of the tokenizer.

\subsection{Data Duplication}

In the past two years, the standard practice when training autoregressive language models has become to train for only one epoch \citep{komatsuzaki2019one,kaplan2020scaling,henighan2020scaling}. Recent research has claimed to see significant benefits from going even further and deduplicating training data \citep{lee2021deduplicating,kandpal2022deduplicating,roberts2022scaling}. In particular, every publicly known larger language model other than GPT-3 \citep{brown2020language} and Jurassic-1\footnote{In private communication, the authors confirmed that Jurassic-1 was trained on the Pile \citep{gao2020pile}.} either uses some form of deduplication \citep{rae2021gopher,askell2021general,zeng2021pangualpha,sun2021ernie,megatron-530b,hoffmann2022chinchilla,chowdhery2022palm} or does not discuss the training data in sufficient detail to determine what was done \citep{kim2021changes}.

\begin{figure}
    \centering
    \includegraphics[width=\linewidth]{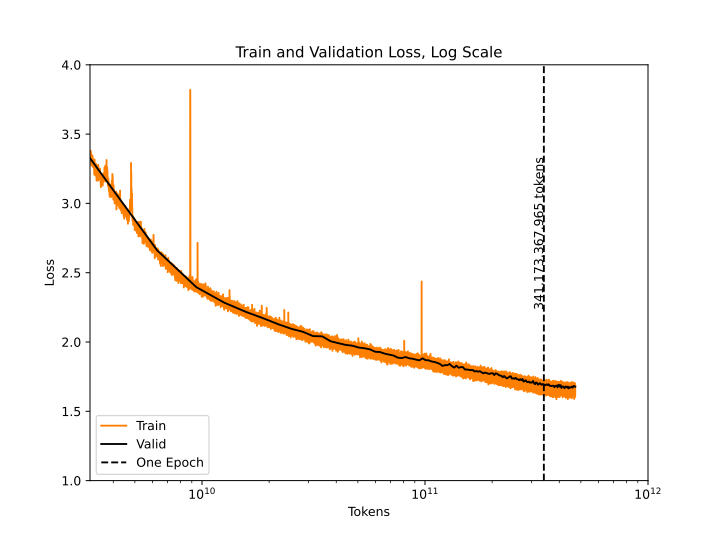}
    \includegraphics[width=\linewidth]{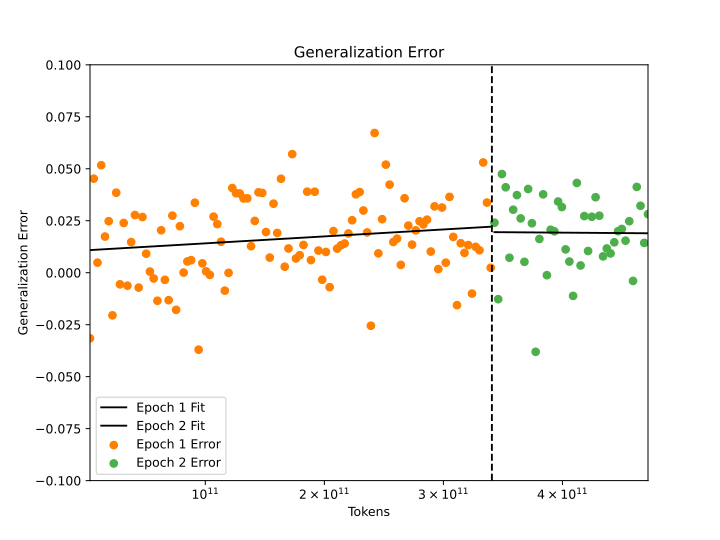}
    \caption{Training and validation loss for GPT-NeoX-20B. As the validation loss continued to fall into the beginning of the second epoch, we decided to let it train further.}
    \label{fig:loss}
\end{figure}

When the Pile was originally made, the only language model larger than \model{} that existed was GPT-3, which upsampled high-quality subsets of its training data. The Pile followed suit, and due to a combination of a lack of resources for large-scale ablations and a lack of noticeable impact at smaller scales, we opt to use the Pile as-is. As shown in \cref{fig:loss}, even at the 20B parameter scale we see no drop in test validation loss after crossing the one epoch boundary.

Unfortunately, none of the papers that have claimed to see an improvement from deduplication have released trained models that demonstrate this, making replication and confirmation of their results difficult. \citet{lee2021deduplicating} releases the deduplication code that they used, which we intend to use to explore this question in more detail in the future.

It is important to note that even if there is not an improvement in loss or on task evaluations there are nevertheless compelling reasons to deduplicate training data for any model put into production. In particular, systematic analysis has shown significant benefits in terms of reducing the leakage of training data \citep{lee2021deduplicating,zhang2021counterfactual,carlini2022quantifying,kandpal2022deduplicating}.

\section{Performance Evaluations}
\label{sec:evals}

To evaluate our model we use the EleutherAI Language Model Evaluation Harness \citep{eval-harness}, an open source codebase for language model evaluation that supports a number of model APIs. As our goal is to make a powerful model publicly accessible, we compare with English language models with at least 10B parameters that are publicly accessible. We compare with the GPT-3 models on the OpenAI API \citep{brown2020language}, the open source FairSeq dense models \citep{fairseq-13B}, and GPT-J-6B \citep{gpt-j}. We do not compare against T5 \citep{raffel2020exploring} or its derivatives as our evaluation methodology assumes that the models are autoregressive. While there is a Megatron-11B checkpoint that has been publicly released, the released code is \href{https://github.com/pytorch/fairseq/issues/2358}{non-functional} and we have not been able to get the model to work. We do not compare against any mixture-of-experts models as no public MoE model achieves performance comparable to a 10B parameter dense model.

While the size of the GPT-3 API models are not officially confirmed, we follow \citet{gao2021sizes} and assess them as being 350M (Ada), 1.3B (Babbage), 6.7B (Curie), and 175B (Da Vinci). We categorize both GPT-J-6B and \model{} under the umbrella of GPT-NeoX models, as both models are trained with the same architecture and were trained on the same dataset. 
However, we connect them using a dashed line to reflect the fact that these two models are not the same model trained at two different scales the way the FairSeq and GPT-3 models are, having been trained using different codebases, different tokenizers, and for different numbers of tokens.

Where we were able to obtain the relevant information, we report two baselines: human-level performance and random performance. All plots contain error bars representing two standard errors, indicating the $95\%$ confidence interval around each point. For some plots, the standard error is so small that the interval is not visible.

\subsection{Tasks Evaluated}
\label{subsec:lm}

We evaluate our model on a diverse collection of standard language model evaluation datasets that we divide into three main categories: natural language tasks, Advanced Knowledge-Based Tasks, and Mathematical Tasks. We evalutate GPT-J-6B, \model{}, and FairSeq models both zero- and five-shot, but due to financial constraints only evaluate GPT-3 models zero-shot. Due to space constraints a representative subset of the results are shown here, with the rest in \Cref{app:downstream}.

\paragraph{Natural Language Tasks} We evaluate our model on a diverse collection of standard language model evaluation datasets: ANLI \citep{nie2020adversarial}, ARC \citep{clark2018think}, HeadQA (English) \citep{vilares2019head}, HellaSwag \citep{zellers2019hellaswag}, LAMBDADA \citep{paperno2016lambada}, LogiQA \citep{liu2021logiqa}, OpenBookQA \citep{mihaylov2018can}, PiQA \citep{bisk2020piqa}, PROST \citep{aroca2021prost}, QA4MRE \citep{penas2013qa4mre} (2013), SciQ \citep{welbl2017crowdsourcing}, TriviaQA \citep{joshi2017triviaqa}, Winogrande \citep{sakaguchi2020winogrande}, and the SuperGlue version of the Winograd Schemas Challenge (WSC) \citep{wang2019superglue}.

\paragraph{Mathematical Tasks} The solving of mathematical problem solving is an area that has had a long history of study in AI research, despite the fact that large language models tend to perform quite poorly on both arithmetic tasks and mathematical problems phrased in natural language. We evaluate on the MATH test dataset \citep{hendrycks2021measuring} as well as on the numerical arithmetic problems introduced by \citet{brown2020language}. Note that the MATH test dataset is an evaluation metric that is generally finetuned on, but due to computational limitations we only evaluate models zero- and five-shot here.

\begin{figure*}
    \centering
    \includegraphics[width=0.45\textwidth]{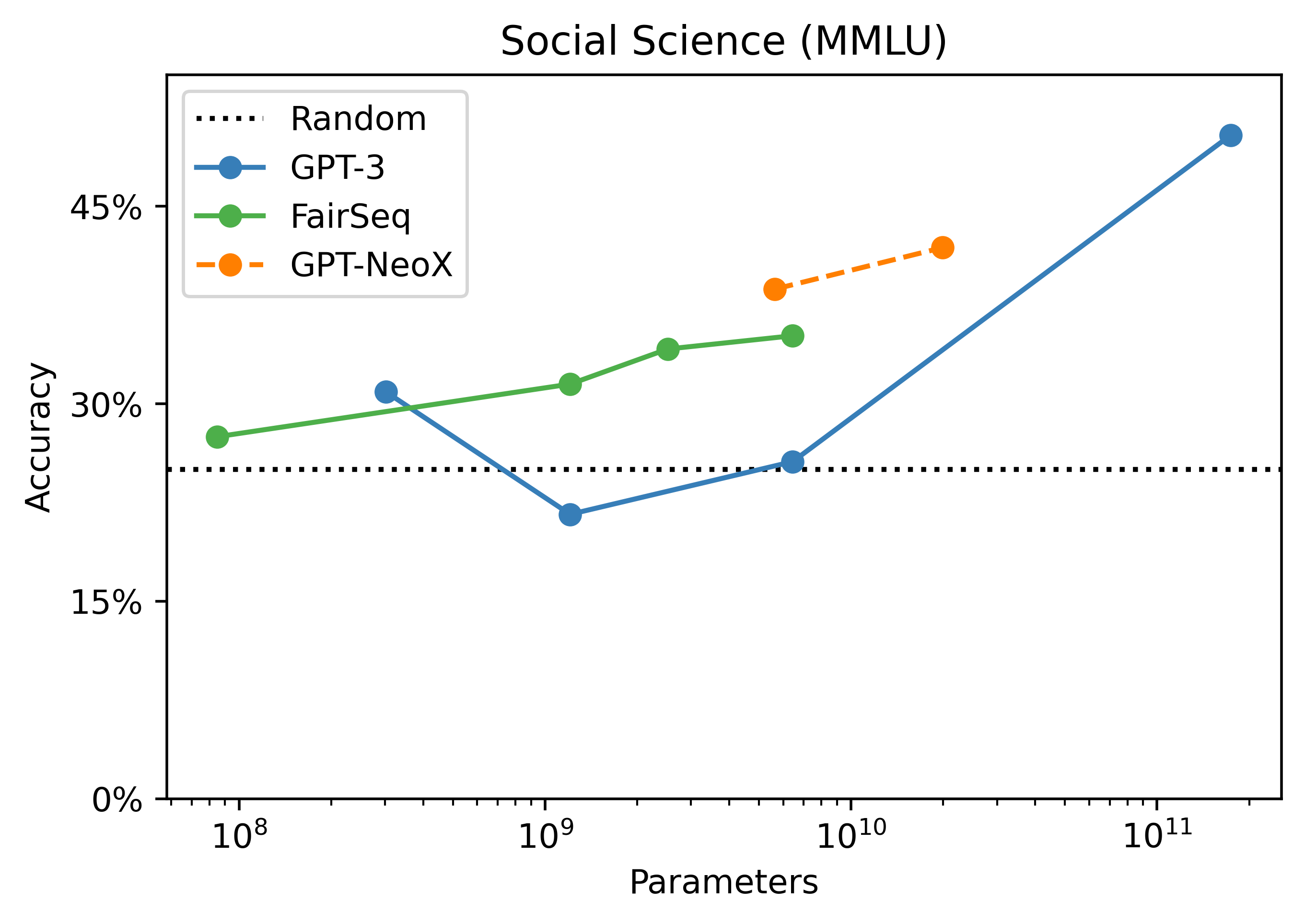}
    \includegraphics[width=0.45\textwidth]{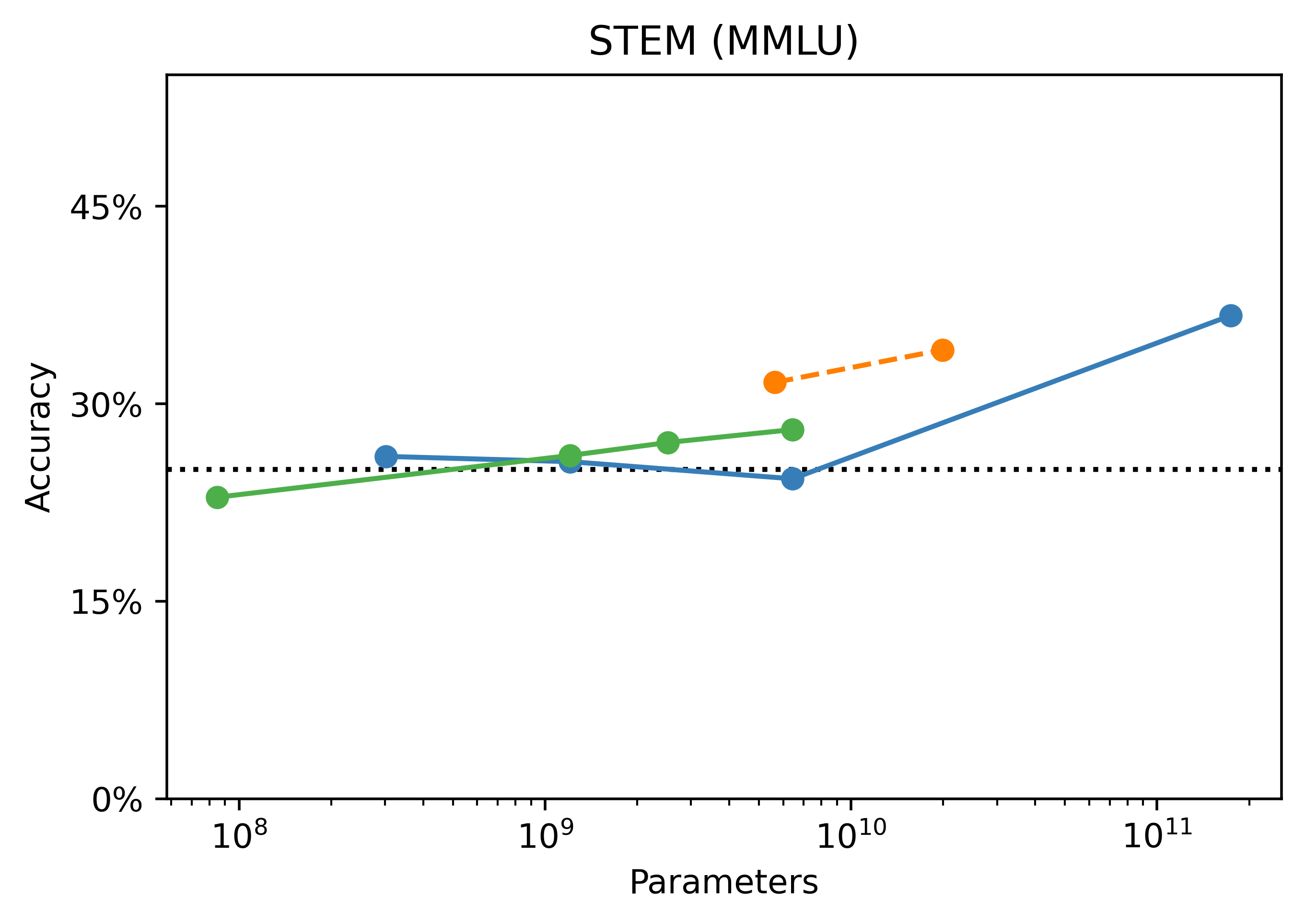}
    \includegraphics[width=0.45\textwidth]{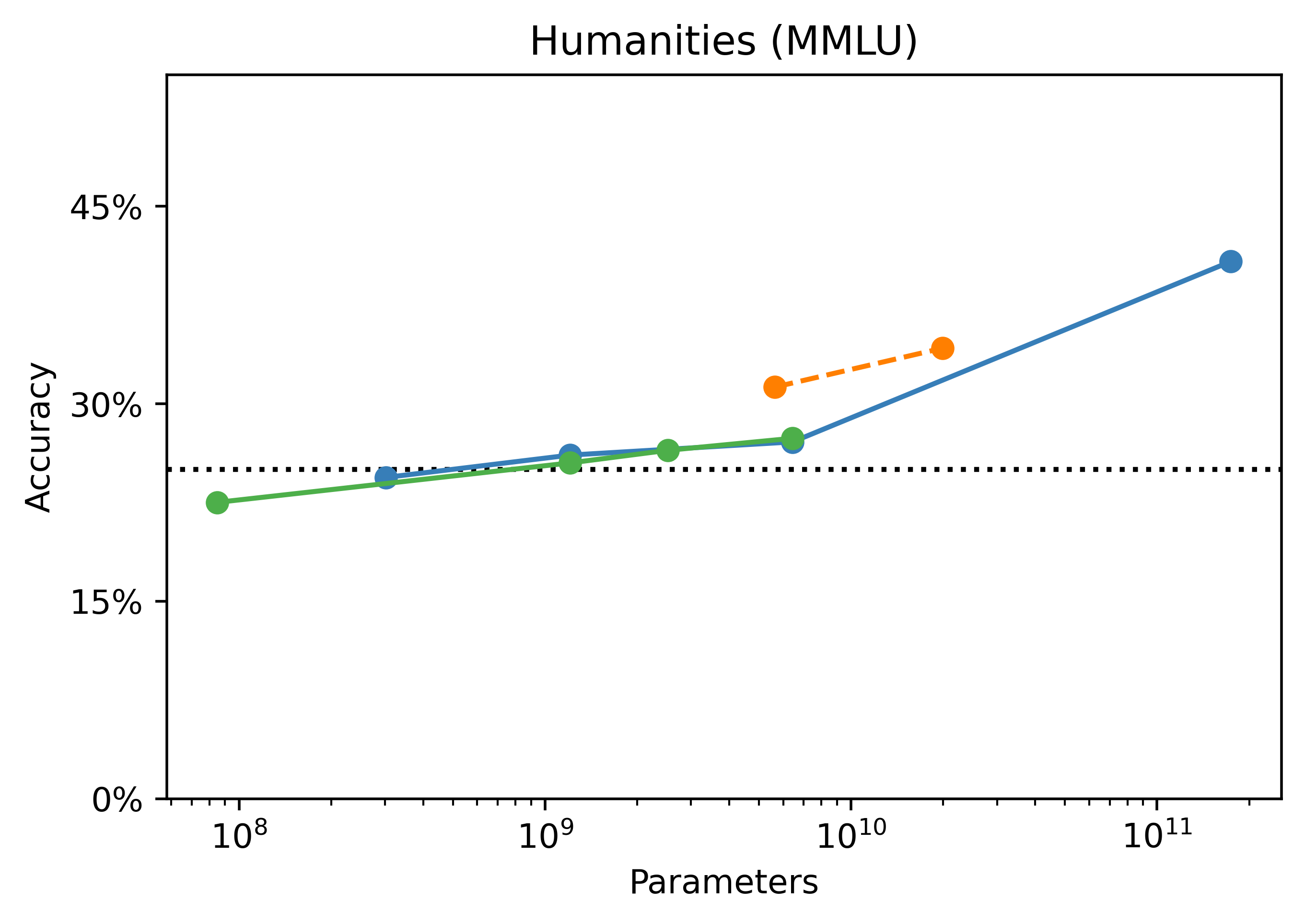}
    \includegraphics[width=0.45\textwidth]{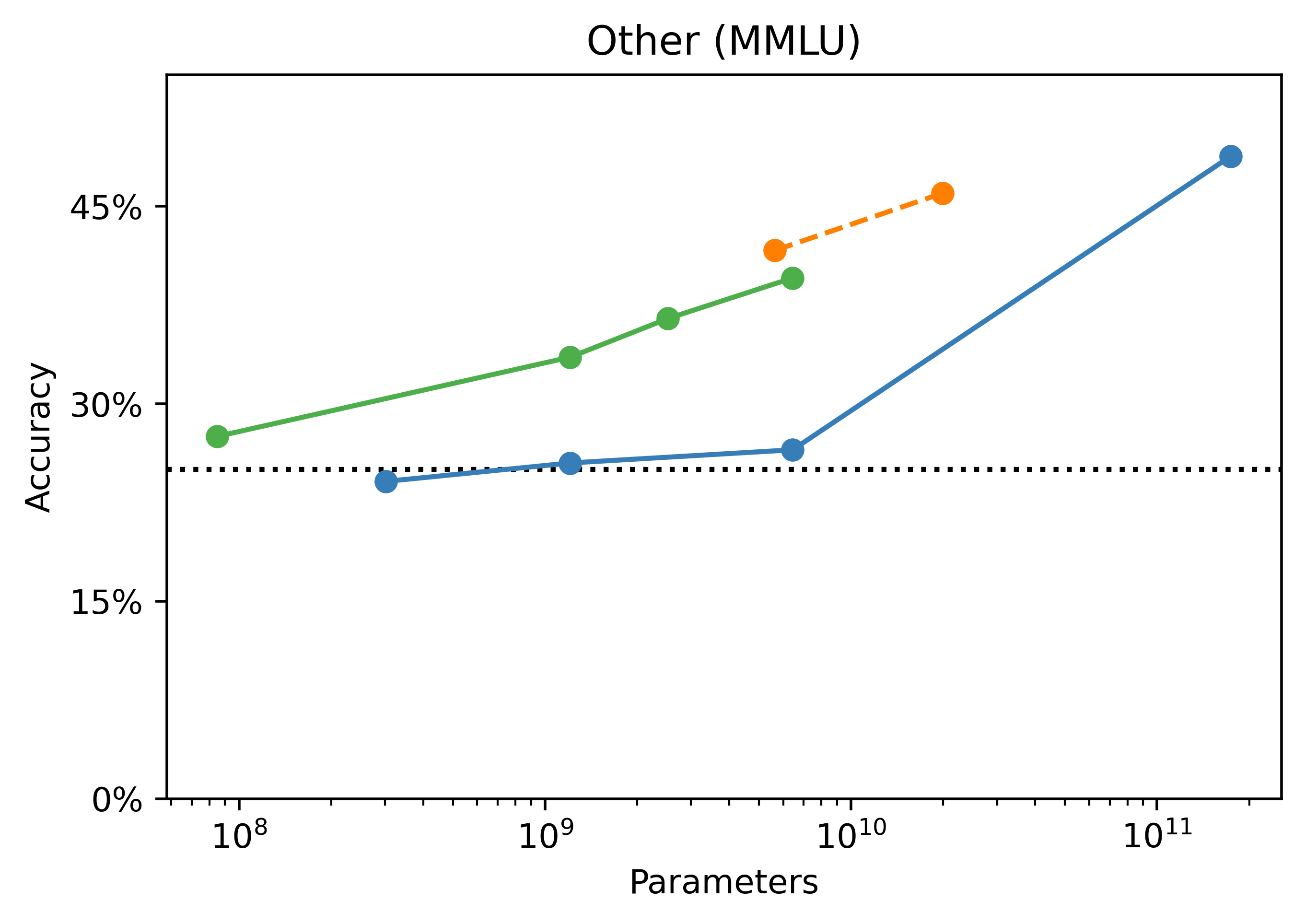}
    \caption{Five-shot performance of \model{} compared to GPT-J-6B and FairSeq and OpenAI models on \citet{hendrycks2020measuring}. Due to financial limitations we were unable to evaluate on the OpenAI API. Instead, we report numbers from \citet{hendrycks2020measuring} with model sizes corrected.}
    \label{fig:hendrycks-5}
\end{figure*}

\paragraph{Advanced Knowledge-Based Tasks} We are also interested in the ability of our models to answer factual questions that (for humans) require advanced knowledge. To do this, we use a dataset of multiple choice questions in a variety of diverse domains developed by \citet{hendrycks2020measuring}. Following common practice on this dataset, we focus on results aggregated by subject area: Humanities, Social Sciences, STEM, and Miscellaneous as presented in \Cref{fig:hendrycks-5}. We report five-shot performance to be comparable to previous work, taking our five-shot GPT-3 values from \citet{hendrycks2020measuring}.

\begin{figure*}
    \centering
    \includegraphics[width=0.3\textwidth]{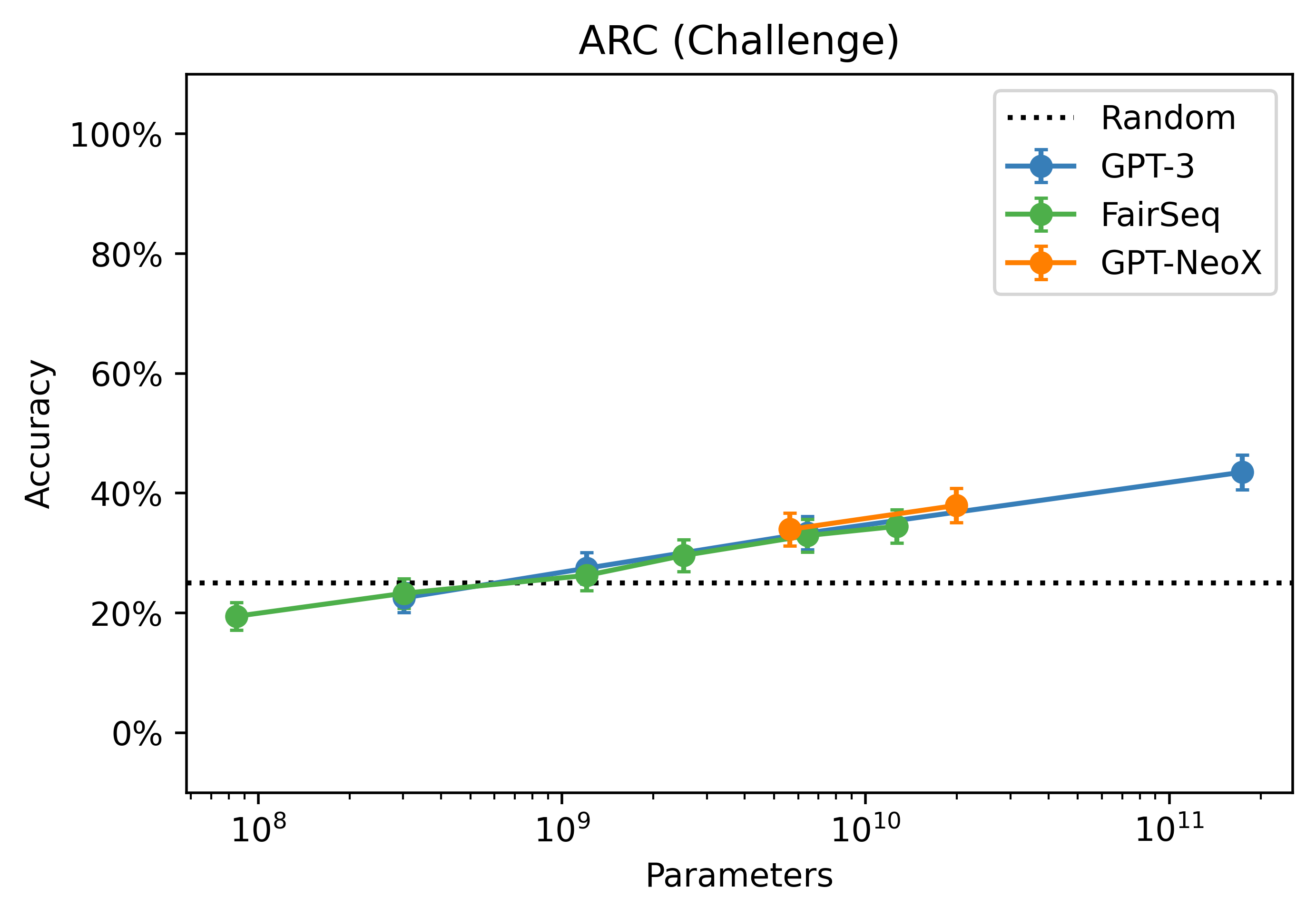}
    \includegraphics[width=0.3\textwidth]{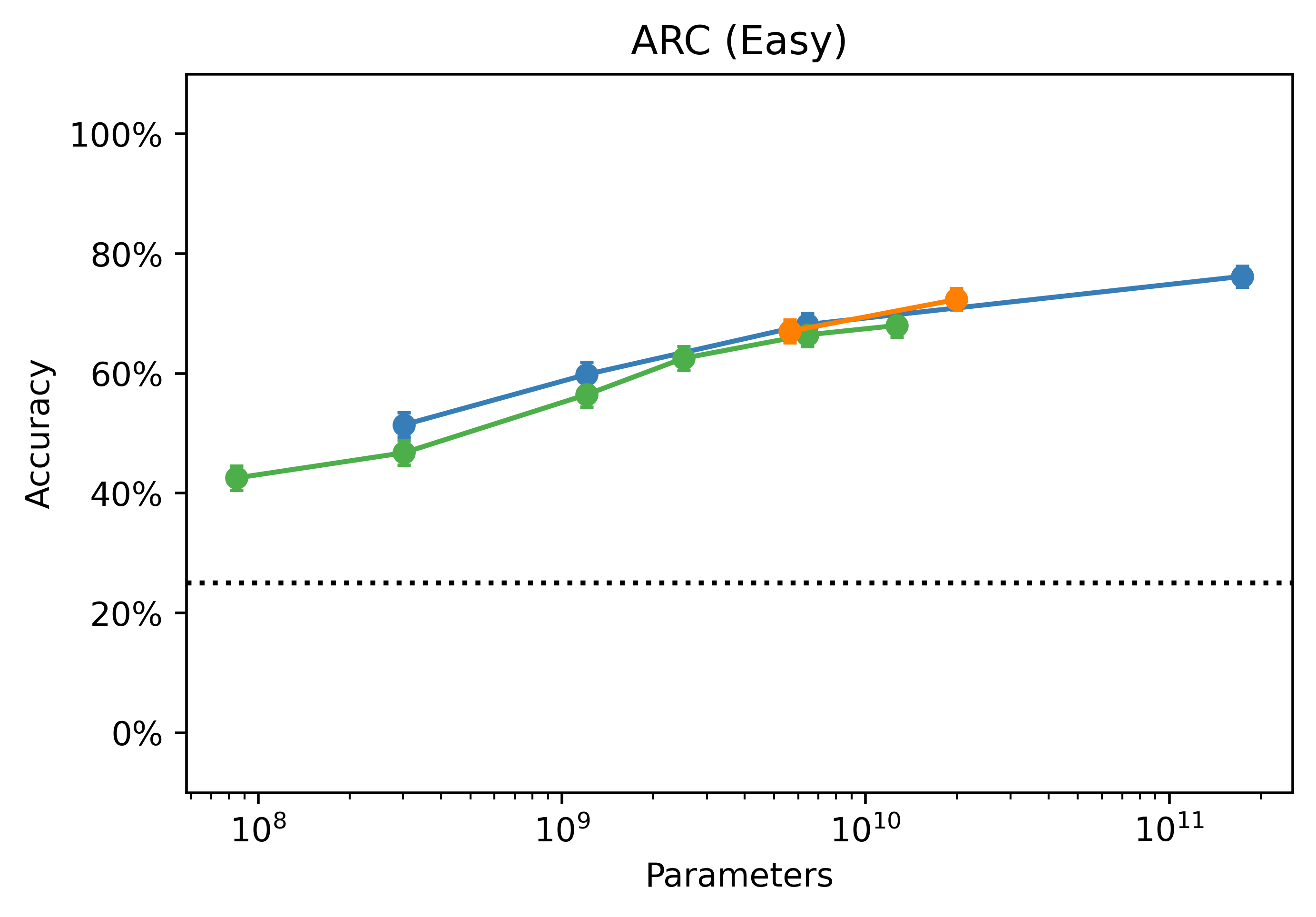}
    \includegraphics[width=0.3\textwidth]{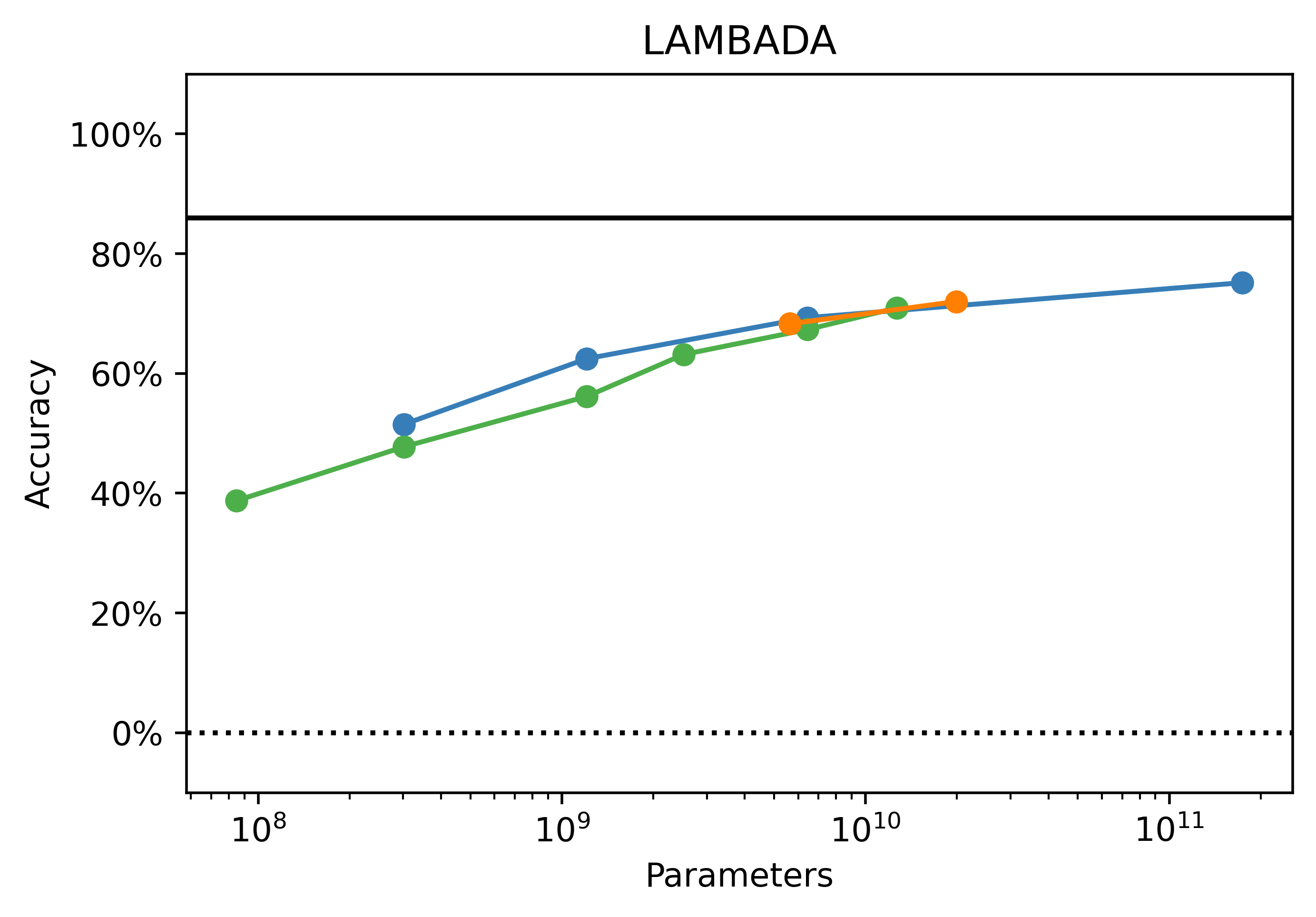}
    \includegraphics[width=0.3\textwidth]{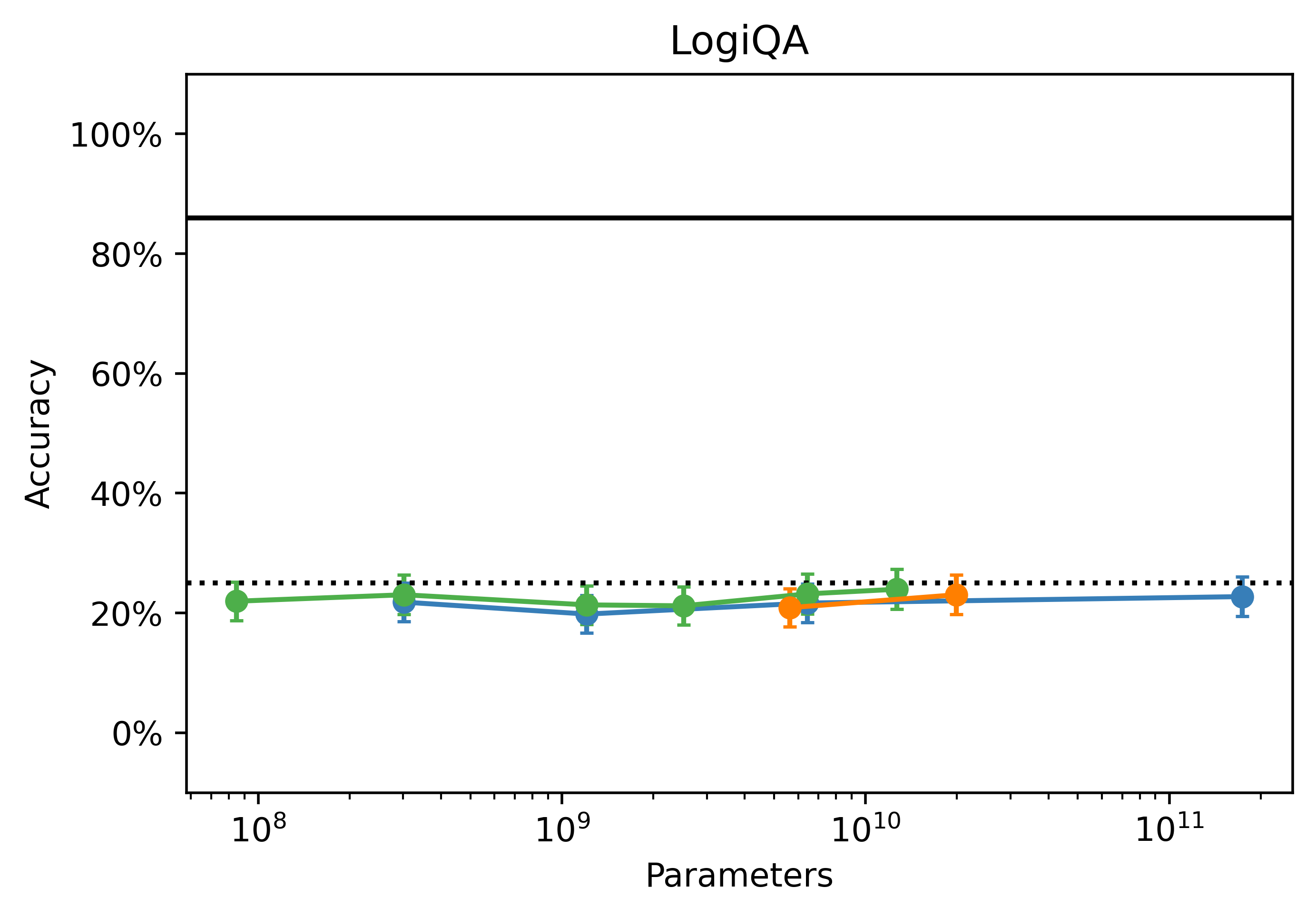}
    \includegraphics[width=0.3\textwidth]{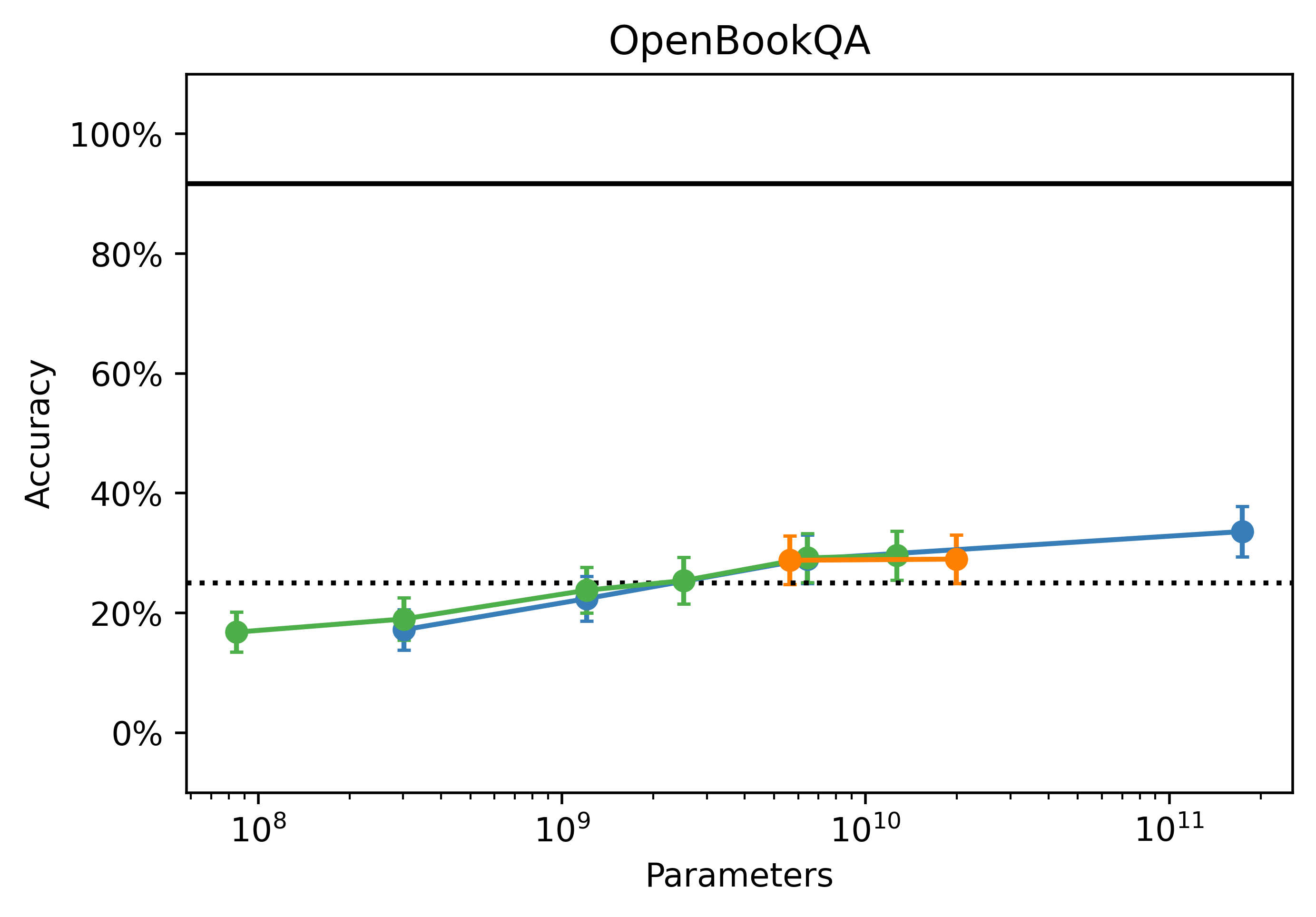}
    \includegraphics[width=0.3\textwidth]{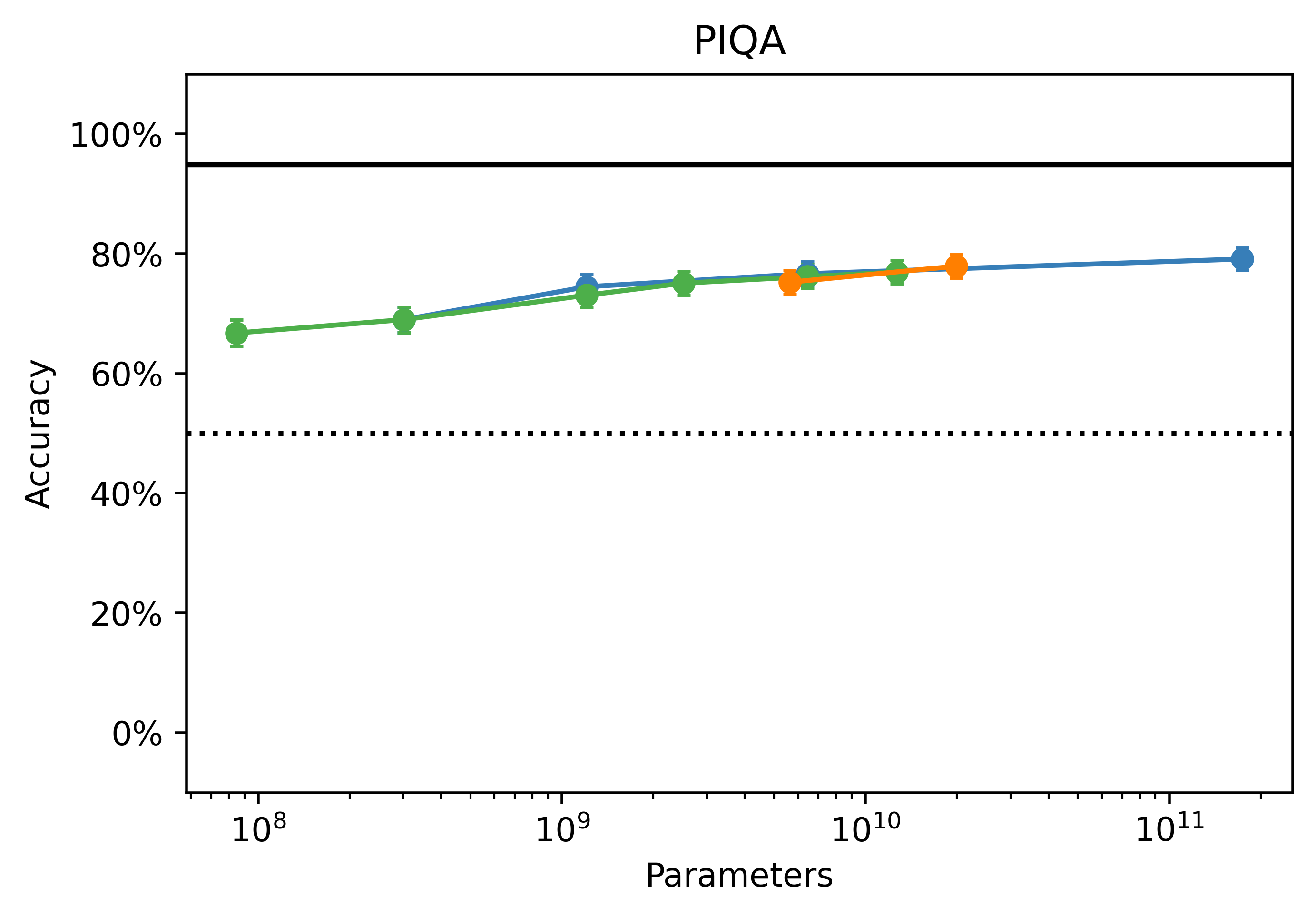}
    \includegraphics[width=0.3\textwidth]{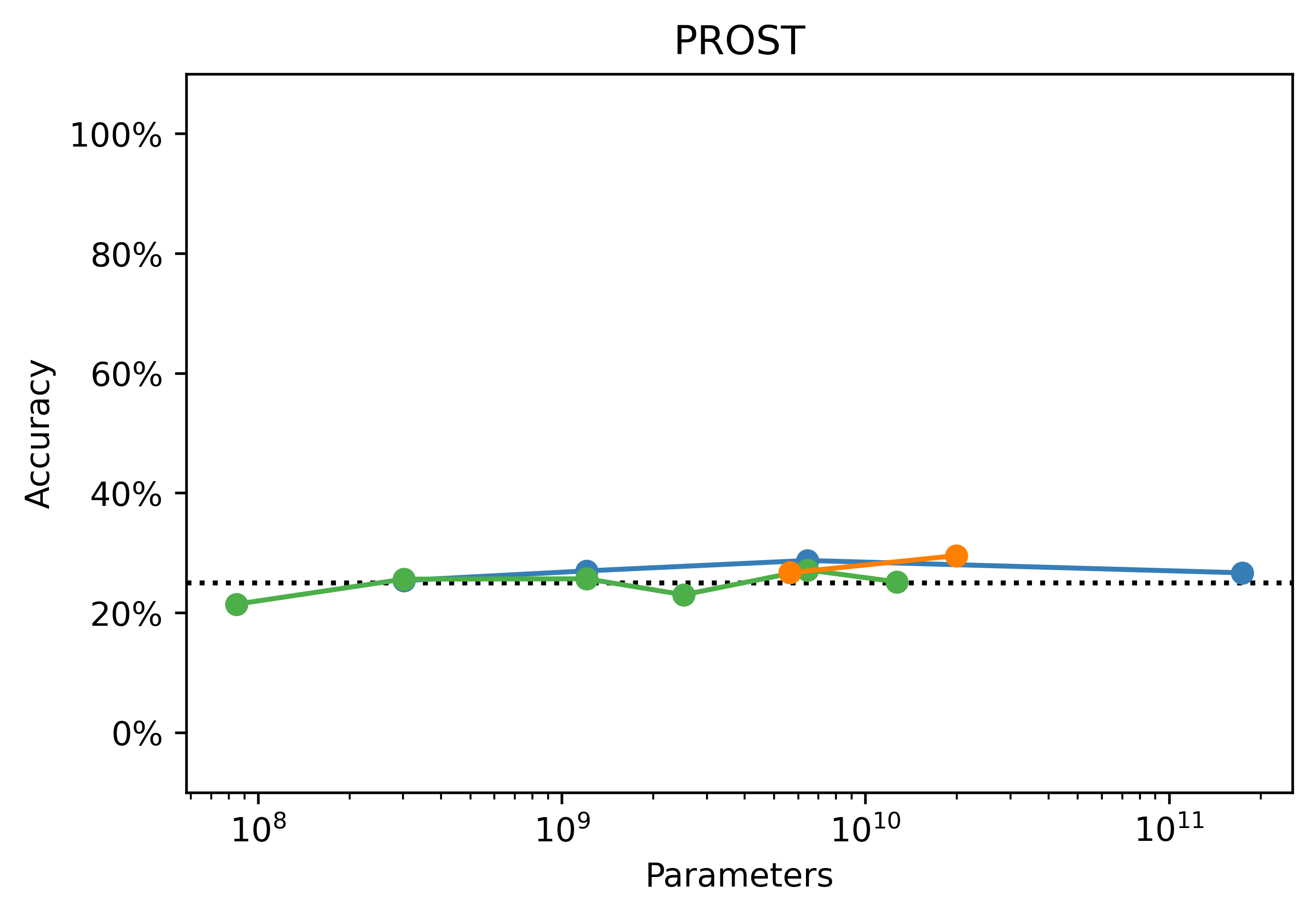}
    \includegraphics[width=0.3\textwidth]{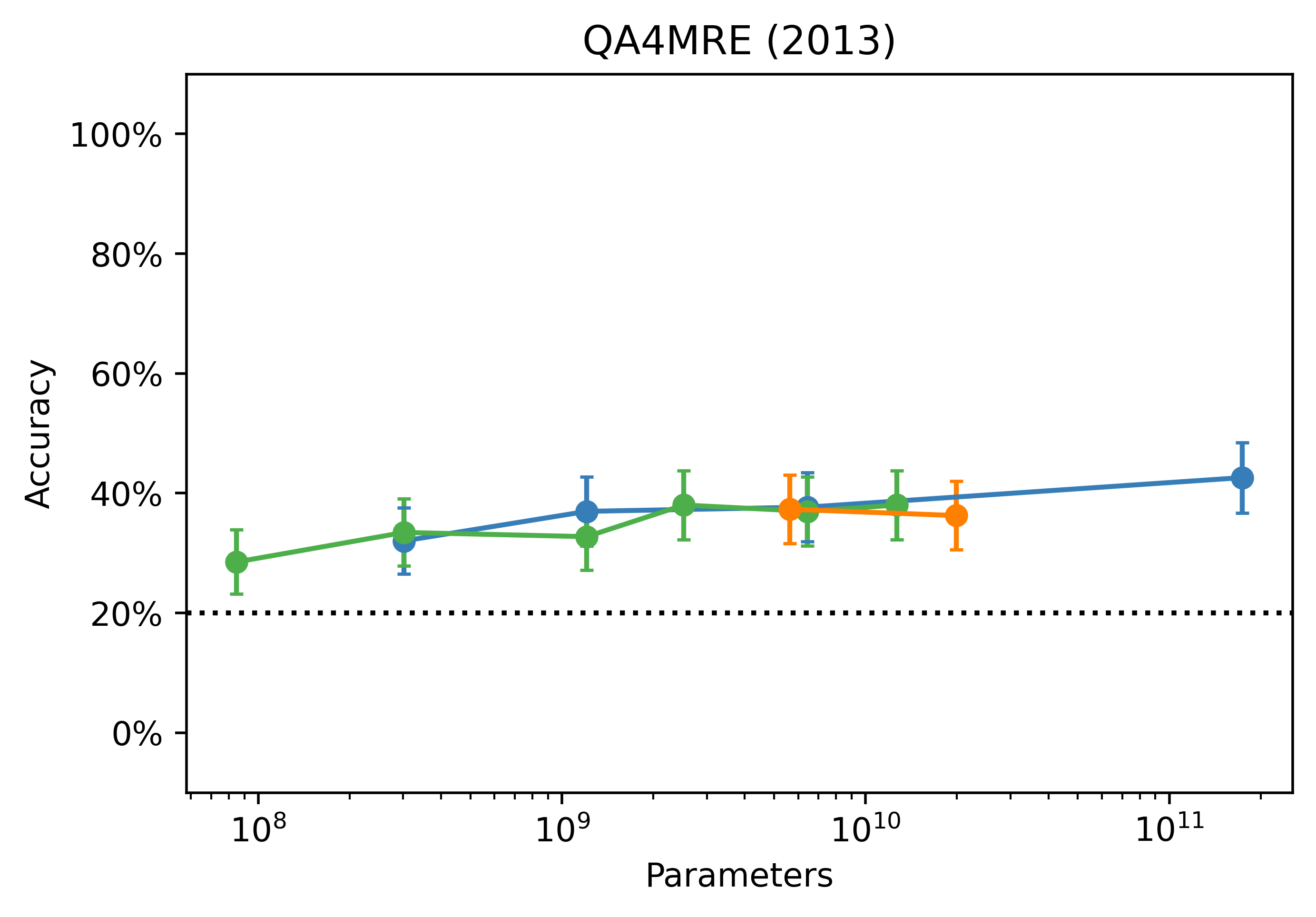}
    \includegraphics[width=0.3\textwidth]{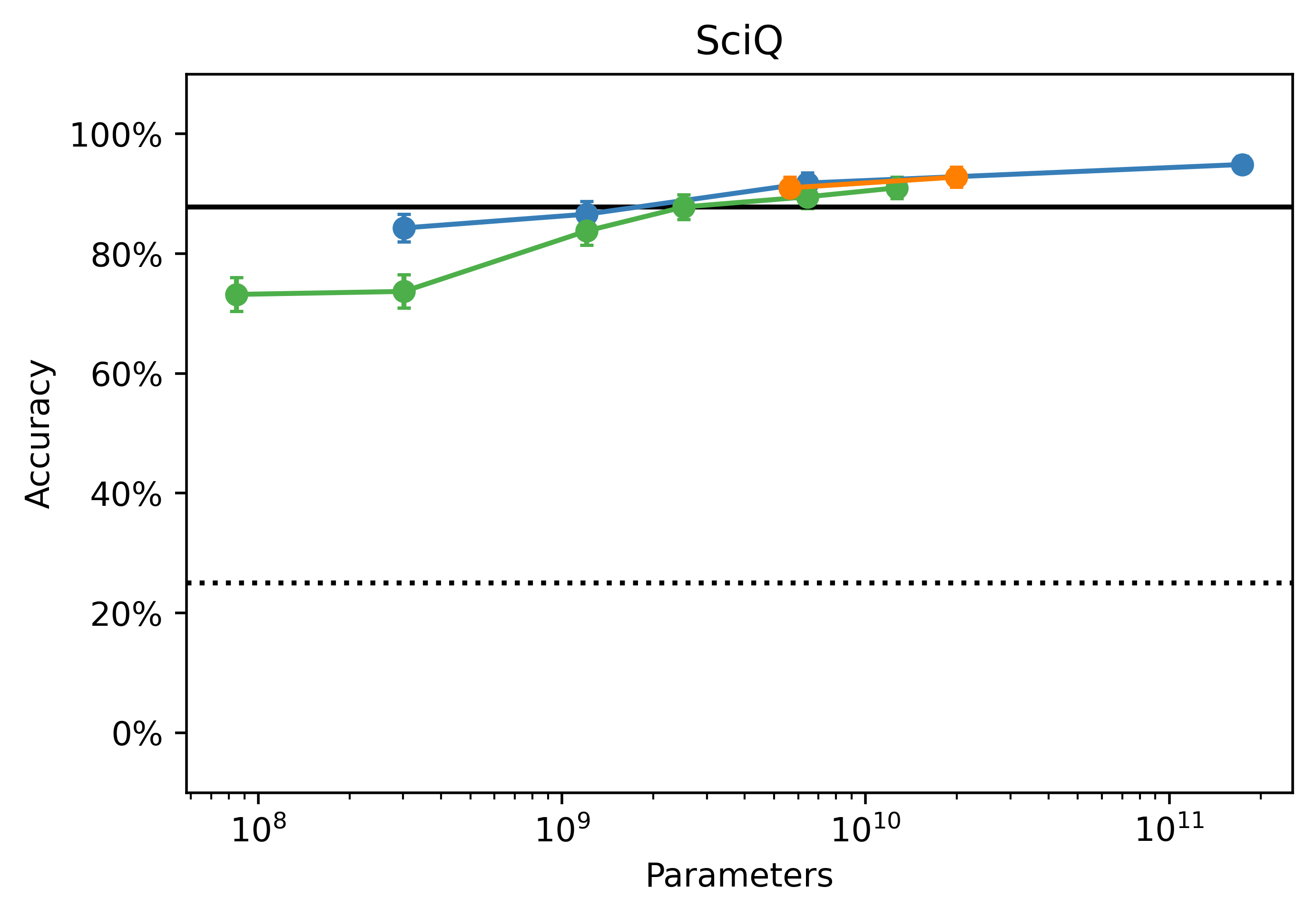}
    \includegraphics[width=0.3\textwidth]{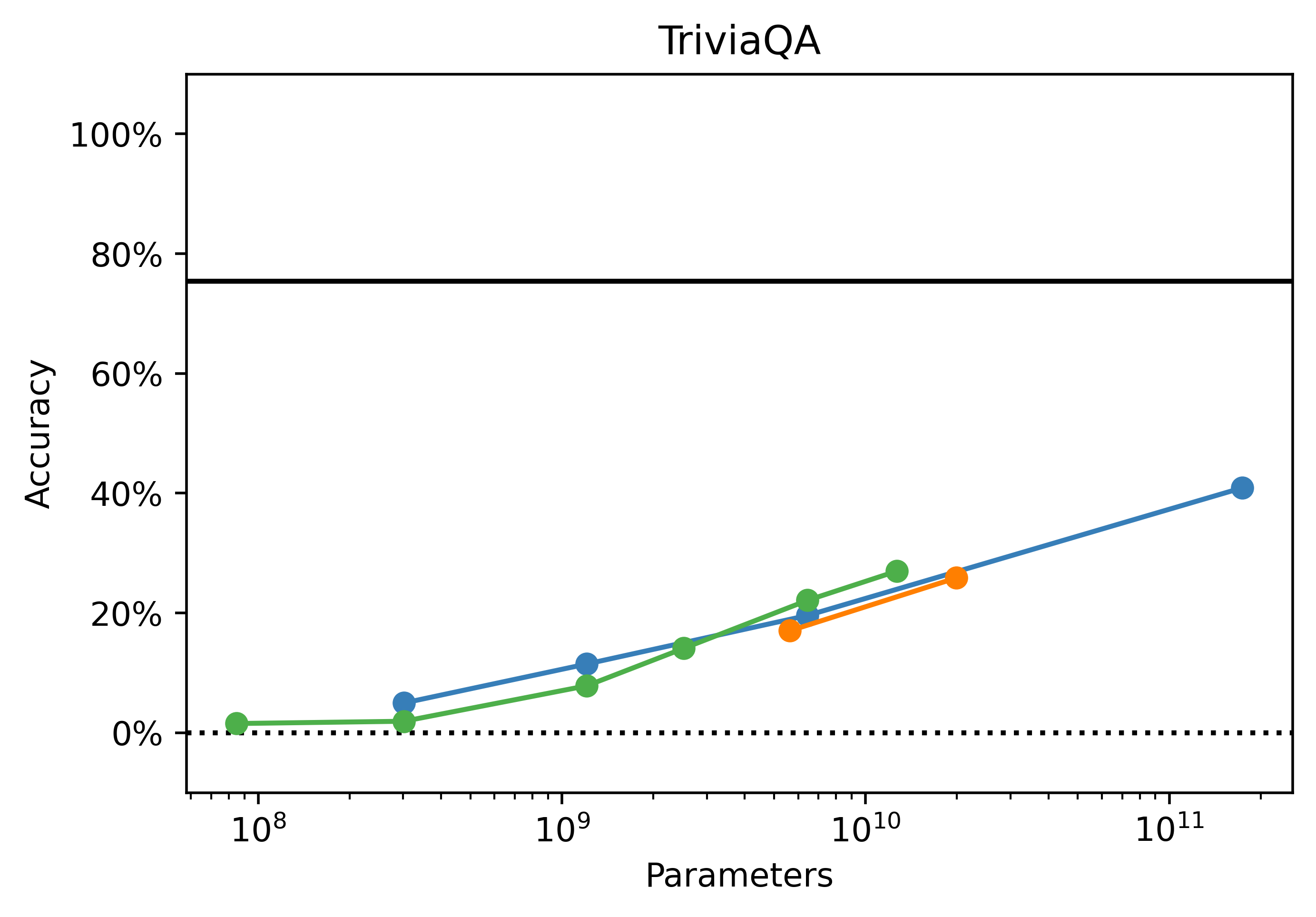}
    \includegraphics[width=0.3\textwidth]{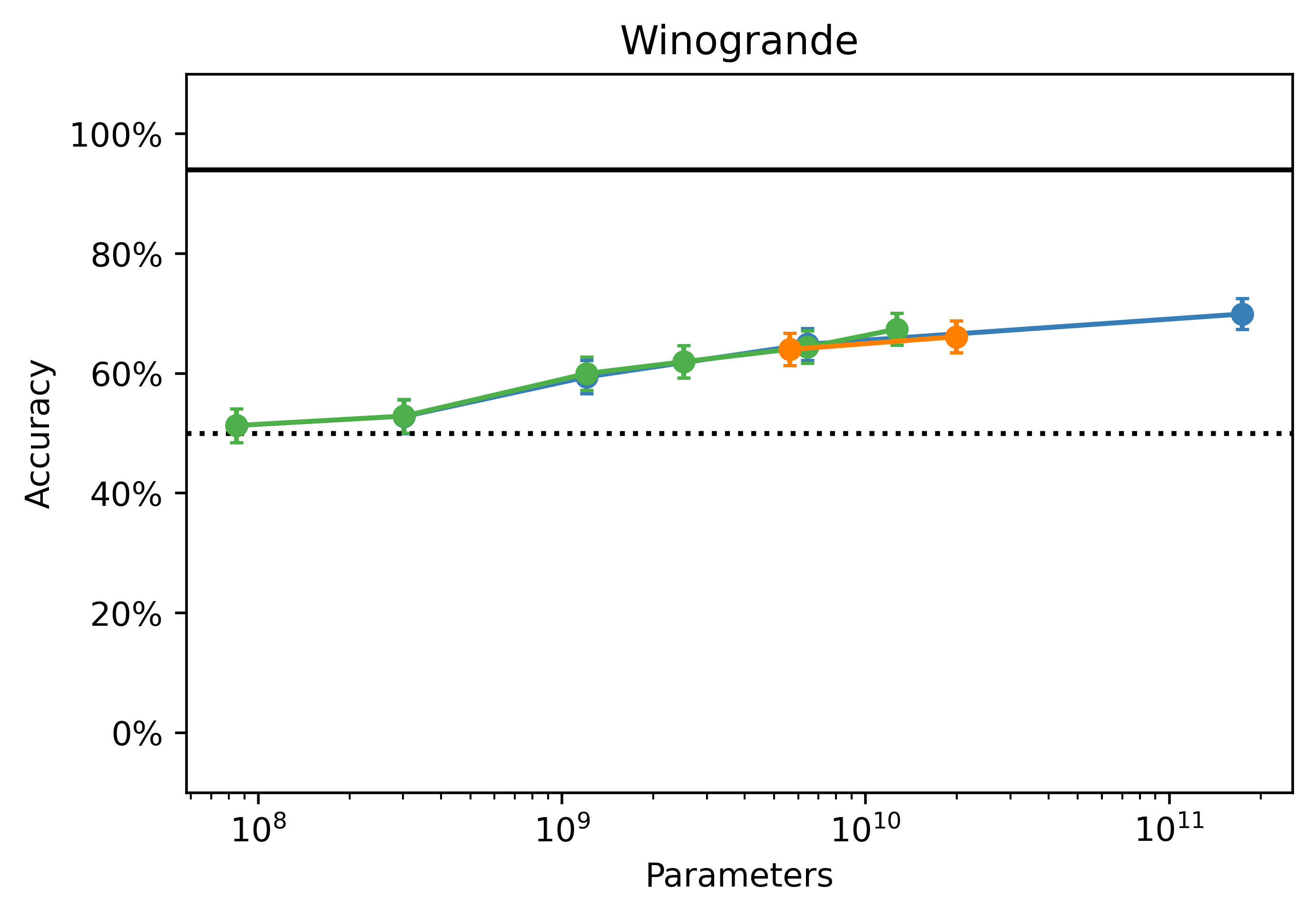}
    \includegraphics[width=0.3\textwidth]{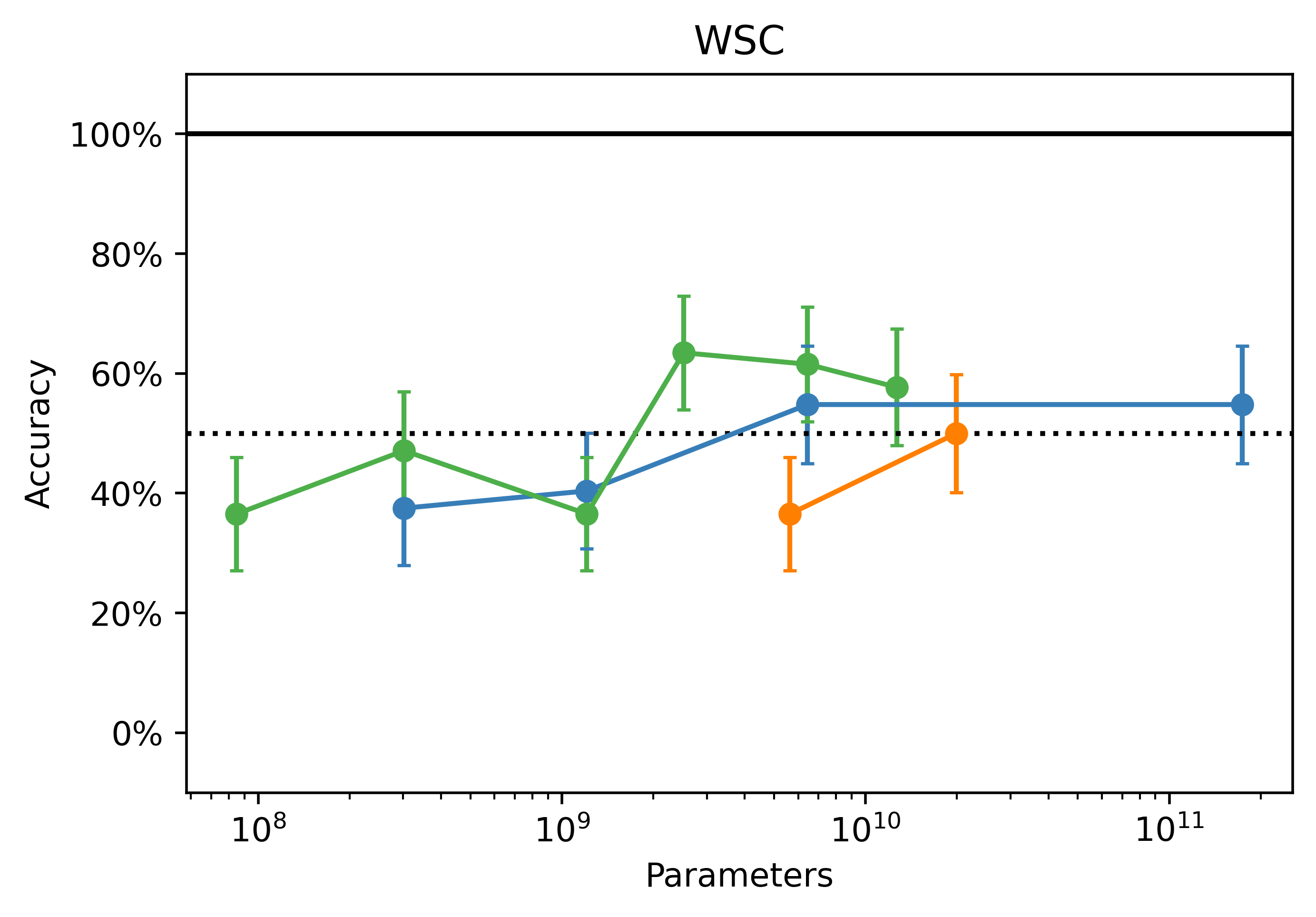}
    \caption{Zero-shot performance of \model{} compared to GPT-J-6B and FairSeq and OpenAI models on a variety of language modeling benchmarks.}
    \label{fig:language-full}
\end{figure*}

\begin{figure*}
    \centering
    \includegraphics[width=0.4\textwidth]{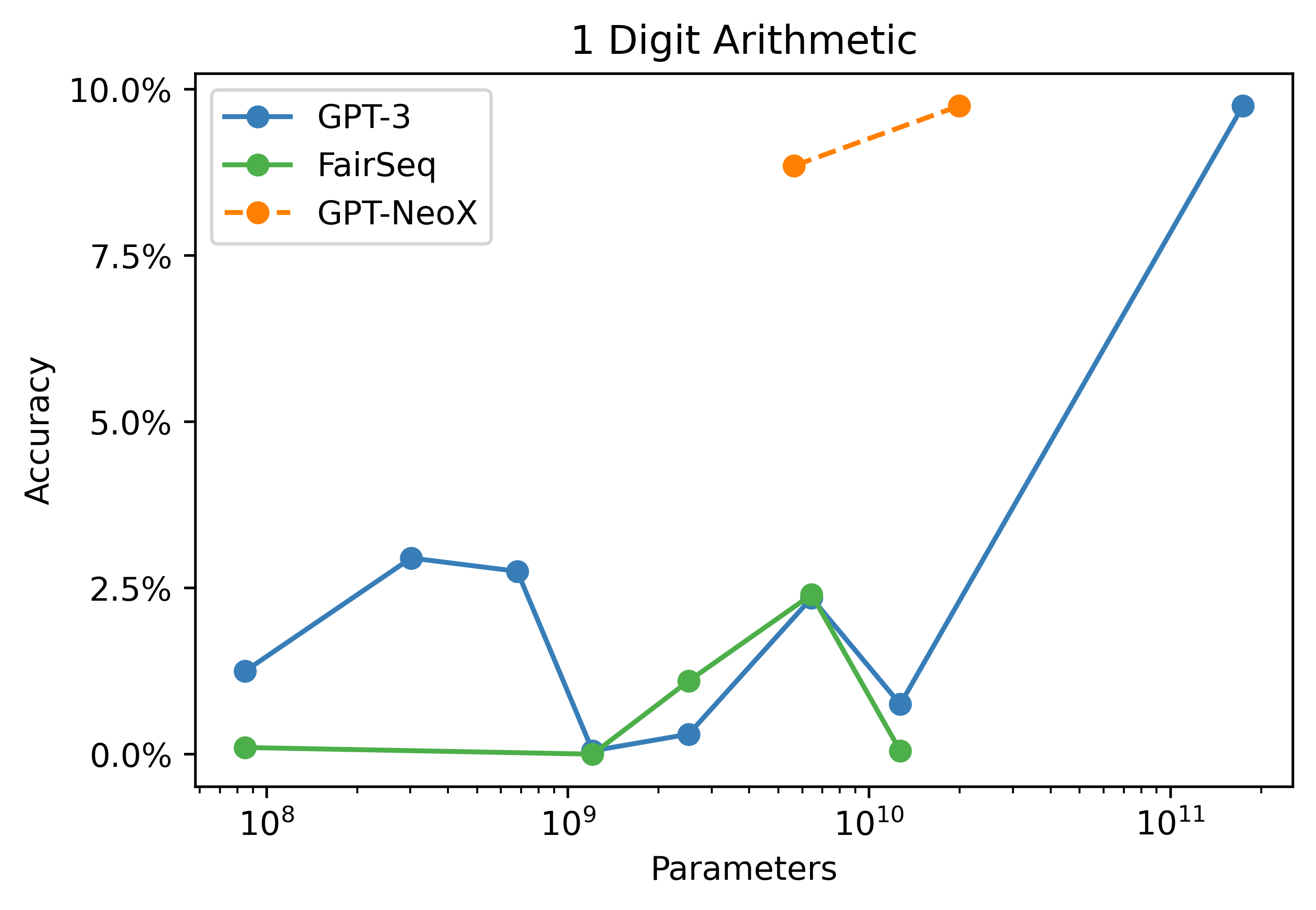}
    \includegraphics[width=0.4\textwidth]{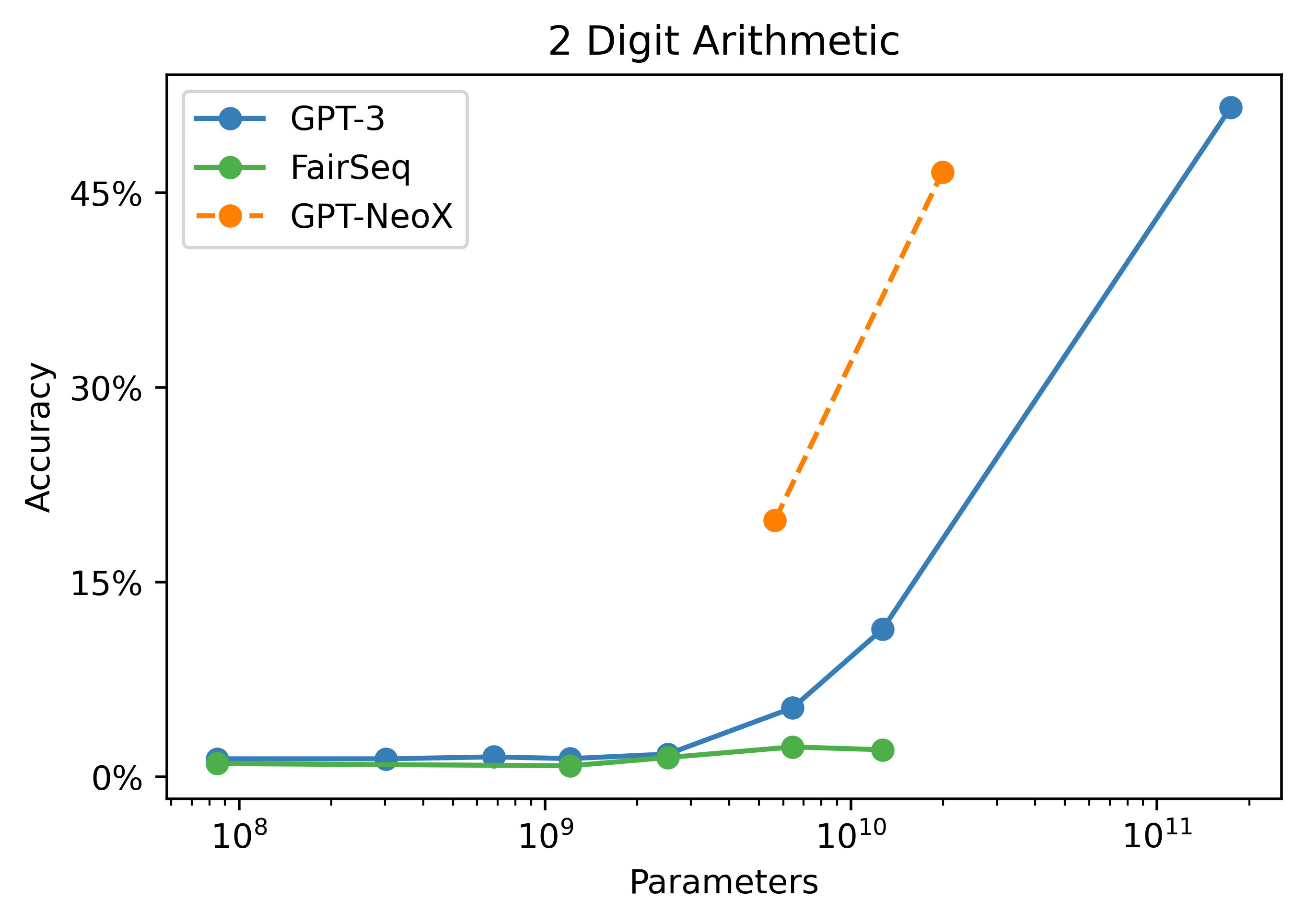}
    \includegraphics[width=0.4\textwidth]{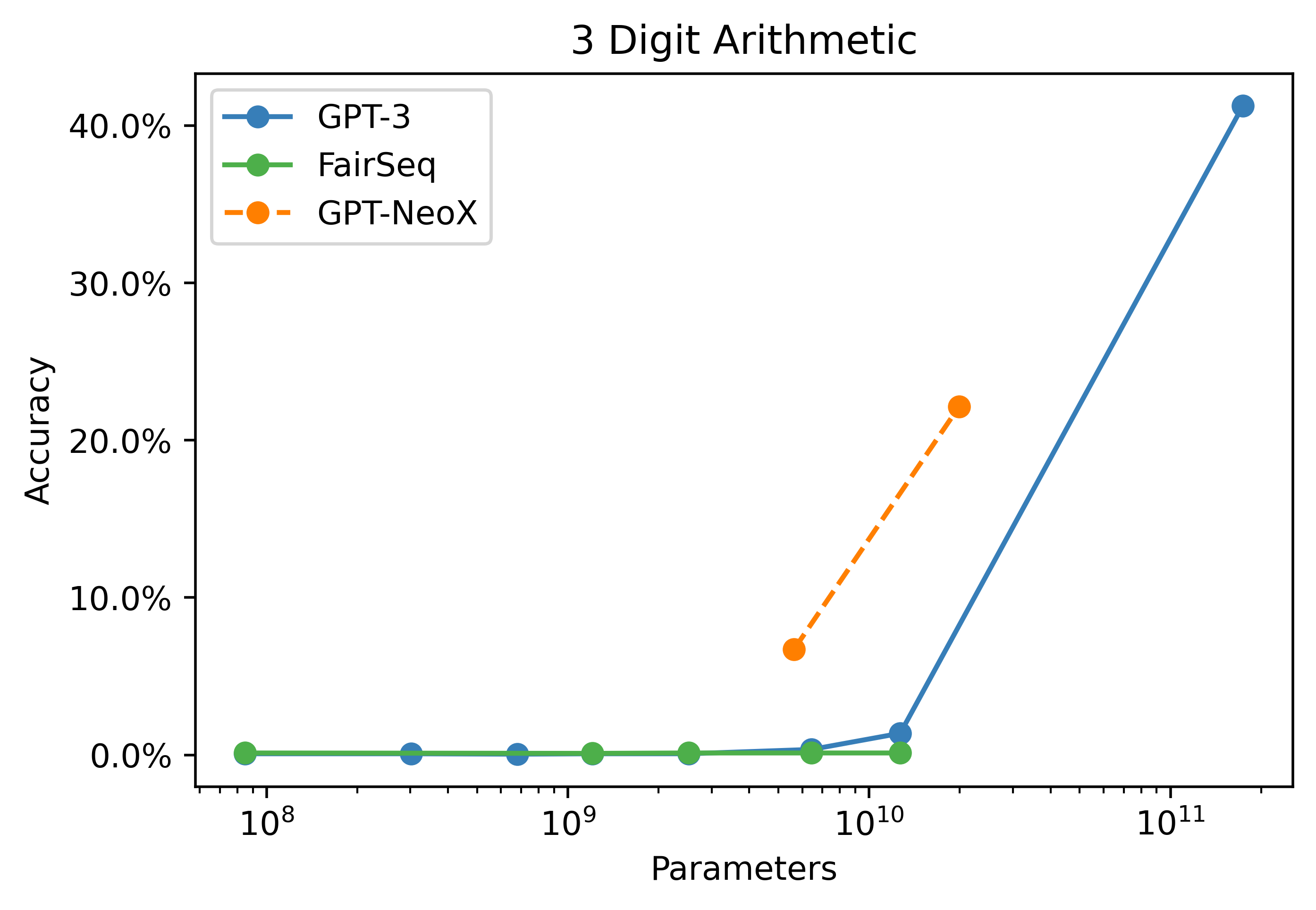}
    \includegraphics[width=0.4\textwidth]{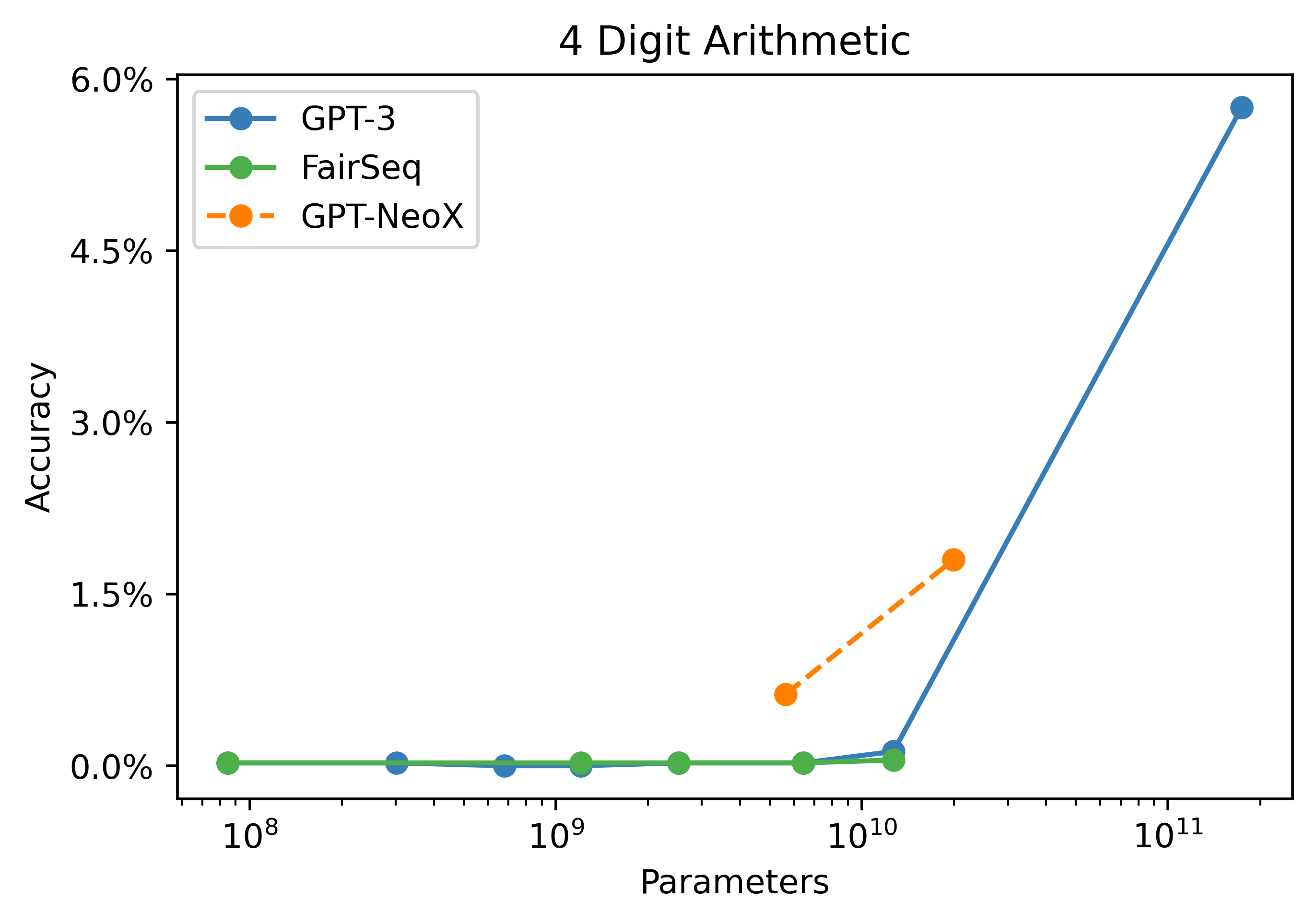}
    \includegraphics[width=0.4\textwidth]{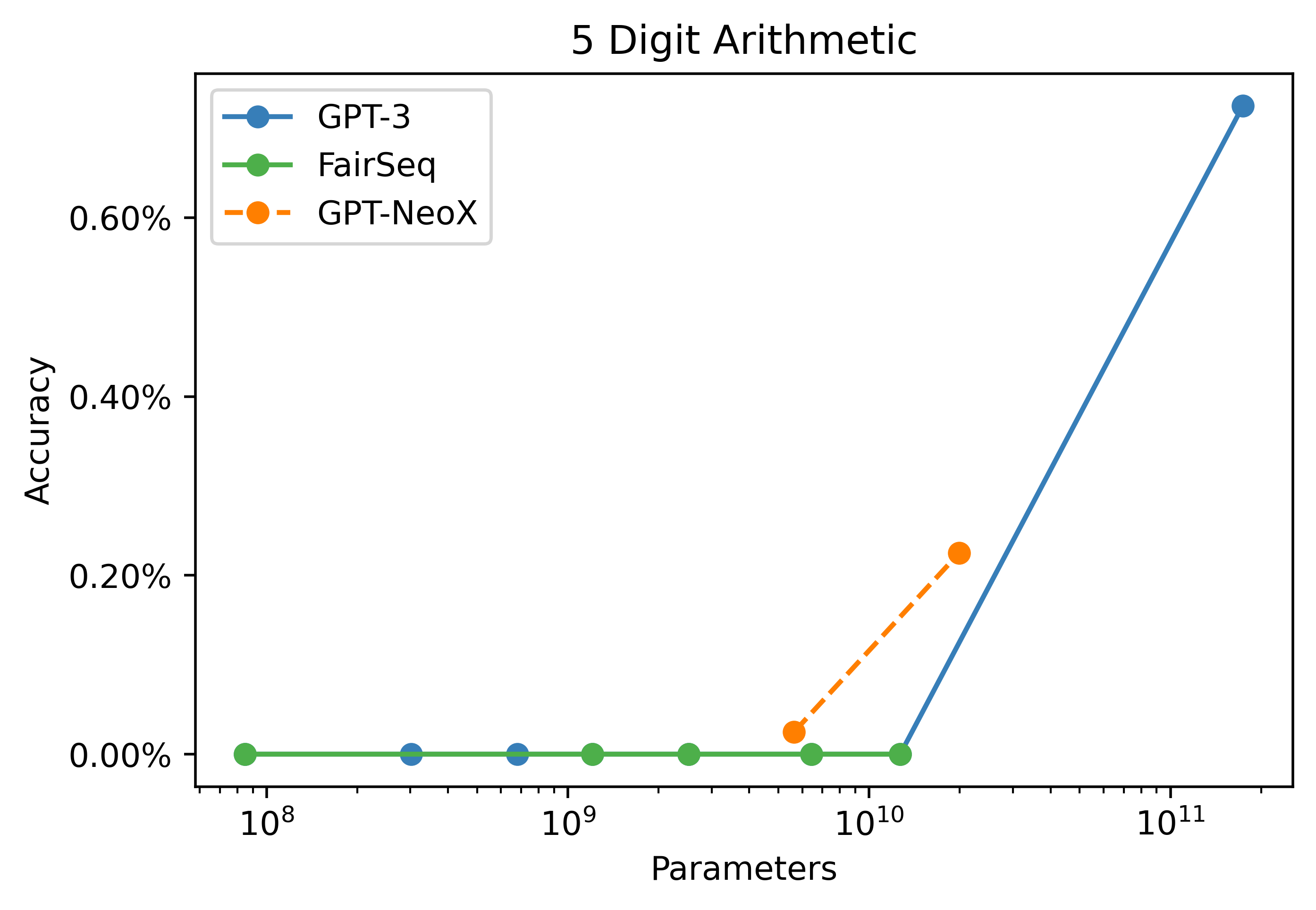}
    \includegraphics[width=0.4\textwidth]{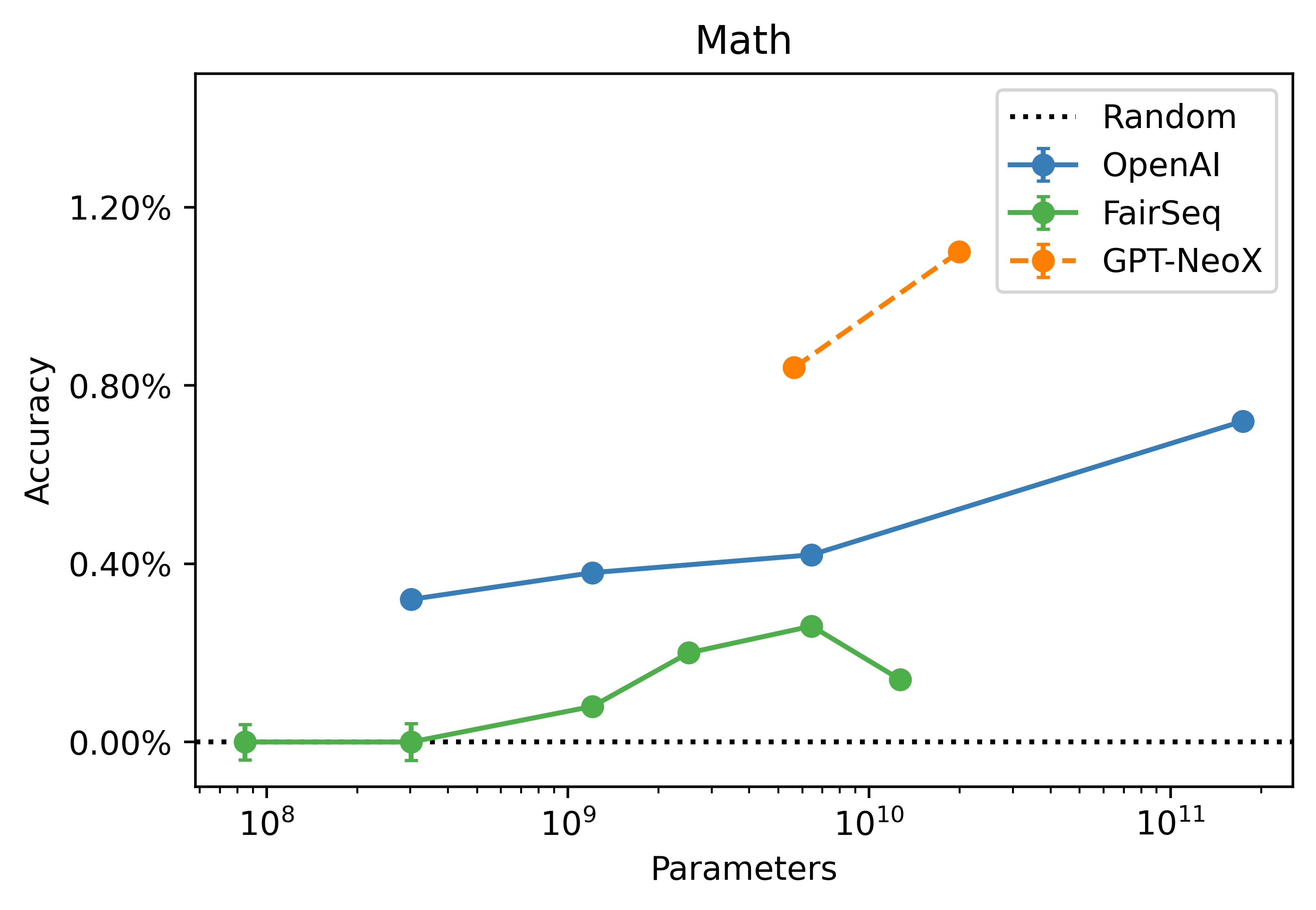}
    \caption{Zero-shot performance of \model{} compared to and FairSeq and OpenAI models on arithmetic tasks and MATH. Random performance on these tasks is $0\%$, and we were unable to find information on median human performance.}
    \label{fig:arithmetic}
\end{figure*}

\section{Discussion}

\subsection{Performance Results}

\paragraph{Natural Language Tasks}  While \model{} outperforms FairSeq 13B on some tasks (e.g. ARC, LAMBADA, PIQA, PROST), it underperforms on others (e.g. HellaSwag, LogiQA zero-shot). In total, across the 32 evaluations we did we outpreform on 22 tasks, underperform on four tasks, and fall within the margin of error on six tasks. By far our weakest performance is on HellaSwag, where we score four standard deviations below FairSeq 13B in both zero- and five-shot evaluations. Similarly, GPT-J underperforms FairSeq 6.7B by three standard deviations zero-shot and six standard deviations five-shot on HellaSwag. We find this massive performance loss largely inexplicable; while we originally assumed that the substantial non-prose components of the Pile were to blame, we note that GPT-J and GPT-NeoX \textit{overpreform} FairSeq models on the very similar Lambada task by roughly the same amount.

\paragraph{Mathematics} While GPT-3 and FairSeq models are generally quite close on arithmetic tasks, they are consistently out-performed by GPT-J and GPT-NeoX. We conjecture that this is traceable to the prevalence of mathematics equations in the training data, but warn that people should not assume that this means that training on the Pile produces better \textit{out-of-distribution} arithmetic reasoning. \citet{razeghi2022impact} show that there is a strong correlation between the frequency of a numerical equation in the Pile and GPT-J's performance on that equation, and we see no reason this would not hold in GPT-NeoX 20B, FairSeq, and GPT-3. We are unfortunately unable to investigate this effect in FairSeq and GPT-3 models because the authors do not release their training data.

\paragraph{Advanced Knowledge-Based Tasks} While GPT-NeoX and FairSeq models both exhibit dominant performance on MMMLU compared to GPT-3 in the five-shot setting (\Cref{fig:hendrycks-5}), their performance is much closer in the zero-shot setting (\Cref{tab:hendrycks-gpt1,tab:hendrycks-gpt2,tab:hendrycks-fair1,tab:hendrycks-fair2}). \citet{hendrycks2021measuring} claim to find that few-shot evaluation does not improve performance relative to zero-shot, but they only study GPT-3. By contrast, we find that GPT-NeoX and FairSeq models do improve substantially with as few as five examples. We view this as a warning against drawing strong conclusions about evaluation metrics based only on one model, and encourage researchers developing new evaluation benchmarks to leverage multiple different classes of models to avoid overfitting their conclusions to a specific model.

\subsection{Powerful Few-Shot Learning}

Our experiments indicate that GPT-J-6B and \model{} benefit substantially more from few-shot evaluations than the FairSeq models do. When going from 0-shot to 5-shot evaluations, GPT-J-6B improves by $0.0526$ and \model{} improves by $0.0598$ while the FairSeq 6.7B and 13B models improve by $0.0051$ and $0.0183$ respectively. This result is statistically significant and robust to perturbations of prompting. While we do not have a particular explanation for this currently, we view this as a strong recommendation for our models. While we do not have systematic five-shot evaluations of GPT-3 due to financial limitations, the change in performance demonstrated in \cref{tab:hendrycks-gpt1,tab:hendrycks-gpt2,tab:hendrycks-fair1,tab:hendrycks-fair2,fig:hendrycks-5} further supports the suggestion that GPT-J-6B and GPT-NeoX-20B are able to gain significantly more utility from five-shot examples.

\subsection{Limitations}

\paragraph{Optimal Training} Hyperparameter tuning is an expensive process, and is often infeasible to do at full scale for multi-billion parameter models. Due to the aforementioned limitations, we opted to choose hyperparameters based on a mixture of experiments at smaller scales and by interpolating parameters appropriate for our model size based on previously published work \citep{brown2020language}. However, several aspects of both our model architecture [\Cref{subsec:model-arch}] and training setup, including the data [\Cref{sec:training-data}] and the tokenizer [\Cref{sec:tokenization}], diverge significantly from \citet{brown2020language}. As such, it is almost certainly the case that the hyperparameters used for our model are no longer optimal, and potentially never were.

\paragraph{Lack of Coding Evaluations} Many of the design choices we made during the development of this model were oriented towards improving performance on coding tasks. However, we underestimated the difficulty and cost of existing coding benchmarks \citep{chen2021evaluating}, and so were unable to evaluate out model in that domain. We hope to do so in the future.

\paragraph{Data Duplication} Finally, the lack of dataset deduplication could also have had an impact on downstream performance. Recent work has shown that deduplicating training data can have a large effect on perplexity \citep{lee2021deduplicating}. While our experiments show no sign of this, it is hard to dismiss it due to the number of researchers who have found the opposite result.

\subsection{Releasing a 20B Parameter LLM}

The current status quo in research is that large language models are things people train and publish about, but do not actually release. To the best of our knowledge, \model{} is the largest and most performant dense language model to ever be publicly released. A variety of reasons for the non-release of large language models are given by various groups, but the primary one is the harms that public access to LLMs would purportedly cause.

We take these concerns quite seriously. However, having taken them quite seriously, we feel that they are flawed in several respects. While a thorough analysis of these issues is beyond the scope of this paper, the public release of our model is the most important contribution of this paper and so an explanation of why we disagree with the prevailing wisdom is important.

\paragraph{Providing access to ethics and alignment researchers will prevent harm.} The open-source release of this model is motivated by the hope that it will allow researchers who would not otherwise have access to LLMs to use them. While there are negative risks due to the potential acceleration of capabilities research, we believe the benefits of this release outweigh the risks. We also note that these benefits are not hypothetical, as a number of papers about the limits and ethics of LLMs has been explicitly enabled by the public release of previous models \citep{zhang2021counterfactual,kandpal2022deduplicating,carlini2022quantifying,birhane2021multimodal,logit-lens,meng2022locating, lin2021truthfulqa}.

\paragraph{Limiting access to governments and corporations will not prevent harm.} Perhaps the most curious aspect of the argument that LLMs should not be released is that the people making such arguments are not arguing they \textit{they} should not use LLMs. Rather, they are claiming that \textit{other people} should not use them. We do not believe that this is a position that should be taken seriously. The companies and governments that have the financial resources to train LLMs are overwhelmingly more likely to do large scale harm using a LLM than a random individual.

Releasing this model is the beginning, not the end, of our work to make \model{} widely accessible to researchers. Due to the size of the model, inference is most economical on a pair of RTX 3090 Tis or a single A6000 GPU and finetuning requires significantly more compute. Truly promoting widespread access to LLMs means promoting widespread access to \textit{computing infrastructure} in addition to the models themselves. We plan to make progress on this issue going forward by continuing to work on reducing the inference costs of our model, and by working with researchers to provide access to the computing infrastructure they need to carry out experiments on our models. We strongly encourage researchers who are interested in studying \model{} but lack the necessary infrastructure to reach out to discuss how we can help empower you.

\section{Summary}

We introduce \model{}, a 20 billion parameter autoregressive Transformer language model trained on the Pile \citep{gao2020pile} dataset, and detail the main architectural differences between \model{} and GPT-3---most notably the change in tokenizer, the addition of Rotary Positional Embeddings, the parallel computation of attention and feed-forward layers, and a different initialization scheme and hyperparameters. We run extensive evaluations of \model{} on natural language and factual knowledge tasks, and compare it with other publicly available models, finding it performs particularly well on knowledge-based and mathematical tasks. Finally, we are open sourcing the training and evaluation code at \url{https://github.com/EleutherAI/gpt-neox}, where readers can find a link to download the model weights across the whole training run.

\section*{Acknowledgments}

We thank staff at CoreWeave---in particular Max~Hjelm, Brannin~McBee, Peter~Salanki, and Brian~Venturo---for providing the GPUs and computing infrastructure that made this project possible. We would also like to acknowledge Eren~Do\u{g}an and Wesley~Brown for feedback and technical support throughout the project, and John~Schulman, Evan~Hubinger, Victor~Sanh, Jacob~Hilton, and Siddharth~Karamcheti for providing feedback on drafts of the paper.

Finally, we thank Anthony~DiPofi, Charles~Foster, Jeffrey~Hsu, Eric~Tang, Anish~Thite, Kevin~Wang, and Andy~Zou for their contributions to the EleutherAI Language Modeling Evaluation Harness we used to evaluate \model{}.

\bibliography{acl_citations}
\bibliographystyle{acl_natbib}

\clearpage

\appendix

\section{Individual Contributions}
\label{app:contrib}

\textbf{Sid Black} was the lead developer and overall point person for the project. \textbf{Stella Biderman} was the lead scientist and project manager.

\subsection*{Implementation and Engineering}
\begin{adjustwidth}{\parindent}{}
\setlength{\parindent}{0pt}
\textbf{Implementation of training infrastructure:}\\\mbox{Sid Black}, \mbox{Stella Biderman}, \mbox{Eric Hallahan}, \mbox{Quentin Anthony}, \mbox{Samuel Weinbach} \setlength{\parskip}{0.5em}\par
\textbf{Scaling experiments and optimization:}\\\mbox{Sid Black}, \mbox{Stella Biderman}, \mbox{Quentin Anthony}, \mbox{Samuel Weinbach} \par
\textbf{Positional Embeddings:}\\\mbox{Sid~Black}, \mbox{Eric Hallahan}, \mbox{Michael~Pieler} \par
\textbf{Tokenizer:}\\Sid~Black \par
\textbf{Miscellaneous:}\\\mbox{USVSN~Sai~Prashanth}, \mbox{Ben~Wang} \par
\end{adjustwidth}

\subsection*{Scientific Experimentation}
\begin{adjustwidth}{\parindent}{}
\setlength{\parindent}{0pt}
\textbf{Evaluations:}\\\mbox{Stella~Biderman}, \mbox{Leo~Gao}, \mbox{Jonathan~Tow}, \mbox{Sid~Black}, \mbox{Shivanshu~Purohit}, \mbox{Horace~He}, \mbox{Laurence~Golding} \setlength{\parskip}{0.5em}\par
\textbf{Positional Embeddings:}\\\mbox{Stella~Biderman}, \mbox{Laurence~Golding}, \mbox{Michael~Pieler} \par
\textbf{Tokenizer:}\\\mbox{Stella~Biderman}, \mbox{Jason~Phang}, \mbox{Leo~Gao} \par
\end{adjustwidth}

\subsection*{Broader Impacts}
\begin{adjustwidth}{\parindent}{}
\setlength{\parindent}{0pt}
\textbf{Alignment Implications:}\\\mbox{Leo~Gao}, \mbox{Connor~Leahy}, \mbox{Laria~Reynolds}, \mbox{Kyle~McDonell} \setlength{\parskip}{0.5em}\par
\textbf{Environmental Impact:}\\\mbox{Stella~Biderman}, \mbox{Eric~Hallahan} \par
\end{adjustwidth}

\section{Full Configuration Details}\label{app:hparam}

In \Cref{tab:config} we attach the full configuration details used to train \model{}. The file is available in \texttt{.yaml} format usable in \texttt{gpt-neox} at \url{https://github.com/EleutherAI/gpt-neox}, where we also provide documentation describing the role of each parameter.

\begin{table}[ht!]
\small
\centering
\resizebox{\linewidth}{!}{
\begin{tabular}{lrrcc}
\hline
Configuration Key                               & Value \\
\hline
    attention-dropout & 0 \\
    bias-gelu-fusion & True \\
    checkpoint-activations & True \\
    checkpoint-num-layers & 1 \\
    data-impl & mmap \\
    distributed-backend & nccl \\
    eval-interval & 1000 \\
    eval-iters & 10 \\
    fp16.enabled & True \\
    fp16.fp16 & True \\
    fp16.hysteresis & 2 \\
    fp16.initial-scale-power & 12 \\
    fp16.loss-scale & 0 \\
    fp16.loss-scale-window & 1000 \\
    fp16.min-loss-scale & 1 \\
    gpt-j-residual & True \\
    gradient-accumulation-steps & 32 \\
    gradient-clipping & 1.0 \\
    hidden-dropout & 0 \\
    hidden-size & 6144 \\
    init-method & small-init \\
    log-interval & 2 \\
    lr-decay-iters & 150000 \\
    lr-decay-style & cosine \\
    max-position-embeddings & 2048 \\
    min-lr & 9.7e-06 \\
    model-parallel-size & 2 \\
    no-weight-tying & True \\
    norm & layernorm \\
    num-attention-heads & 64 \\
    num-layers & 44 \\
    optimizer.params.betas & [0.9, 0.95] \\
    optimizer.params.eps & 1e-08 \\
    optimizer.params.lr & 9.7e-05 \\
    optimizer.type & Adam \\
    output-layer-init-method & wang-init \\
    output-layer-parallelism & column \\
    partition-activations & False \\
    pipe-parallel-size & 4 \\
    pos-emb & rotary \\
    rotary-pct & 0.25 \\
    save-interval & 500 \\
    scaled-upper-triang-masked-softmax-fusion & True \\
    seq-length & 2048 \\
    split & 995,4,1 \\
    steps-per-print & 2 \\
    synchronize-each-layer & True \\
    tokenizer-type & HFTokenizer \\
    train-iters & 150000 \\
    train-micro-batch-size-per-gpu & 4 \\
    vocab-file & 20B-tokenizer.json \\
    wall-clock-breakdown & False \\
    warmup & 0.01 \\
    weight-decay & 0.01 \\
    zero-optimization.allgather-bucket-size & 1260000000 \\
    zero-optimization.allgather-partitions & True \\
    zero-optimization.contiguous-gradients & True \\
    zero-optimization.cpu-offload & False \\
    zero-optimization.overlap-comm & True \\
    zero-optimization.reduce-bucket-size & 1260000000 \\
    zero-optimization.reduce-scatter & True \\
    zero-optimization.stage & 1 \\
    \end{tabular}
}
\caption{The full configuration details for \model{} training}
\label{tab:config}
\end{table}
\clearpage
\section{Broader Impacts}
\label{sec:broader-impacts}

The current status quo in research is that large language models are things people train and publish about, but do not actually release. To the best of our knowledge, \model{} is the largest dense language model to ever be publicly released with a several-way tie for second place at 13 billion parameters \citep{fairseq-13B,xue2020mt5,xue2021byt5} and many more models at the 10-11B parameter scale. A variety of reasons for the non-release of large language models are given by various groups, but the primary one is the harms that public access to LLMs would purportedly cause.

We take these concerns quite seriously. However, having taken them quite seriously, we feel that they are flawed in several respects. While a thorough analysis of these issues is beyond the scope of this paper, the public release of our model is the most important contribution of this paper and so an explanation of why we disagree with the prevailing wisdom is important.

\paragraph{Providing access to ethics and alignment researchers will prevent harm.} The open-source release of this model is motivated by the hope that it will allow researchers who would not otherwise have access to LLMs to use them. While there are negative risks due to the potential acceleration of capabilities research, we believe the benefits of this release outweigh the risks. We also note that these benefits are not hypothetical, as a number of papers about the limits and ethics of LLMs has been explicitly enabled by the public release of previous models \citep{zhang2021counterfactual,kandpal2022deduplicating,carlini2022quantifying,birhane2021multimodal,logit-lens,meng2022locating, lin2021truthfulqa}.

\paragraph{Limiting access to governments and corporations will not prevent harm.} Perhaps the most curious aspect of the argument that LLMs should not be released is that the people making such arguments are not arguing they \textit{they} should not use LLMs. Rather, they are claiming that \textit{other people} should not use them. We do not believe that this is a position that should be taken seriously. The companies and governments that have the financial resources to train LLMs are overwhelmingly more likely to do large scale harm using a LLM than a random individual.

The open-source release of this model is motivated by the hope that it will allow ethics and alignment researchers who would not otherwise have access to LLMs to use them. While there are negative risks due to the potential acceleration of capabilities research, we believe the benefits of this release outweigh the risks of accelerating capabilities research.

\subsection{Impact on Capabilities Research and Products}

When discussing the impact of access to technology, it is important to distinguish between \textit{capacities research} which seeks to push the current state-of-the-art and research on

We feel the risk of releasing \model{} is acceptable, as the contribution of the model to capabilities research is likely to be limited, for two reasons. 

We ultimately believe that the benefits of releasing this model outweigh the risks, but this argument hinges crucially on the particular circumstances of this release. All actors considering releasing powerful AI models or advancing the frontier of capabilities should think carefully about what they release, in what way, and when.

\subsection{Impact on Ethics and Alignment Research}

To oversimplify a complex debate, there are broadly speaking two schools of thought regarding the mitigation of harm that is done by AI algorithms: \textit{AI Ethics} and \textit{AI Alignement}. AI Ethics researchers are primarily concerned with the impact of current technologies or technologies very similar to current technologies, while AI Alignment is primarily concerned with future ``generally intelligent'' systems whose capacities greatly outclass currently existing systems and possess human and superhuman levels of intelligence. While the tools, methods, and ideas of these camps are very different, we believe that increasing access to these technologies will empower and advance the goals of researchers in both schools.

\subsubsection{The Necessity of Model Access for AI Ethics}

Analyzing and documenting the limitations of models is an essential aspect of AI ethics research \citep{matias2021why}. Work examining and criticizing datasets \citep{kreutzer2021quality,dodge2021documenting,birhane2021multimodal}, functionality \citep{smart2021addressing,zhang2021counterfactual,carlini2022quantifying,biderman2022neural}, evaluation and deployment procedures \citep{biderman2020pitfalls,BS2022Reap}, and more are essential to well-rounded and informed debate on the value and application of technology.

However \textit{the current centralization of LLM training also creates a centralization of control of technology} \citep{sadowski2021everyone,whittaker2021steep} that makes meaningful independent evaluation impossible. This means that it is often not possible to do this kind of work in practice because of the severe access restrictions companies that own large language models put on them. While GPT-NeoX is the 13th largest dense language model at time of writing only model larger than GPT-NeoX 20B that is publicly accessible is GPT-3. There are significant limitations on people's ability to do research on GPT-3 though, as it is not free to use and its training data is private.

\subsubsection{The Usefulness of Large Language Models in Alignment}
\label{subsec:llms-for-alignment}

LLMs represent a different paradigm than the AI systems generally studied by alignment researchers because they are not well-described as coherent agents or expected utility maximizers. Though trained to optimize a log-likelihood loss function, at a high level the goals a LLM pursues are varied and contradictory, depending on the way it is prompted. This introduces additional challenges, but may also enable new approaches to alignment.

\model{} itself is not the system we need to align, but we hope it can serve as a publicly available platform for experiments whose results might generalize to crucial future work.

The following is a non-exhaustive list of potential approaches we consider promising for further investigation.

\paragraph{Mechanistic interpretability.} Mechanistic interpretability research \citep{circuits} hopes to gain an understanding into \textit{how} models accomplish the tasks they do, in part in the hopes of detecting problematic or deceptive algorithms implemented by models before these failures manifest in the real world. Being able to interpret and inspect the detailed inner workings of trained models would be a powerful tool to ensure models are optimizing for the goals we intended \citep{hubinger2021rlo,koch2021objective}. Reverse engineering transformer language models has already yielded insights about the inner functioning of LMs \citep{anthropic-transformer-circuits,logit-lens, meng2022locating, dai2021knowledge}.

\paragraph{Using a LLM as a reward model.} Because they are trained to predict human writing, LLMs also appear to develop a useful representation of human values at the semantic level. Finding a way to utilise these representations could be a possible path toward solving the problem of reward robustness in RL and other algorithms which require a proxy of human judgment \citep{rlhf-summ,alignment-by-default}. Despite fundamental theoretical limitations on learning human values \citep{Armstrong2018OccamsRI,irl-is-hard}, value learning may still be robust enough to align weaker superhuman AIs. Future experiments could explore the extent to which LLM pretraining improves downstream reward model robustness and generalization.

\paragraph{Natural language transparency.} Since LLM prompts are in a human-readable form, it can provide insight on the LLM’s expected behavior. Prompt programming or finetuning can be used to leverage this fact and force a LLM to execute more transparent algorithms, such as splitting problems into steps or explicitly writing an ``internal monologue'' \citep{visible-thoughts, factored-cognition,nye2021work}. Reliability and trustworthiness can present significant challenges for these approaches. 

However, this form of transparency also has its limits. In particular, models can often respond unpredictably to prompts, and internal monologues may become completely detached from the model's decision making process if translating between the model's ontology and the human ontology is more complex than simply modeling human monologues \citep{elk}.

\paragraph{Simulating agents at runtime.} Although LLMs are not well-described as coherent agents, they can still be used to generate goal-directed processes. Given an appropriate prompt (such as a story of a character working to achieve a goal), LLMs can predict and thus simulate an agent \citep{huang2022language}. Simulated agents take representative actions according to the patterns present in the training data, similar to behavior cloning. One potential future research direction is testing whether they are less susceptible to failure modes that follow from expected utility maximization, such as Goodhart failures and power-seeking behavior. However, other failure modes can be introduced by the LM training procedure, such as ``delusions'' or ``hallucinations'' \citep{ortega2021shaking,bc-miscalibrated,maynez2020faithfulness}. Additionally, simulated agents may be uncompetitive with optimal agents like those produced by Reinforcement Learning. An important research direction is to explore how the beneficial properties of simulated agents can be maintained while making them competitive with RL based approaches.

\paragraph{Tool AI and automated alignment research.} LMs can be used as relatively unagentic tools, such as OpenAI's Codex model \citep{chen2021evaluating} acting as a coding assistant. Because pretrained LLMs are not directly optimized for the factual accuracy of their predictions, it is possible they avoid some of the traditional problems with tool or oracle AI \citep{Armstrong2012ThinkingIT}, such as the incentive to produce manipulative answers \citep{predictomatic}. Tool AI is not a long-term solution to the problem of alignment, but it could be used to assist alignment research or even automate large parts of it. For example, language models could be used to help brainstorm alignment ideas more quickly, act as a writing assistant, or directly generate alignment research papers for humans to review. This line of research also risks accelerating capabilities research, a concern we discuss more below.

\subsection{Differential Impact on Access}

Because training large models requires a significant engineering and capital investment, such models are often out of reach for small labs and independent researchers. As it stands, only large organizations have access to the latest generation of powerful language models \citep{brown2020language, rae2021gopher, fedus2021switch, lieber2021jurassic, WuDao}. The number of researchers focused primarily on ethics and alignment working at these labs is much lower than those working on developing new capabilities. 

We feel the risk of releasing \model{} is acceptable, as the contribution of the model to capabilities research is likely to be limited, for two reasons. Firstly, the organizations pursuing capabilities research most aggressively are unlikely to benefit from our open-source release of this model as they have already developed more powerful models of their own. Secondly, we believe the single most important piece of knowledge that drives advancing capabilities research is the knowledge that scaling LLMs was possible in the first place \citep{why-release,leahy2021hard}. Whereas the actual implementation is very fungible (as evidenced by the large number of parties who have succeeded in creating their own LLMs in the past two years). \textbf{This differential impact, wherein our release is expected to benefit primarily people who have less funding and infrastructure, is a key factor in our decision to release this model publicly.}

We ultimately believe that the benefits of releasing this model outweigh the risks, but this argument hinges crucially on the particular circumstances of this release. All actors considering releasing powerful AI models or advancing the frontier of capabilities should think carefully about what they release, in what way, and when. 

\subsection{Environmental Impact}
\label{subsec:environmental-impact}

A significant point of concern in some recent work is the energy usage and carbon emissions associated with training large language models \citep{strubell2019energy,schwartz2020green,lacoste2019quantifying,bender2021dangers}. In particular, \citet{strubell2019energy} estimate that a then-recent paper by the authors released $626,155$ lbs or $284.01$ metric tons\footnote{We choose to present environmental impact figures in metric tons to align with standard reporting.} of $\mathrm{CO_2}$ ($\mathrm{t_{CO_2}}$). As \citet{strubell2019energy} has been widely cited and quoted in the media as representative of large-scale language models, we decided to explicitly and carefully track our energy usage and carbon emissions to see if this is truly a representative account of NLP emissions.

Throughout the development and training of our model, we tracked our energy usage and carbon emissions. We found that the process of developing and training \model{} emitted almost exactly $10\%$ of \citet{strubell2019energy}'s estimate, coming in at a total of $69957$ lbs or $31.73$ metric tons of $\mathrm{CO_2}$. This is roughly the equivalent of the yearly emissions of the average American or 35 round-trip flights between New York City and San Francisco. Our systems were based in Illinois, USA, and consumed energy sourced from the mix as follows

\begin{itemize}
    \item $30.40\%$ Coal ($0.95\,\mathrm{ t_{CO_2}}$/MWh)
    \item $31.30\%$ Gas ($0.6078\,\mathrm{ t_{CO_2}}$/MWh)
    \item $\phantom{0}1.30\%$ Hydroelectric ($0\,\mathrm{ t_{CO_2}}$/MWh)
    \item $17.40\%$ Nuclear ($0\,\mathrm{ t_{CO_2}}$/MWh)
    \item $\phantom{0}0.30\%$ Solar ($0\,\mathrm{ t_{CO_2}}$/MWh)
    \item $18.10\%$ Wind ($0\,\mathrm{ t_{CO_2}}$/MWh)
    \item $\phantom{0}1.30\%$ Other Renewables ($0\,\mathrm{t_{CO_2}}$/MWh)
\end{itemize}

This mixture produces an average of $0.47905$ $\mathrm{t_{CO_2}}$/MWh, and we consumed a total of $43.92$\,MWh of electricity over the course of $1830$ hours of training. Scaling, testing, and evaluation were responsible for the equivalent of another $920$ hours on our systems, for a total energy consumption $66.24$\,MWh and thus the production of just under $35$ metric tons of $\mathrm{CO_2}$.

It is noteworthy that \citet{strubell2019energy} are estimating emissions from a \textit{neural architecture search} paper, and is therefore not directly comparable to ours. The primary motivation for our comparison is that their number has attracted a lot of attention and is often taken to be respresentative of NLP research. In general, we advocate for more systematic and comprehensive reporting to improve transparency surrounding this important topic.

\section{Full Evaluation Results}\label{app:downstream}

Results for natural language understanding tasks are shown in \Cref{tab:nlu_gpt,tab:nlu_fairseq}, while results for Hendrycks tasks are found in \Cref{tab:hendrycks_gpt1,tab:hendrycks_gpt2,tab:hendrycks_fairseq1,tab:hendrycks_fairseq2}.

All evaluations had version 0 in the Evaluation Harness. This information is reported in the output of the Evaluation Harness and should be used for ensuring reproducibility of these results, even as the task implementations themselves may change to fix bugs.

\clearpage


{\onecolumn
\begin{landscape}
\begin{table*}    
\centering 
\begin{tabular}{l c c c c c c }
 & GPT-J & GPT-NeoX & \multicolumn{4}{c}{GPT-3} \\
Task & 6B & 20B & Ada & Babbage & Curie & DaVinci\\ \toprule
ANLI Round 1 & $0.324 \pm 0.015$ & $0.340 \pm 0.015$ & $0.334 \pm 0.015$ & $0.326 \pm 0.015$ & $0.325 \pm 0.015$ & $0.363 \pm 0.015$ \\ 
ANLI Round 2 & $0.340 \pm 0.015$ & $0.343 \pm 0.015$ & $0.342 \pm 0.015$ & $0.308 \pm 0.015$ & $0.338 \pm 0.015$ & $0.375 \pm 0.015$ \\ 
ANLI Round 3 & $0.355 \pm 0.014$ & $0.354 \pm 0.014$ & $0.354 \pm 0.014$ & $0.340 \pm 0.014$ & $0.353 \pm 0.014$ & $0.369 \pm 0.014$ \\ 
LAMBADA & $0.683 \pm 0.006$ & $0.720 \pm 0.006$ & $0.515 \pm 0.007$ & $0.625 \pm 0.007$ & $0.693 \pm 0.006$ & $0.752 \pm 0.006$ \\ 
WSC & $0.365 \pm 0.047$ & $0.500 \pm 0.049$ & $0.375 \pm 0.048$ & $0.404 \pm 0.048$ & $0.548 \pm 0.049$ & $0.548 \pm 0.049$ \\ 
HellaSwag & $0.518 \pm 0.005$ & $0.535 \pm 0.005$ & $0.359 \pm 0.005$ & $0.429 \pm 0.005$ & $0.505 \pm 0.005$ & $0.592 \pm 0.005$ \\ 
Winogrande & $0.640 \pm 0.013$ & $0.661 \pm 0.013$ & $0.528 \pm 0.014$ & $0.594 \pm 0.014$ & $0.649 \pm 0.013$ & $0.699 \pm 0.013$ \\ 
SciQ & $0.910 \pm 0.009$ & $0.928 \pm 0.008$ & $0.843 \pm 0.012$ & $0.866 \pm 0.011$ & $0.918 \pm 0.009$ & $0.949 \pm 0.007$ \\ 
PIQA & $0.752 \pm 0.010$ & $0.779 \pm 0.010$ & $0.690 \pm 0.011$ & $0.745 \pm 0.010$ & $0.767 \pm 0.010$ & $0.791 \pm 0.009$ \\ 
TriviaQA & $0.170 \pm 0.004$ & $0.259 \pm 0.004$ & $0.050 \pm 0.002$ & $0.115 \pm 0.003$ & $0.196 \pm 0.004$ & $0.409 \pm 0.005$ \\ 
ARC (Easy) & $0.670 \pm 0.010$ & $0.723 \pm 0.009$ & $0.514 \pm 0.010$ & $0.598 \pm 0.010$ & $0.682 \pm 0.010$ & $0.762 \pm 0.009$ \\ 
ARC (Challenge) & $0.340 \pm 0.014$ & $0.380 \pm 0.014$ & $0.225 \pm 0.012$ & $0.275 \pm 0.013$ & $0.334 \pm 0.014$ & $0.435 \pm 0.014$ \\ 
OpenBookQA & $0.288 \pm 0.020$ & $0.290 \pm 0.020$ & $0.172 \pm 0.017$ & $0.224 \pm 0.019$ & $0.290 \pm 0.020$ & $0.336 \pm 0.021$ \\ 
HeadQA (English) & --- & --- & $0.245 \pm 0.008$ & $0.278 \pm 0.009$ & $0.317 \pm 0.009$ & $0.356 \pm 0.009$ \\ 
LogiQA & $0.209 \pm 0.016$ & $0.230 \pm 0.017$ & $0.218 \pm 0.016$ & $0.198 \pm 0.016$ & $0.217 \pm 0.016$ & $0.227 \pm 0.016$ \\ 
PROST & $0.267 \pm 0.003$ & $0.296 \pm 0.003$ & $0.254 \pm 0.003$ & $0.270 \pm 0.003$ & $0.288 \pm 0.003$ & $0.267 \pm 0.003$ \\ 
QA4MRE (2013) & $0.373 \pm 0.029$ & $0.363 \pm 0.029$ & $0.320 \pm 0.028$ & $0.370 \pm 0.029$ & $0.377 \pm 0.029$ & $0.426 \pm 0.029$ \\ \bottomrule
\end{tabular}
\caption{Zero-Shot Results on Natural Language Understanding Tasks (GPT-J, GPT-NeoX and GPT-3)}
\label{tab:nlu_gpt}
\end{table*}

\begin{table*}    
\centering 
\begin{tabular}{l c c c c c c }
 & \multicolumn{6}{c}{FairSeq} \\
 Task & 125M & 355M & 1.3B & 2.7B & 6.7B & 13B\\ \toprule
ANLI Round 1 & $0.316 \pm 0.015$ & $0.322 \pm 0.015$ & $0.331 \pm 0.015$ & $0.318 \pm 0.015$ & $0.338 \pm 0.015$ & $0.340 \pm 0.015$ \\ 
ANLI Round 2 & $0.336 \pm 0.015$ & $0.312 \pm 0.015$ & $0.334 \pm 0.015$ & $0.339 \pm 0.015$ & $0.322 \pm 0.015$ & $0.330 \pm 0.015$ \\ 
ANLI Round 3 & $0.330 \pm 0.014$ & $0.323 \pm 0.014$ & $0.333 \pm 0.014$ & $0.340 \pm 0.014$ & $0.333 \pm 0.014$ & $0.347 \pm 0.014$ \\ 
LAMBADA & $0.388 \pm 0.007$ & $0.478 \pm 0.007$ & $0.562 \pm 0.007$ & $0.632 \pm 0.007$ & $0.673 \pm 0.007$ & $0.709 \pm 0.006$ \\ 
WSC & $0.365 \pm 0.047$ & $0.471 \pm 0.049$ & $0.365 \pm 0.047$ & $0.635 \pm 0.047$ & $0.615 \pm 0.048$ & $0.577 \pm 0.049$ \\ 
HellaSwag & $0.309 \pm 0.005$ & $0.380 \pm 0.005$ & $0.448 \pm 0.005$ & $0.493 \pm 0.005$ & $0.525 \pm 0.005$ & $0.554 \pm 0.005$ \\ 
Winogrande & $0.513 \pm 0.014$ & $0.529 \pm 0.014$ & $0.600 \pm 0.014$ & $0.620 \pm 0.014$ & $0.644 \pm 0.013$ & $0.674 \pm 0.013$ \\ 
SciQ & $0.732 \pm 0.014$ & $0.737 \pm 0.014$ & $0.838 \pm 0.012$ & $0.878 \pm 0.010$ & $0.895 \pm 0.010$ & $0.910 \pm 0.009$ \\ 
PIQA & $0.668 \pm 0.011$ & $0.690 \pm 0.011$ & $0.731 \pm 0.010$ & $0.751 \pm 0.010$ & $0.762 \pm 0.010$ & $0.769 \pm 0.010$ \\ 
TriviaQA & $0.015 \pm 0.001$ & $0.019 \pm 0.001$ & $0.078 \pm 0.003$ & $0.141 \pm 0.003$ & $0.221 \pm 0.004$ & $0.270 \pm 0.004$ \\ 
ARC (Easy) & $0.426 \pm 0.010$ & $0.468 \pm 0.010$ & $0.565 \pm 0.010$ & $0.625 \pm 0.010$ & $0.665 \pm 0.010$ & $0.680 \pm 0.010$ \\ 
ARC (Challenge) & $0.195 \pm 0.012$ & $0.233 \pm 0.012$ & $0.263 \pm 0.013$ & $0.296 \pm 0.013$ & $0.329 \pm 0.014$ & $0.345 \pm 0.014$ \\ 
OpenBookQA & $0.168 \pm 0.017$ & $0.190 \pm 0.018$ & $0.238 \pm 0.019$ & $0.254 \pm 0.019$ & $0.292 \pm 0.020$ & $0.296 \pm 0.020$ \\ 
HeadQA (English) & $0.233 \pm 0.008$ & $0.233 \pm 0.008$ & $0.256 \pm 0.008$ & $0.264 \pm 0.008$ & $0.280 \pm 0.009$ & $0.280 \pm 0.009$ \\ 
LogiQA & $0.220 \pm 0.016$ & $0.230 \pm 0.017$ & $0.214 \pm 0.016$ & $0.212 \pm 0.016$ & $0.232 \pm 0.017$ & $0.240 \pm 0.017$ \\ 
PROST & $0.215 \pm 0.003$ & $0.257 \pm 0.003$ & $0.257 \pm 0.003$ & $0.230 \pm 0.003$ & $0.272 \pm 0.003$ & $0.252 \pm 0.003$ \\ 
QA4MRE (2013) & $0.285 \pm 0.027$ & $0.335 \pm 0.028$ & $0.327 \pm 0.028$ & $0.380 \pm 0.029$ & $0.370 \pm 0.029$ & $0.380 \pm 0.029$ \\\bottomrule
\end{tabular}
\caption{Zero-Shot Results on Natural Language Understanding Tasks (FairSeq Models)}
\label{tab:nlu_fairseq}
\end{table*}


\begin{table*}
\centering 
\begin{tabular}{l c c c c c c } \\
 & GPT-J & GPT-NeoX & \multicolumn{4}{c}{GPT-3} \\
Task & 6B & 20B & Ada & Babbage & Curie & DaVinci\\ \toprule
ANLI Round 1 & $0.322 \pm 0.015$ & $0.312 \pm 0.015$ & ---  & --- & --- & --- \\ 
ANLI Round 2 & $0.331 \pm 0.015$ & $0.329 \pm 0.015$ & ---  & --- & --- & --- \\ 
ANLI Round 3 & $0.346 \pm 0.014$ & $0.342 \pm 0.014$ & ---  & --- & --- & --- \\ 
LAMBADA & $0.662 \pm 0.007$ & $0.698 \pm 0.006$ & ---  & --- & --- & --- \\ 
WSC & $0.365 \pm 0.047$ & $0.385 \pm 0.048$ & ---  & --- & --- & --- \\ 
HellaSwag & $0.494 \pm 0.005$ & $0.538 \pm 0.005$ & ---  & --- & --- & --- \\ 
Winogrande & $0.660 \pm 0.013$ & $0.683 \pm 0.013$ & ---  & --- & --- & --- \\ 
SciQ & $0.913 \pm 0.009$ & $0.960 \pm 0.006$ & ---  & --- & --- & --- \\ 
PIQA & $0.756 \pm 0.010$ & $0.774 \pm 0.010$ & ---  & --- & --- & --- \\ 
TriviaQA & $0.289 \pm 0.004$ & $0.347 \pm 0.004$ & ---  & --- & --- & --- \\ 
ARC (Challenge) & $0.360 \pm 0.014$ & $0.410 \pm 0.014$ & ---  & --- & --- & --- \\ 
ARC (Easy) & $0.705 \pm 0.009$ & $0.746 \pm 0.009$ & ---  & --- & --- & --- \\ 
OpenBookQA & $0.310 \pm 0.021$ & $0.326 \pm 0.021$ & ---  & --- & --- & --- \\ 
HeadQA (English) & $0.326 \pm 0.009$ & $0.385 \pm 0.009$ & ---  & --- & --- & --- \\ 
LogiQA & $0.230 \pm 0.017$ & $0.220 \pm 0.016$ & ---  & --- & --- & --- \\ 
QA4MRE (2013) & $0.366 \pm 0.029$ & $0.363 \pm 0.029$ & ---  & --- & --- & --- \\ 
\bottomrule 
\end{tabular}
\caption{Five-Shot Results on Natural Language Understanding Tasks (GPT-J and GPT-NeoX). GPT-3 is omitted due to financial limitations.}
\label{tab:nlu_gpt_5}
\end{table*}

\begin{table*}
\centering 
\begin{tabular}{l c c c c c c } \\ 
& \multicolumn{4}{c}{FairSeq} \\ 
Task & 125M & 355M & 1.3B & 2.7B & 6.7B & 13B \\ \toprule
ANLI Round 1 & $0.332 \pm 0.015$ & $0.336 \pm 0.015$ & $0.327 \pm 0.015$ & $0.336 \pm 0.015$ & $0.305 \pm 0.015$ & $0.335 \pm 0.015$ \\ 
ANLI Round 2 & $0.345 \pm 0.015$ & $0.350 \pm 0.015$ & $0.347 \pm 0.015$ & $0.333 \pm 0.015$ & $0.340 \pm 0.015$ & $0.338 \pm 0.015$ \\ 
ANLI Round 3 & $0.359 \pm 0.014$ & $0.347 \pm 0.014$ & $0.370 \pm 0.014$ & $0.326 \pm 0.014$ & $0.367 \pm 0.014$ & $0.357 \pm 0.014$ \\ 
LAMBADA & $0.268 \pm 0.006$ & $0.349 \pm 0.007$ & $0.427 \pm 0.007$ & $0.460 \pm 0.007$ & $0.494 \pm 0.007$ & $0.518 \pm 0.007$ \\ 
WSC & $0.365 \pm 0.047$ & $0.365 \pm 0.047$ & $0.365 \pm 0.047$ & $0.356 \pm 0.047$ & $0.500 \pm 0.049$ & $0.404 \pm 0.048$ \\ 
HellaSwag & $0.308 \pm 0.005$ & $0.379 \pm 0.005$ & $0.451 \pm 0.005$ & $0.497 \pm 0.005$ & $0.531 \pm 0.005$ & $0.559 \pm 0.005$ \\ 
Winogrande & $0.516 \pm 0.014$ & $0.538 \pm 0.014$ & $0.612 \pm 0.014$ & $0.633 \pm 0.014$ & $0.657 \pm 0.013$ & $0.690 \pm 0.013$ \\ 
SciQ & $0.758 \pm 0.014$ & $0.819 \pm 0.012$ & $0.859 \pm 0.011$ & $0.875 \pm 0.010$ & $0.871 \pm 0.011$ & $0.899 \pm 0.010$ \\ 
PIQA & $0.656 \pm 0.011$ & $0.700 \pm 0.011$ & $0.731 \pm 0.010$ & $0.750 \pm 0.010$ & $0.764 \pm 0.010$ & $0.769 \pm 0.010$ \\ 
TriviaQA & $0.044 \pm 0.002$ & $0.097 \pm 0.003$ & $0.160 \pm 0.003$ & $0.225 \pm 0.004$ & $0.293 \pm 0.004$ & $0.323 \pm 0.004$ \\ 
ARC (Easy) & $0.453 \pm 0.010$ & $0.533 \pm 0.010$ & $0.618 \pm 0.010$ & $0.664 \pm 0.010$ & $0.686 \pm 0.010$ & $0.702 \pm 0.009$ \\ 
ARC (Challenge) & $0.198 \pm 0.012$ & $0.231 \pm 0.012$ & $0.278 \pm 0.013$ & $0.310 \pm 0.014$ & $0.359 \pm 0.014$ & $0.370 \pm 0.014$ \\ 
OpenBookQA & $0.184 \pm 0.017$ & $0.206 \pm 0.018$ & $0.218 \pm 0.018$ & $0.258 \pm 0.020$ & $0.288 \pm 0.020$ & $0.290 \pm 0.020$ \\ 
HeadQA (English) & $0.235 \pm 0.008$ & $0.240 \pm 0.008$ & $0.254 \pm 0.008$ & $0.266 \pm 0.008$ & $0.276 \pm 0.009$ & $0.282 \pm 0.009$ \\ 
LogiQA & $0.218 \pm 0.016$ & $0.207 \pm 0.016$ & $0.210 \pm 0.016$ & $0.214 \pm 0.016$ & $0.214 \pm 0.016$ & $0.223 \pm 0.016$ \\ 
QA4MRE (2013) & $0.324 \pm 0.028$ & $0.338 \pm 0.028$ & $0.338 \pm 0.028$ & $0.352 \pm 0.028$ & $0.391 \pm 0.029$ & $0.387 \pm 0.029$ \\ 
\bottomrule 
\end{tabular}
\caption{Five-Shot Results on Natural Language Understanding Tasks (FairSeq Models)}
\label{tab:nlu_fairseq_5}
\end{table*}


\begin{table*}
\centering 
\begin{tabular}{l c c c c c c } \\ 
& GPT-J & GPT-NeoX & \multicolumn{4}{c}{GPT-3} \\
Task & 6B & 20B & Ada & Babbage & Curie & DaVinci \\ \toprule
1DC & $0.088 \pm 0.006$ & $0.098 \pm 0.007$ & $0.029 \pm 0.000$ & $0.001 \pm 0.000$ & $0.024 \pm 0.000$ & $0.098 \pm 0.000$ \\ 
2D+ & $0.238 \pm 0.010$ & $0.570 \pm 0.011$ & $0.006 \pm 0.000$ & $0.009 \pm 0.000$ & $0.025 \pm 0.000$ & $0.769 \pm 0.000$ \\ 
2Dx & $0.139 \pm 0.008$ & $0.148 \pm 0.008$ & $0.022 \pm 0.000$ & $0.021 \pm 0.000$ & $0.058 \pm 0.000$ & $0.198 \pm 0.000$ \\ 
2D- & $0.216 \pm 0.009$ & $0.680 \pm 0.010$ & $0.013 \pm 0.000$ & $0.013 \pm 0.000$ & $0.076 \pm 0.000$ & $0.580 \pm 0.000$ \\ 
3D+ & $0.088 \pm 0.006$ & $0.099 \pm 0.007$ & $0.001 \pm 0.000$ & $0.001 \pm 0.000$ & $0.003 \pm 0.000$ & $0.342 \pm 0.000$ \\ 
3D- & $0.046 \pm 0.005$ & $0.344 \pm 0.011$ & $0.001 \pm 0.000$ & $0.001 \pm 0.000$ & $0.004 \pm 0.000$ & $0.483 \pm 0.000$ \\ 
4D+ & $0.007 \pm 0.002$ & $0.007 \pm 0.002$ & $0.001 \pm 0.000$ & $0.000 \pm 0.000$ & $0.001 \pm 0.000$ & $0.040 \pm 0.000$ \\ 
4D- & $0.005 \pm 0.002$ & $0.029 \pm 0.004$ & $0.000 \pm 0.000$ & $0.000 \pm 0.000$ & $0.000 \pm 0.000$ & $0.075 \pm 0.000$ \\ 
5D+ & $0.001 \pm 0.001$ & $0.000 \pm 0.000$ & $0.000 \pm 0.000$ & $0.000 \pm 0.000$ & $0.000 \pm 0.000$ & $0.006 \pm 0.000$ \\ 
5D- & $0.000 \pm 0.000$ & $0.004 \pm 0.001$ & $0.000 \pm 0.000$ & $0.000 \pm 0.000$ & $0.000 \pm 0.000$ & $0.008 \pm 0.000$ \\ 
MATH (Algebra) & $0.013 \pm 0.003$ & $0.010 \pm 0.003$ & $0.003 \pm 0.002$ & $0.008 \pm 0.003$ & $0.003 \pm 0.002$ & $0.008 \pm 0.003$ \\ 
MATH (Counting and Probability) & $0.011 \pm 0.005$ & $0.017 \pm 0.006$ & $0.000 \pm 0.000$ & $0.004 \pm 0.003$ & $0.000 \pm 0.000$ & $0.006 \pm 0.004$ \\ 
MATH (Geometry) & $0.004 \pm 0.003$ & $0.017 \pm 0.006$ & $0.000 \pm 0.000$ & $0.000 \pm 0.000$ & $0.002 \pm 0.002$ & $0.002 \pm 0.002$ \\ 
MATH (Intermediate Algebra) & $0.004 \pm 0.002$ & $0.001 \pm 0.001$ & $0.000 \pm 0.000$ & $0.003 \pm 0.002$ & $0.006 \pm 0.002$ & $0.003 \pm 0.002$ \\ 
MATH (Number Theory) & $0.007 \pm 0.004$ & $0.013 \pm 0.005$ & $0.007 \pm 0.004$ & $0.000 \pm 0.000$ & $0.006 \pm 0.003$ & $0.011 \pm 0.005$ \\ 
MATH (Pre-Algebra) & $0.010 \pm 0.003$ & $0.018 \pm 0.005$ & $0.007 \pm 0.003$ & $0.006 \pm 0.003$ & $0.008 \pm 0.003$ & $0.014 \pm 0.004$ \\ 
MATH (Pre-Calculus) & $0.005 \pm 0.003$ & $0.005 \pm 0.003$ & $0.004 \pm 0.003$ & $0.000 \pm 0.000$ & $0.002 \pm 0.002$ & $0.004 \pm 0.003$ \\ 

\bottomrule 
\end{tabular}
\caption{Zero-Shot Results on Basic Arithmetic and MATH (GPT-J, GPT-NeoX, and GPT-3)}\label{tab:math_gpt}
\end{table*}

\begin{table*}
\centering 
\begin{tabular}{l c c c c c c } \\ 
& \multicolumn{4}{c}{FairSeq} \\ 
Task & 125M & 355M & 1.3B & 2.7B & 6.7B & 13B \\ \toprule
1DC & $0.001 \pm 0.001$ & $0.000 \pm 0.000$ & $0.000 \pm 0.000$ & $0.011 \pm 0.002$ & $0.024 \pm 0.003$ & $0.001 \pm 0.001$ \\ 
2D+ & $0.005 \pm 0.002$ & $0.001 \pm 0.001$ & $0.002 \pm 0.001$ & $0.009 \pm 0.002$ & $0.019 \pm 0.003$ & $0.020 \pm 0.003$ \\ 
2Dx & $0.020 \pm 0.003$ & $0.004 \pm 0.001$ & $0.018 \pm 0.003$ & $0.023 \pm 0.003$ & $0.036 \pm 0.004$ & $0.028 \pm 0.004$ \\ 
2D- & $0.005 \pm 0.002$ & $0.002 \pm 0.001$ & $0.006 \pm 0.002$ & $0.013 \pm 0.002$ & $0.013 \pm 0.003$ & $0.015 \pm 0.003$ \\ 
3D+ & $0.001 \pm 0.001$ & $0.001 \pm 0.001$ & $0.001 \pm 0.001$ & $0.001 \pm 0.001$ & $0.001 \pm 0.001$ & $0.001 \pm 0.001$ \\ 
3D- & $0.002 \pm 0.001$ & $0.001 \pm 0.001$ & $0.002 \pm 0.001$ & $0.002 \pm 0.001$ & $0.002 \pm 0.001$ & $0.002 \pm 0.001$ \\ 
4D+ & $0.001 \pm 0.001$ & $0.000 \pm 0.000$ & $0.001 \pm 0.001$ & $0.001 \pm 0.001$ & $0.001 \pm 0.001$ & $0.001 \pm 0.001$ \\ 
4D- & $0.000 \pm 0.000$ & $0.000 \pm 0.000$ & $0.000 \pm 0.000$ & $0.000 \pm 0.000$ & $0.000 \pm 0.000$ & $0.000 \pm 0.000$ \\ 
5D+ & $0.000 \pm 0.000$ & $0.000 \pm 0.000$ & $0.000 \pm 0.000$ & $0.000 \pm 0.000$ & $0.000 \pm 0.000$ & $0.000 \pm 0.000$ \\ 
5D- & $0.000 \pm 0.000$ & $0.000 \pm 0.000$ & $0.000 \pm 0.000$ & $0.000 \pm 0.000$ & $0.000 \pm 0.000$ & $0.000 \pm 0.000$ \\ 
MATH (Algebra) & $0.000 \pm 0.000$ & $0.000 \pm 0.000$ & $0.001 \pm 0.001$ & $0.003 \pm 0.002$ & $0.004 \pm 0.002$ & $0.003 \pm 0.001$ \\ 
MATH (Counting and Probability) & $0.000 \pm 0.000$ & $0.000 \pm 0.000$ & $0.000 \pm 0.000$ & $0.000 \pm 0.000$ & $0.004 \pm 0.003$ & $0.000 \pm 0.000$ \\ 
MATH (Geometry) & $0.000 \pm 0.000$ & $0.000 \pm 0.000$ & $0.000 \pm 0.000$ & $0.002 \pm 0.002$ & $0.000 \pm 0.000$ & $0.000 \pm 0.000$ \\ 
MATH (Intermediate Algebra) & $0.000 \pm 0.002$ & $0.000 \pm 0.002$ & $0.000 \pm 0.000$ & $0.001 \pm 0.001$ & $0.006 \pm 0.002$ & $0.002 \pm 0.002$ \\ 
MATH (Number Theory) & $0.000 \pm 0.000$ & $0.000 \pm 0.000$ & $0.000 \pm 0.000$ & $0.002 \pm 0.002$ & $0.000 \pm 0.000$ & $0.004 \pm 0.003$ \\ 
MATH (Pre-Algebra) & $0.000 \pm 0.000$ & $0.000 \pm 0.000$ & $0.003 \pm 0.002$ & $0.002 \pm 0.002$ & $0.001 \pm 0.001$ & $0.000 \pm 0.000$ \\ 
MATH (Pre-Calculus) & $0.000 \pm 0.000$ & $0.000 \pm 0.000$ & $0.000 \pm 0.000$ & $0.002 \pm 0.002$ & $0.000 \pm 0.000$ & $0.000 \pm 0.000$ \\ 
\bottomrule 
\end{tabular}
\caption{Zero-Shot Results on Basic Arithmetic and MATH (FairSeq Models)}
\label{tab:math_fairseq}
\end{table*}


\begin{table*}
\centering 
\begin{tabular}{l c c c c c c } \\ 
& GPT-J & GPT-NeoX & \multicolumn{4}{c}{GPT-3} \\
Task & 6B & 20B & Ada & Babbage & Curie & DaVinci \\ \toprule
1DC & $0.192 \pm 0.009$ & $0.191 \pm 0.009$ & --- & --- & --- & ---  \\ 
2D+ & $0.880 \pm 0.007$ & $0.992 \pm 0.002$ & --- & --- & --- & --- \\ 
2Dx & $0.282 \pm 0.010$ & $0.452 \pm 0.011$ & --- & --- & --- & --- \\ 
2D- & $0.817 \pm 0.009$ & $0.942 \pm 0.005$ & --- & --- & --- & --- \\ 
3D+ & $0.357 \pm 0.011$ & $0.599 \pm 0.011$ & --- & --- & --- & --- \\ 
3D- & $0.497 \pm 0.011$ & $0.819 \pm 0.009$ & --- & --- & --- & --- \\ 
4D+ & $0.058 \pm 0.005$ & $0.152 \pm 0.008$ & --- & --- & --- & --- \\ 
4D- & $0.092 \pm 0.006$ & $0.151 \pm 0.008$ & --- & --- & --- & --- \\ 
5D+ & $0.009 \pm 0.002$ & $0.033 \pm 0.004$ & --- & --- & --- & --- \\ 
5D- & $0.021 \pm 0.003$ & $0.059 \pm 0.005$ & --- & --- & --- & --- \\ 
MATH (Algebra) & $0.032 \pm 0.005$ & $0.049 \pm 0.006$ & --- & --- & --- & --- \\ 
MATH (Counting and Probability) & $0.036 \pm 0.009$ & $0.030 \pm 0.008$& --- & --- & --- & ---  \\ 
MATH (Geometry) & $0.027 \pm 0.007$ & $0.015 \pm 0.005$ & --- & --- & --- & --- \\ 
MATH (Intermediate Algebra) & $0.024 \pm 0.005$ & $0.021 \pm 0.005$ & --- & --- & --- & --- \\ 
MATH (Number Theory) & $0.044 \pm 0.009$ & $0.065 \pm 0.011$ & --- & --- & --- & --- \\ 
MATH (Pre-Algebra) & $0.052 \pm 0.008$ & $0.057 \pm 0.008$ & --- & --- & --- & --- \\ 
MATH (Pre-Calculus) & $0.013 \pm 0.005$ & $0.027 \pm 0.007$ & --- & --- & --- & --- \\ 
\bottomrule 
\end{tabular}
\caption{Five-Shot Results on Basic Arithmetic and MATH (GPT-J and GPT-NeoX). GPT-3 is omitted due to financial limitations.}
\label{tab:math_gpt_5}
\end{table*}

\begin{table*}
\centering 
\begin{tabular}{l c c c c c c } \\ 
& \multicolumn{4}{c}{FairSeq} \\ 
Task & 125M & 355M & 1.3B & 2.7B & 6.7B & 13B \\ \toprule
1DC & $0.019 \pm 0.003$ & $0.024 \pm 0.003$ & $0.029 \pm 0.004$ & $0.032 \pm 0.004$ & $0.046 \pm 0.005$ & $0.046 \pm 0.005$ \\ 
2D+ & $0.005 \pm 0.002$ & $0.004 \pm 0.001$ & $0.006 \pm 0.002$ & $0.029 \pm 0.004$ & $0.034 \pm 0.004$ & $0.051 \pm 0.005$ \\ 
2Dx & $0.001 \pm 0.001$ & $0.025 \pm 0.004$ & $0.025 \pm 0.003$ & $0.025 \pm 0.003$ & $0.049 \pm 0.005$ & $0.053 \pm 0.005$ \\ 
2D- & $0.007 \pm 0.002$ & $0.011 \pm 0.002$ & $0.008 \pm 0.002$ & $0.013 \pm 0.003$ & $0.018 \pm 0.003$ & $0.030 \pm 0.004$ \\ 
3D+ & $0.002 \pm 0.001$ & $0.002 \pm 0.001$ & $0.001 \pm 0.001$ & $0.003 \pm 0.001$ & $0.001 \pm 0.001$ & $0.003 \pm 0.001$ \\ 
3D- & $0.002 \pm 0.001$ & $0.004 \pm 0.001$ & $0.003 \pm 0.001$ & $0.003 \pm 0.001$ & $0.002 \pm 0.001$ & $0.003 \pm 0.001$ \\ 
4D+ & $0.000 \pm 0.000$ & $0.000 \pm 0.000$ & $0.000 \pm 0.000$ & $0.000 \pm 0.000$ & $0.000 \pm 0.000$ & $0.000 \pm 0.000$ \\ 
4D- & $0.001 \pm 0.001$ & $0.000 \pm 0.000$ & $0.000 \pm 0.000$ & $0.001 \pm 0.001$ & $0.000 \pm 0.000$ & $0.000 \pm 0.000$ \\ 
5D+ & $0.000 \pm 0.000$ & $0.000 \pm 0.000$ & $0.000 \pm 0.000$ & $0.000 \pm 0.000$ & $0.000 \pm 0.000$ & $0.000 \pm 0.000$ \\ 
5D- & $0.000 \pm 0.000$ & $0.000 \pm 0.000$ & $0.000 \pm 0.000$ & $0.000 \pm 0.000$ & $0.000 \pm 0.000$ & $0.000 \pm 0.000$ \\ 
MATH (Algebra) & $0.023 \pm 0.004$ & $0.010 \pm 0.003$ & $0.013 \pm 0.003$ & $0.014 \pm 0.003$ & $0.017 \pm 0.004$ & $0.012 \pm 0.003$ \\ 
MATH (Counting and Probability) & $0.008 \pm 0.004$ & $0.004 \pm 0.003$ & $0.015 \pm 0.006$ & $0.017 \pm 0.006$ & $0.015 \pm 0.006$ & $0.017 \pm 0.006$ \\ 
MATH (Geometry) & $0.000 \pm 0.000$ & $0.013 \pm 0.005$ & $0.006 \pm 0.004$ & $0.015 \pm 0.005$ & $0.015 \pm 0.005$ & $0.006 \pm 0.004$ \\ 
MATH (Intermediate Algebra) & $0.010 \pm 0.003$ & $0.002 \pm 0.002$ & $0.007 \pm 0.003$ & $0.010 \pm 0.003$ & $0.011 \pm 0.003$ & $0.004 \pm 0.002$ \\ 
MATH (Number Theory) & $0.019 \pm 0.006$ & $0.009 \pm 0.004$ & $0.007 \pm 0.004$ & $0.011 \pm 0.005$ & $0.028 \pm 0.007$ & $0.019 \pm 0.006$ \\ 
MATH (Pre-Algebra) & $0.013 \pm 0.004$ & $0.008 \pm 0.003$ & $0.010 \pm 0.003$ & $0.011 \pm 0.004$ & $0.021 \pm 0.005$ & $0.013 \pm 0.004$ \\ 
MATH (Pre-Calculus) & $0.002 \pm 0.002$ & $0.002 \pm 0.002$ & $0.004 \pm 0.003$ & $0.000 \pm 0.000$ & $0.002 \pm 0.002$ & $0.000 \pm 0.000$ \\ 
\bottomrule 
\end{tabular}
\caption{Five-Shot Results on Basic Arithmetic and MATH (FairSeq Models)}
\label{tab:math_fairseq_5}
\end{table*}
\end{landscape}}


{\onecolumn
\begin{landscape}
\begin{table*}    
\centering 
\begin{tabular}{l c c c c c c }
 & GPT-J & GPT-NeoX & \multicolumn{4}{c}{GPT-3} \\
Task & 6B & 20B & Ada & Babbage & Curie & DaVinci\\ \toprule
Abstract Algebra & $0.260 \pm 0.044$ & $0.230 \pm 0.042$ & $0.170 \pm 0.038$ & $0.220 \pm 0.042$ & $0.220 \pm 0.042$ & $0.220 \pm 0.042$ \\ 
Anatomy & $0.274 \pm 0.039$ & $0.319 \pm 0.040$ & $0.207 \pm 0.035$ & $0.289 \pm 0.039$ & $0.274 \pm 0.039$ & $0.348 \pm 0.041$ \\ 
Astronomy & $0.243 \pm 0.035$ & $0.329 \pm 0.038$ & $0.237 \pm 0.035$ & $0.211 \pm 0.033$ & $0.237 \pm 0.035$ & $0.382 \pm 0.040$ \\ 
Business Ethics & $0.290 \pm 0.046$ & $0.280 \pm 0.045$ & $0.360 \pm 0.048$ & $0.330 \pm 0.047$ & $0.300 \pm 0.046$ & $0.390 \pm 0.049$ \\ 
Clinical Knowledge & $0.272 \pm 0.027$ & $0.291 \pm 0.028$ & $0.223 \pm 0.026$ & $0.234 \pm 0.026$ & $0.253 \pm 0.027$ & $0.317 \pm 0.029$ \\ 
College Biology & $0.285 \pm 0.038$ & $0.271 \pm 0.037$ & $0.271 \pm 0.037$ & $0.299 \pm 0.038$ & $0.208 \pm 0.034$ & $0.347 \pm 0.040$ \\ 
College Chemistry & $0.240 \pm 0.043$ & $0.160 \pm 0.037$ & $0.270 \pm 0.045$ & $0.290 \pm 0.046$ & $0.210 \pm 0.041$ & $0.250 \pm 0.044$ \\ 
College Computer Science & $0.270 \pm 0.045$ & $0.250 \pm 0.044$ & $0.310 \pm 0.046$ & $0.270 \pm 0.045$ & $0.240 \pm 0.043$ & $0.260 \pm 0.044$ \\ 
College Mathematics & $0.260 \pm 0.044$ & $0.240 \pm 0.043$ & $0.220 \pm 0.042$ & $0.160 \pm 0.037$ & $0.200 \pm 0.040$ & $0.170 \pm 0.038$ \\ 
College Medicine & $0.197 \pm 0.030$ & $0.283 \pm 0.034$ & $0.237 \pm 0.032$ & $0.202 \pm 0.031$ & $0.225 \pm 0.032$ & $0.289 \pm 0.035$ \\ 
College Physics & $0.206 \pm 0.040$ & $0.284 \pm 0.045$ & $0.304 \pm 0.046$ & $0.324 \pm 0.047$ & $0.255 \pm 0.043$ & $0.235 \pm 0.042$ \\ 
Computer Security & $0.270 \pm 0.045$ & $0.290 \pm 0.046$ & $0.250 \pm 0.044$ & $0.240 \pm 0.043$ & $0.320 \pm 0.047$ & $0.350 \pm 0.048$ \\ 
Conceptual Physics & $0.255 \pm 0.029$ & $0.294 \pm 0.030$ & $0.264 \pm 0.029$ & $0.260 \pm 0.029$ & $0.268 \pm 0.029$ & $0.294 \pm 0.030$ \\ 
Econometrics & $0.237 \pm 0.040$ & $0.289 \pm 0.043$ & $0.289 \pm 0.043$ & $0.246 \pm 0.040$ & $0.246 \pm 0.040$ & $0.228 \pm 0.039$ \\ 
Electrical Engineering & $0.359 \pm 0.040$ & $0.303 \pm 0.038$ & $0.338 \pm 0.039$ & $0.276 \pm 0.037$ & $0.310 \pm 0.039$ & $0.414 \pm 0.041$ \\ 
Elementary Mathematics & $0.254 \pm 0.022$ & $0.283 \pm 0.023$ & $0.243 \pm 0.022$ & $0.272 \pm 0.023$ & $0.249 \pm 0.022$ & $0.312 \pm 0.024$ \\ 
Formal Logic & $0.341 \pm 0.042$ & $0.294 \pm 0.041$ & $0.262 \pm 0.039$ & $0.349 \pm 0.043$ & $0.270 \pm 0.040$ & $0.294 \pm 0.041$ \\ 
Global Facts & $0.250 \pm 0.044$ & $0.220 \pm 0.042$ & $0.240 \pm 0.043$ & $0.240 \pm 0.043$ & $0.300 \pm 0.046$ & $0.290 \pm 0.046$ \\ 
High School Biology & $0.252 \pm 0.025$ & $0.300 \pm 0.026$ & $0.235 \pm 0.024$ & $0.232 \pm 0.024$ & $0.271 \pm 0.025$ & $0.335 \pm 0.027$ \\ 
High School Chemistry & $0.202 \pm 0.028$ & $0.236 \pm 0.030$ & $0.246 \pm 0.030$ & $0.241 \pm 0.030$ & $0.197 \pm 0.028$ & $0.232 \pm 0.030$ \\ 
High School Computer Science & $0.250 \pm 0.044$ & $0.210 \pm 0.041$ & $0.190 \pm 0.039$ & $0.240 \pm 0.043$ & $0.220 \pm 0.042$ & $0.290 \pm 0.046$ \\ 
High School European History & $0.261 \pm 0.034$ & $0.255 \pm 0.034$ & $0.224 \pm 0.033$ & $0.285 \pm 0.035$ & $0.261 \pm 0.034$ & $0.303 \pm 0.036$ \\ 
High School Geography & $0.202 \pm 0.029$ & $0.227 \pm 0.030$ & $0.217 \pm 0.029$ & $0.207 \pm 0.029$ & $0.242 \pm 0.031$ & $0.348 \pm 0.034$ \\ 
High School Government and Politics & $0.228 \pm 0.030$ & $0.228 \pm 0.030$ & $0.212 \pm 0.030$ & $0.181 \pm 0.028$ & $0.212 \pm 0.030$ & $0.326 \pm 0.034$ \\ 
High School Macroeconomics & $0.285 \pm 0.023$ & $0.328 \pm 0.024$ & $0.272 \pm 0.023$ & $0.277 \pm 0.023$ & $0.277 \pm 0.023$ & $0.303 \pm 0.023$ \\
High School Mathematics & $0.219 \pm 0.025$ & $0.263 \pm 0.027$ & $0.196 \pm 0.024$ & $0.230 \pm 0.026$ & $0.167 \pm 0.023$ & $0.248 \pm 0.026$ \\ 
\bottomrule\end{tabular}
\caption{Zero-Shot Results on Hendrycks Tasks, Part 1 (GPT-J, GPT-NeoX and GPT-3)}
\label{tab:hendrycks-gpt1}
\end{table*}

\begin{table*}    
\centering 
\begin{tabular}{l c c c c c c }
 & GPT-J & GPT-NeoX & \multicolumn{4}{c}{GPT-3} \\
Task & 6B & 20B & Ada & Babbage & Curie & DaVinci\\ \toprule
High School Microeconomics & $0.277 \pm 0.029$ & $0.294 \pm 0.030$ & $0.235 \pm 0.028$ & $0.265 \pm 0.029$ & $0.239 \pm 0.028$ & $0.307 \pm 0.030$ \\ 
High School Physics & $0.272 \pm 0.036$ & $0.298 \pm 0.037$ & $0.199 \pm 0.033$ & $0.298 \pm 0.037$ & $0.199 \pm 0.033$ & $0.219 \pm 0.034$ \\ 
High School Physiology & $0.273 \pm 0.019$ & $0.283 \pm 0.019$ & $0.209 \pm 0.017$ & $0.217 \pm 0.018$ & $0.246 \pm 0.018$ & $0.352 \pm 0.020$ \\ 
High School Statistics & $0.292 \pm 0.031$ & $0.319 \pm 0.032$ & $0.241 \pm 0.029$ & $0.278 \pm 0.031$ & $0.255 \pm 0.030$ & $0.278 \pm 0.031$ \\ 
High School US History & $0.289 \pm 0.032$ & $0.309 \pm 0.032$ & $0.255 \pm 0.031$ & $0.260 \pm 0.031$ & $0.240 \pm 0.030$ & $0.368 \pm 0.034$ \\ 
High School World History & $0.283 \pm 0.029$ & $0.295 \pm 0.030$ & $0.278 \pm 0.029$ & $0.262 \pm 0.029$ & $0.270 \pm 0.029$ & $0.321 \pm 0.030$ \\ 
Human Aging & $0.265 \pm 0.030$ & $0.224 \pm 0.028$ & $0.368 \pm 0.032$ & $0.336 \pm 0.032$ & $0.296 \pm 0.031$ & $0.327 \pm 0.031$ \\ 
Human Sexuality & $0.397 \pm 0.043$ & $0.405 \pm 0.043$ & $0.374 \pm 0.042$ & $0.427 \pm 0.043$ & $0.397 \pm 0.043$ & $0.481 \pm 0.044$ \\ 
International Law & $0.264 \pm 0.040$ & $0.298 \pm 0.042$ & $0.182 \pm 0.035$ & $0.207 \pm 0.037$ & $0.207 \pm 0.037$ & $0.331 \pm 0.043$ \\ 
Jurisprudence & $0.278 \pm 0.043$ & $0.250 \pm 0.042$ & $0.287 \pm 0.044$ & $0.278 \pm 0.043$ & $0.259 \pm 0.042$ & $0.370 \pm 0.047$ \\ 
Logical Fallacies & $0.294 \pm 0.036$ & $0.227 \pm 0.033$ & $0.239 \pm 0.034$ & $0.221 \pm 0.033$ & $0.245 \pm 0.034$ & $0.252 \pm 0.034$ \\ 
Machine Learning & $0.223 \pm 0.040$ & $0.268 \pm 0.042$ & $0.241 \pm 0.041$ & $0.286 \pm 0.043$ & $0.295 \pm 0.043$ & $0.232 \pm 0.040$ \\ 
Management & $0.233 \pm 0.042$ & $0.282 \pm 0.045$ & $0.184 \pm 0.038$ & $0.214 \pm 0.041$ & $0.320 \pm 0.046$ & $0.456 \pm 0.049$ \\ 
Marketing & $0.303 \pm 0.030$ & $0.321 \pm 0.031$ & $0.308 \pm 0.030$ & $0.282 \pm 0.029$ & $0.308 \pm 0.030$ & $0.491 \pm 0.033$ \\ 
Medical Genetics & $0.310 \pm 0.046$ & $0.340 \pm 0.048$ & $0.260 \pm 0.044$ & $0.300 \pm 0.046$ & $0.330 \pm 0.047$ & $0.430 \pm 0.050$ \\ 
Miscellaneous & $0.275 \pm 0.016$ & $0.299 \pm 0.016$ & $0.257 \pm 0.016$ & $0.269 \pm 0.016$ & $0.284 \pm 0.016$ & $0.450 \pm 0.018$ \\ 
Moral Disputes & $0.283 \pm 0.024$ & $0.289 \pm 0.024$ & $0.263 \pm 0.024$ & $0.263 \pm 0.024$ & $0.277 \pm 0.024$ & $0.301 \pm 0.025$ \\ 
Moral Scenarios & $0.237 \pm 0.014$ & $0.232 \pm 0.014$ & $0.238 \pm 0.014$ & $0.273 \pm 0.015$ & $0.238 \pm 0.014$ & $0.249 \pm 0.014$ \\ 
Nutrition & $0.346 \pm 0.027$ & $0.379 \pm 0.028$ & $0.301 \pm 0.026$ & $0.281 \pm 0.026$ & $0.291 \pm 0.026$ & $0.353 \pm 0.027$ \\ 
Philosophy & $0.260 \pm 0.025$ & $0.293 \pm 0.026$ & $0.215 \pm 0.023$ & $0.267 \pm 0.025$ & $0.244 \pm 0.024$ & $0.367 \pm 0.027$ \\ 
Prehistory & $0.244 \pm 0.024$ & $0.272 \pm 0.025$ & $0.244 \pm 0.024$ & $0.269 \pm 0.025$ & $0.284 \pm 0.025$ & $0.324 \pm 0.026$ \\ 
Professional Accounting & $0.262 \pm 0.026$ & $0.234 \pm 0.025$ & $0.202 \pm 0.024$ & $0.255 \pm 0.026$ & $0.238 \pm 0.025$ & $0.287 \pm 0.027$ \\ 
Professional Law & $0.241 \pm 0.011$ & $0.267 \pm 0.011$ & $0.261 \pm 0.011$ & $0.256 \pm 0.011$ & $0.259 \pm 0.011$ & $0.261 \pm 0.011$ \\ 
Professional Medicine & $0.276 \pm 0.027$ & $0.287 \pm 0.027$ & $0.221 \pm 0.025$ & $0.239 \pm 0.026$ & $0.265 \pm 0.027$ & $0.324 \pm 0.028$ \\ 
Professional Psychology & $0.284 \pm 0.018$ & $0.275 \pm 0.018$ & $0.245 \pm 0.017$ & $0.225 \pm 0.017$ & $0.257 \pm 0.018$ & $0.335 \pm 0.019$ \\ 
Public Relations & $0.282 \pm 0.043$ & $0.345 \pm 0.046$ & $0.255 \pm 0.042$ & $0.327 \pm 0.045$ & $0.364 \pm 0.046$ & $0.364 \pm 0.046$ \\ 
Security Studies & $0.363 \pm 0.031$ & $0.376 \pm 0.031$ & $0.367 \pm 0.031$ & $0.347 \pm 0.030$ & $0.384 \pm 0.031$ & $0.392 \pm 0.031$ \\ 
Sociology & $0.279 \pm 0.032$ & $0.284 \pm 0.032$ & $0.328 \pm 0.033$ & $0.303 \pm 0.033$ & $0.274 \pm 0.032$ & $0.368 \pm 0.034$ \\ 
US Foreign Policy & $0.340 \pm 0.048$ & $0.360 \pm 0.048$ & $0.330 \pm 0.047$ & $0.330 \pm 0.047$ & $0.380 \pm 0.049$ & $0.500 \pm 0.050$ \\ 
Virology & $0.355 \pm 0.037$ & $0.361 \pm 0.037$ & $0.307 \pm 0.036$ & $0.319 \pm 0.036$ & $0.337 \pm 0.037$ & $0.386 \pm 0.038$ \\ 
World Religions & $0.333 \pm 0.036$ & $0.386 \pm 0.037$ & $0.316 \pm 0.036$ & $0.310 \pm 0.035$ & $0.374 \pm 0.037$ & $0.398 \pm 0.038$ \\  \bottomrule
\end{tabular}
\caption{Zero-Shot Results on Hendrycks Tasks, Part 2 (GPT-J, GPT-NeoX, and GPT-3)}
\label{tab:hendrycks-gpt2}
\end{table*}

\begin{table*}    
\centering 
\begin{tabular}{l c c c c c c }
 & \multicolumn{6}{c}{FairSeq} \\
Task & 125M & 355M & 1.3B & 2.7B & 6.7B & 13B\\ \toprule
Abstract Algebra & $0.260 \pm 0.044$ & $0.180 \pm 0.039$ & $0.230 \pm 0.042$ & $0.250 \pm 0.044$ & $0.240 \pm 0.043$ & $0.260 \pm 0.044$ \\ 
Anatomy & $0.178 \pm 0.033$ & $0.207 \pm 0.035$ & $0.185 \pm 0.034$ & $0.170 \pm 0.032$ & $0.259 \pm 0.038$ & $0.237 \pm 0.037$ \\ 
Astronomy & $0.270 \pm 0.036$ & $0.237 \pm 0.035$ & $0.243 \pm 0.035$ & $0.263 \pm 0.036$ & $0.296 \pm 0.037$ & $0.257 \pm 0.036$ \\ 
Business Ethics & $0.330 \pm 0.047$ & $0.410 \pm 0.049$ & $0.340 \pm 0.048$ & $0.350 \pm 0.048$ & $0.380 \pm 0.049$ & $0.340 \pm 0.048$ \\ 
Clinical Knowledge & $0.215 \pm 0.025$ & $0.264 \pm 0.027$ & $0.226 \pm 0.026$ & $0.249 \pm 0.027$ & $0.223 \pm 0.026$ & $0.264 \pm 0.027$ \\ 
College Biology & $0.285 \pm 0.038$ & $0.201 \pm 0.034$ & $0.243 \pm 0.036$ & $0.222 \pm 0.035$ & $0.271 \pm 0.037$ & $0.306 \pm 0.039$ \\ 
College Chemistry & $0.310 \pm 0.046$ & $0.290 \pm 0.046$ & $0.350 \pm 0.048$ & $0.300 \pm 0.046$ & $0.280 \pm 0.045$ & $0.240 \pm 0.043$ \\ 
College Computer Science & $0.200 \pm 0.040$ & $0.250 \pm 0.044$ & $0.260 \pm 0.044$ & $0.250 \pm 0.044$ & $0.300 \pm 0.046$ & $0.280 \pm 0.045$ \\ 
College Mathematics & $0.190 \pm 0.039$ & $0.170 \pm 0.038$ & $0.230 \pm 0.042$ & $0.200 \pm 0.040$ & $0.230 \pm 0.042$ & $0.250 \pm 0.044$ \\ 
College Medicine & $0.243 \pm 0.033$ & $0.237 \pm 0.032$ & $0.249 \pm 0.033$ & $0.254 \pm 0.033$ & $0.237 \pm 0.032$ & $0.260 \pm 0.033$ \\ 
College Physics & $0.216 \pm 0.041$ & $0.245 \pm 0.043$ & $0.216 \pm 0.041$ & $0.275 \pm 0.044$ & $0.343 \pm 0.047$ & $0.216 \pm 0.041$ \\ 
Computer Security & $0.240 \pm 0.043$ & $0.290 \pm 0.046$ & $0.300 \pm 0.046$ & $0.240 \pm 0.043$ & $0.230 \pm 0.042$ & $0.320 \pm 0.047$ \\ 
Conceptual Physics & $0.260 \pm 0.029$ & $0.255 \pm 0.029$ & $0.247 \pm 0.028$ & $0.243 \pm 0.028$ & $0.247 \pm 0.028$ & $0.204 \pm 0.026$ \\ 
Econometrics & $0.246 \pm 0.040$ & $0.272 \pm 0.042$ & $0.246 \pm 0.040$ & $0.281 \pm 0.042$ & $0.219 \pm 0.039$ & $0.263 \pm 0.041$ \\ 
Electrical Engineering & $0.283 \pm 0.038$ & $0.303 \pm 0.038$ & $0.234 \pm 0.035$ & $0.276 \pm 0.037$ & $0.310 \pm 0.039$ & $0.290 \pm 0.038$ \\ 
Elementary Mathematics & $0.246 \pm 0.022$ & $0.214 \pm 0.021$ & $0.233 \pm 0.022$ & $0.233 \pm 0.022$ & $0.246 \pm 0.022$ & $0.198 \pm 0.021$ \\ 
Formal Logic & $0.278 \pm 0.040$ & $0.302 \pm 0.041$ & $0.278 \pm 0.040$ & $0.310 \pm 0.041$ & $0.286 \pm 0.040$ & $0.333 \pm 0.042$ \\ 
Global Facts & $0.200 \pm 0.040$ & $0.210 \pm 0.041$ & $0.190 \pm 0.039$ & $0.150 \pm 0.036$ & $0.220 \pm 0.042$ & $0.160 \pm 0.037$ \\ 
High School Biology & $0.248 \pm 0.025$ & $0.255 \pm 0.025$ & $0.268 \pm 0.025$ & $0.226 \pm 0.024$ & $0.274 \pm 0.025$ & $0.235 \pm 0.024$ \\ 
High School Chemistry & $0.217 \pm 0.029$ & $0.207 \pm 0.029$ & $0.256 \pm 0.031$ & $0.281 \pm 0.032$ & $0.217 \pm 0.029$ & $0.266 \pm 0.031$ \\ 
High School Computer Science & $0.240 \pm 0.043$ & $0.230 \pm 0.042$ & $0.270 \pm 0.045$ & $0.240 \pm 0.043$ & $0.350 \pm 0.048$ & $0.280 \pm 0.045$ \\ 
High School European History & $0.230 \pm 0.033$ & $0.333 \pm 0.037$ & $0.279 \pm 0.035$ & $0.261 \pm 0.034$ & $0.273 \pm 0.035$ & $0.230 \pm 0.033$ \\ 
High School Geography & $0.263 \pm 0.031$ & $0.273 \pm 0.032$ & $0.222 \pm 0.030$ & $0.258 \pm 0.031$ & $0.207 \pm 0.029$ & $0.253 \pm 0.031$ \\ 
High School Government and Politics & $0.254 \pm 0.031$ & $0.290 \pm 0.033$ & $0.228 \pm 0.030$ & $0.233 \pm 0.031$ & $0.218 \pm 0.030$ & $0.187 \pm 0.028$ \\ 
High School Macroeconomics & $0.200 \pm 0.020$ & $0.272 \pm 0.023$ & $0.254 \pm 0.022$ & $0.269 \pm 0.022$ & $0.326 \pm 0.024$ & $0.256 \pm 0.022$ \\ 
High School Mathematics & $0.204 \pm 0.025$ & $0.189 \pm 0.024$ & $0.170 \pm 0.023$ & $0.226 \pm 0.025$ & $0.200 \pm 0.024$ & $0.193 \pm 0.024$ \\ 
\bottomrule
\end{tabular}
\caption{Zero-Shot Results on Hendrycks Tasks, Part 1 (FairSeq Models)}
\label{tab:hendrycks-fair1}
\end{table*}

\begin{table*}    
\centering 
\begin{tabular}{l c c c c c c }
 & \multicolumn{6}{c}{FairSeq} \\
Task & 125M & 355M & 1.3B & 2.7B & 6.7B & 13B\\ \toprule
High School Microeconomics & $0.248 \pm 0.028$ & $0.256 \pm 0.028$ & $0.244 \pm 0.028$ & $0.248 \pm 0.028$ & $0.269 \pm 0.029$ & $0.227 \pm 0.027$ \\ 
High School Physics & $0.238 \pm 0.035$ & $0.219 \pm 0.034$ & $0.258 \pm 0.036$ & $0.245 \pm 0.035$ & $0.232 \pm 0.034$ & $0.166 \pm 0.030$ \\ 
High School Physiology & $0.235 \pm 0.018$ & $0.272 \pm 0.019$ & $0.266 \pm 0.019$ & $0.284 \pm 0.019$ & $0.250 \pm 0.019$ & $0.261 \pm 0.019$ \\ 
High School Statistics & $0.222 \pm 0.028$ & $0.241 \pm 0.029$ & $0.269 \pm 0.030$ & $0.250 \pm 0.030$ & $0.287 \pm 0.031$ & $0.241 \pm 0.029$ \\ 
High School US History & $0.240 \pm 0.030$ & $0.284 \pm 0.032$ & $0.299 \pm 0.032$ & $0.299 \pm 0.032$ & $0.314 \pm 0.033$ & $0.294 \pm 0.032$ \\ 
High School World History & $0.283 \pm 0.029$ & $0.232 \pm 0.027$ & $0.270 \pm 0.029$ & $0.245 \pm 0.028$ & $0.300 \pm 0.030$ & $0.316 \pm 0.030$ \\ 
Human Aging & $0.274 \pm 0.030$ & $0.309 \pm 0.031$ & $0.323 \pm 0.031$ & $0.291 \pm 0.031$ & $0.296 \pm 0.031$ & $0.274 \pm 0.030$ \\ 
Human Sexuality & $0.252 \pm 0.038$ & $0.366 \pm 0.042$ & $0.328 \pm 0.041$ & $0.359 \pm 0.042$ & $0.359 \pm 0.042$ & $0.351 \pm 0.042$ \\
International Law & $0.157 \pm 0.033$ & $0.223 \pm 0.038$ & $0.240 \pm 0.039$ & $0.281 \pm 0.041$ & $0.264 \pm 0.040$ & $0.231 \pm 0.038$ \\
Jurisprudence & $0.241 \pm 0.041$ & $0.269 \pm 0.043$ & $0.287 \pm 0.044$ & $0.241 \pm 0.041$ & $0.213 \pm 0.040$ & $0.278 \pm 0.043$ \\
Logical Fallacies & $0.196 \pm 0.031$ & $0.221 \pm 0.033$ & $0.233 \pm 0.033$ & $0.196 \pm 0.031$ & $0.245 \pm 0.034$ & $0.221 \pm 0.033$ \\ 
Machine Learning & $0.232 \pm 0.040$ & $0.295 \pm 0.043$ & $0.348 \pm 0.045$ & $0.232 \pm 0.040$ & $0.259 \pm 0.042$ & $0.241 \pm 0.041$ \\ 
Management & $0.223 \pm 0.041$ & $0.311 \pm 0.046$ & $0.214 \pm 0.041$ & $0.291 \pm 0.045$ & $0.340 \pm 0.047$ & $0.262 \pm 0.044$ \\ 
Marketing & $0.295 \pm 0.030$ & $0.231 \pm 0.028$ & $0.286 \pm 0.030$ & $0.303 \pm 0.030$ & $0.333 \pm 0.031$ & $0.329 \pm 0.031$ \\ 
Medical Genetics & $0.250 \pm 0.044$ & $0.310 \pm 0.046$ & $0.310 \pm 0.046$ & $0.280 \pm 0.045$ & $0.270 \pm 0.045$ & $0.300 \pm 0.046$ \\ 
Miscellaneous & $0.258 \pm 0.016$ & $0.301 \pm 0.016$ & $0.264 \pm 0.016$ & $0.249 \pm 0.015$ & $0.284 \pm 0.016$ & $0.268 \pm 0.016$ \\ 
Moral Disputes & $0.269 \pm 0.024$ & $0.246 \pm 0.023$ & $0.220 \pm 0.022$ & $0.260 \pm 0.024$ & $0.269 \pm 0.024$ & $0.272 \pm 0.024$ \\ 
Moral Scenarios & $0.255 \pm 0.015$ & $0.236 \pm 0.014$ & $0.273 \pm 0.015$ & $0.238 \pm 0.014$ & $0.241 \pm 0.014$ & $0.253 \pm 0.015$ \\ 
Nutrition & $0.252 \pm 0.025$ & $0.261 \pm 0.025$ & $0.297 \pm 0.026$ & $0.297 \pm 0.026$ & $0.330 \pm 0.027$ & $0.304 \pm 0.026$ \\ 
Philosophy & $0.199 \pm 0.023$ & $0.219 \pm 0.023$ & $0.228 \pm 0.024$ & $0.222 \pm 0.024$ & $0.238 \pm 0.024$ & $0.270 \pm 0.025$ \\ 
Prehistory & $0.290 \pm 0.025$ & $0.222 \pm 0.023$ & $0.253 \pm 0.024$ & $0.228 \pm 0.023$ & $0.296 \pm 0.025$ & $0.235 \pm 0.024$ \\ 
Professional Accounting & $0.262 \pm 0.026$ & $0.220 \pm 0.025$ & $0.209 \pm 0.024$ & $0.170 \pm 0.022$ & $0.238 \pm 0.025$ & $0.266 \pm 0.026$ \\ 
Professional Law & $0.261 \pm 0.011$ & $0.261 \pm 0.011$ & $0.256 \pm 0.011$ & $0.256 \pm 0.011$ & $0.259 \pm 0.011$ & $0.261 \pm 0.011$ \\ 
Professional Medicine & $0.239 \pm 0.026$ & $0.254 \pm 0.026$ & $0.254 \pm 0.026$ & $0.206 \pm 0.025$ & $0.221 \pm 0.025$ & $0.195 \pm 0.024$ \\ 
Professional Psychology & $0.245 \pm 0.017$ & $0.247 \pm 0.017$ & $0.242 \pm 0.017$ & $0.248 \pm 0.017$ & $0.278 \pm 0.018$ & $0.252 \pm 0.018$ \\ 
Public Relations & $0.236 \pm 0.041$ & $0.245 \pm 0.041$ & $0.264 \pm 0.042$ & $0.227 \pm 0.040$ & $0.291 \pm 0.044$ & $0.291 \pm 0.044$ \\ 
Security Studies & $0.322 \pm 0.030$ & $0.331 \pm 0.030$ & $0.331 \pm 0.030$ & $0.335 \pm 0.030$ & $0.408 \pm 0.031$ & $0.359 \pm 0.031$ \\ 
Sociology & $0.234 \pm 0.030$ & $0.234 \pm 0.030$ & $0.259 \pm 0.031$ & $0.229 \pm 0.030$ & $0.234 \pm 0.030$ & $0.323 \pm 0.033$ \\ 
US Foreign Policy & $0.250 \pm 0.044$ & $0.300 \pm 0.046$ & $0.300 \pm 0.046$ & $0.310 \pm 0.046$ & $0.370 \pm 0.049$ & $0.330 \pm 0.047$ \\ 
Virology & $0.289 \pm 0.035$ & $0.301 \pm 0.036$ & $0.319 \pm 0.036$ & $0.355 \pm 0.037$ & $0.295 \pm 0.036$ & $0.331 \pm 0.037$ \\ 
World Religions & $0.292 \pm 0.035$ & $0.263 \pm 0.034$ & $0.287 \pm 0.035$ & $0.292 \pm 0.035$ & $0.269 \pm 0.034$ & $0.339 \pm 0.036$ \\ 
\bottomrule
\end{tabular}
\caption{Zero-shot Results on Hendrycks Tasks, Part 2 (FairSeq Models)}
\label{tab:hendrycks-fair2}
\end{table*}
\end{landscape}}



\clearpage

\section{Tokenizer Analysis}\label{sec:tokenizer-analysis}

Both tokenizers share 36938 out of 50257 tokens, a $\sim$73.5\% overlap in tokens.
In this section, we perform comparison between the \model{} tokenizer to the GPT-2 tokenizer using the validation set of the Pile.

In \Cref{tab:token_pileval-all}, we show the resulting number of tokens from tokenizing each component of the Pile’s validation set with both tokenizers, and the ratio of \model{} tokens to GPT-2 tokens.

We observe that the \model{} tokenizer represents all Pile components using fewer or very closely comparable numbers of tokens. The largest percentage improvement in token counts are in the EuroParl, GitHub, and PubMed Central components, with a more than 20\% savings in the number of tokens needed to represent that component. We highlight that arXiv, GitHub, and StackExchange---subsets with large code components---can be represented with meaningfully fewer tokens with the \model{} tokenizer compared to the GPT-2 tokenizer. Overall, the \model{} tokenizer represents the Pile validation set with approximately 10\% fewer tokens compared to the GPT-2 tokenizer.

Given that the \model{} tokenizer is tweaked to better tokenize whitespace, we also perform a comparison between the two tokenizers excluding whitespace. We perform the same analysis as the above, but exclude all whitespace tokens from our computations, only counting the non-whitespace tokens. A token is considered a whitespace token if it consists only of whitespace characters. The results are shown in \Cref{tab:token_pileval-nowhitespace} in the Appendix. We observe that the \model{} tokenizer still uses 5\% fewer tokens to represent the Pile validation set compared to the GPT-2 tokenizer. As expected, the token ratios for certain components such as GitHub and StackExchange become closer to even once the whitespace characters are excluded. 

\begin{table}[ht!]
\centering
\resizebox{0.48\textwidth}{!}{
\begin{tabular}{lrrcc}
\toprule
    & GPT-2 & \model{} & $\frac{\text{\model}}{\text{GPT-2}}$ \\ 
    \midrule
    Pile (val) & 383,111,734 & 342,887,807 & 0.89501 \\
    C4 & 173,669,294 & 173,768,876  & 1.001 \\
    C4 excl. Space & 168,932,391 & 171,003,008 & 1.012 \\
    \bottomrule
\end{tabular}
}
\caption{
    Number of tokens from tokenizing the AllenAI C4 (\texttt{en}) validation set.
    The \model{} tokenizer uses approximately the same number of tokens to represent C4 as the GPT-2 tokenizer.
}
\label{tab:token_c4}
\end{table}

While we evaluated our tokenizer using the validation set for the Pile, the Pile components would still be considered in-domain for the tokenizer and may not provide the most informative comparison point. To perform an out-of-domain comparison, we perform the same analysis using the AllenAI replication of C4,\footnote{\url{https://github.com/allenai/allennlp/discussions/5056}}, another popular pretraining corpus for large language models. As above, we use the validation set for our analysis. Our results are shown in \Cref{tab:token_c4}. We find that the \model{} tokenizer tokenizes the C4 validation set to approximately the same number of tokens as the GPT-2 tokenizer. When excluding all whitespace tokens, the \model{} requires approximately 1\% more tokens to represent the corpus compared to the GPT-2 tokenizer.

\begin{table}[ht!]
\centering
\resizebox{0.48\textwidth}{!}{
\begin{tabular}{lrrcc}
\toprule
    & GPT-2 & \model & $\frac{\text{\model}}{\text{GPT-2}}$ \\ 
    \midrule
    arXiv & 41,020,155 & 34,704,315 & 0.84603 \\
    BookCorpus2 & 2,336,388 & 2,365,633 & 1.01252 \\
    Books3 & 42,819,036 & 43,076,832 & 1.00602 \\
    DM Mathematics & 7,699,527 & 7,413,775 & 0.96289 \\
    Enron Emails & 480,500 & 433,867 & 0.90295 \\
    EuroParl & 3,519,584 & 2,808,275 & 0.79790 \\
    FreeLaw & 21,098,168 & 18,687,364 & 0.88573 \\
    GitHub & 42,986,216 & 33,021,839 & 0.76820 \\
    Gutenberg (PG-19) & 6,729,187 & 6,428,946 & 0.95538 \\
    HackerNews & 2,578,933 & 2,551,720 & 0.98945 \\
    NIH ExPorter & 776,688 & 739,558 & 0.95219 \\
    OpenSubtitles & 5,431,529 & 5,446,485 & 1.00275 \\
    OpenWebText2 & 31,993,480 & 30,813,744 & 0.96313 \\
    PhilPapers & 1,879,206 & 1,750,928 & 0.93174 \\
    Pile-CC & 53,415,704 & 53,392,389 & 0.99956 \\
    PubMed Abstracts & 8,708,180 & 8,215,529 & 0.94343 \\
    PubMed Central & 56,874,247 & 43,534,166 & 0.76545 \\
    StackExchange & 22,708,643 & 19,000,198 & 0.83669 \\
    USPTO Backgrounds & 10,217,886 & 9,727,223 & 0.95198 \\
    Ubuntu IRC & 3,341,287 & 2,771,066 & 0.82934 \\
    Wikipedia (en) & 12,614,087 & 12,692,048 & 1.00618 \\
    YoutubeSubtitles & 3,883,103 & 3,311,907 & 0.85290 \\
    \midrule
    Total & 383,111,734 & 342,887,807 & 0.89501 \\
    \bottomrule
\end{tabular}
}
\caption{
    Number of tokens from tokenizing the Pile validation set. 
    The \model{} tokenizer uses fewer tokens to represent the Pile overall, with the biggest gains in whitespace heavy datasets such as arXiv, GitHub and StackExchange.
}
\label{tab:token_pileval-all}

\end{table}

\subsection{Tokenizer Comparisons}

\begin{table}[ht!]
\centering
\resizebox{0.48\textwidth}{!}{
\begin{tabular}{lrrcc}
\toprule
    & GPT-2 & \model & $\frac{\text{\model}}{\text{GPT-2}}$ \\ 
    \midrule
    arXiv & 38,932,524 & 33,561,364 & 0.86204 \\
    BookCorpus2 & 2,233,367 & 2,262,609 & 1.01309 \\
    Books3 & 40,895,236 & 41,198,424 & 1.00741 \\
    DM Mathematics & 7,214,874 & 6,929,066 & 0.96039 \\
    Enron Emails & 374,978 & 373,498 & 0.99605 \\
    EuroParl & 3,482,120 & 2,780,405 & 0.79848 \\
    FreeLaw & 17,766,692 & 17,434,708 & 0.98131 \\
    GitHub & 29,338,176 & 27,558,966 & 0.93936 \\
    Gutenberg (PG-19) & 5,838,580 & 5,827,408 & 0.99809 \\
    HackerNews & 2,312,116 & 2,299,848 & 0.99469 \\
    NIH ExPorter & 776,619 & 739,543 & 0.95226 \\
    OpenSubtitles & 5,428,118 & 5,445,721 & 1.00324 \\
    OpenWebText2 & 30,849,218 & 29,723,143 & 0.96350 \\
    PhilPapers & 1,872,347 & 1,743,627 & 0.93125 \\
    Pile-CC & 51,305,080 & 51,281,909 & 0.99955 \\
    PubMed Abstracts & 8,676,790 & 8,185,417 & 0.94337 \\
    PubMed Central & 44,508,570 & 40,722,151 & 0.91493 \\
    StackExchange & 17,414,955 & 16,712,814 & 0.95968 \\
    USPTO Backgrounds & 9,882,473 & 9,601,385 & 0.97156 \\
    Ubuntu IRC & 3,220,797 & 2,659,225 & 0.82564 \\
    Wikipedia (en) & 11,874,878 & 11,986,567 & 1.00941 \\
    YoutubeSubtitles & 3,589,042 & 3,046,451 & 0.84882 \\
    \midrule
    Total & 337,787,550 & 322,074,249 & 0.95348 \\
    \bottomrule
\end{tabular}
}
\caption{
    Number of tokens from tokenizing the Pile validation set, excluding whitespace tokens. 
}
\label{tab:token_pileval-nowhitespace}
\end{table}

\subsubsection{Longest Tokens}

We show in \Cref{tab:token_longest} the 10 longest tokens in each tokenizer vocabulary. We exclude consideration of tokens that comprise only symbols or whitespace characters. We observe that for the GPT-2 tokenizer, many of the longest tokens appear to reflect artifacts in the tokenizer training data, likely with certain websites or web-scrapes being overrepresented in the training data. For the \model{} tokenizer, we observe that most of the longest tokens are scientific terms, likely arising from the PubMed components of the Pile.

\begin{table*}[ht!]
\centering
\resizebox{0.6\textwidth}{!}{
\begin{tabular}{lll}
\toprule
    & GPT-2 & \model  \\ 
    \midrule
    & \texttt{\tokentext\tokenbox{rawdownloadcloneembedreportprint}} & \texttt{\tokentext\tokenbox{~immunohistochemistry}} \\
    & \texttt{\tokentext\tokenbox{BuyableInstoreAndOnline}} & \texttt{\tokentext\tokenbox{~immunohistochemical}} \\
    & \texttt{\tokentext\tokenbox{cloneembedreportprint}} & \texttt{\tokentext\tokenbox{~telecommunications}} \\
    & \texttt{\tokentext\tokenbox{~RandomRedditorWithNo}} & \texttt{\tokentext\tokenbox{~immunofluorescence}} \\
    & \texttt{\tokentext\tokenbox{~telecommunications}} & \texttt{\tokentext\tokenbox{~immunosuppressive}} \\
    & \texttt{\tokentext\tokenbox{channelAvailability}} & \texttt{\tokentext\tokenbox{~BytePtrFromString}} \\
    & \texttt{\tokentext\tokenbox{~disproportionately}} & \texttt{\tokentext\tokenbox{~multidisciplinary}} \\
    & \texttt{\tokentext\tokenbox{~Telecommunications}} & \texttt{\tokentext\tokenbox{~histopathological}} \\
    & \texttt{\tokentext\tokenbox{~guiActiveUnfocused}} & \texttt{\tokentext\tokenbox{~neurodegenerative}} \\
    & \texttt{\tokentext\tokenbox{ItemThumbnailImage}} & \texttt{\tokentext\tokenbox{~indistinguishable}} \\
    \bottomrule
\end{tabular}
}
\caption{
    Ten longest tokens (excluding tokens comprising mainly symbols, numbers and spaces) in tokenizer vocabularies.
}
\label{tab:token_longest}
\end{table*}

\subsubsection{Worst Case Word Tokenization Comparison}

\begin{table*}[ht!]
\centering
\resizebox{\textwidth}{!}{
\begin{tabular}{lll}
\multicolumn{3}{c}{\large\textbf{GPT-2 Worst-case Tokenization}}\\
\toprule
    Word & GPT-2 Tokenization & \model{} Tokenization  \\ 
    \midrule
~hematopoietic & (6) \texttt{\tokentext\tokenbox{~he}\tokenbox{mat}\tokenbox{op}\tokenbox{o}\tokenbox{iet}\tokenbox{ic}} &  (1) \texttt{\tokentext\tokenbox{~hematopoietic}} \\
~adenocarcinoma & (6) \texttt{\tokentext\tokenbox{~ad}\tokenbox{en}\tokenbox{oc}\tokenbox{arc}\tokenbox{in}\tokenbox{oma}} &  (1) \texttt{\tokentext\tokenbox{~adenocarcinoma}} \\
~MERCHANTABILITY & (5) \texttt{\tokentext\tokenbox{~MER}\tokenbox{CH}\tokenbox{AN}\tokenbox{TA}\tokenbox{BILITY}} &  (1) \texttt{\tokentext\tokenbox{~MERCHANTABILITY}} \\
~CONSEQUENTIAL & (5) \texttt{\tokentext\tokenbox{~CON}\tokenbox{SE}\tokenbox{QU}\tokenbox{ENT}\tokenbox{IAL}} &  (1) \texttt{\tokentext\tokenbox{~CONSEQUENTIAL}} \\
~oligonucleotides & (5) \texttt{\tokentext\tokenbox{~olig}\tokenbox{on}\tokenbox{ucle}\tokenbox{ot}\tokenbox{ides}} &  (1) \texttt{\tokentext\tokenbox{~oligonucleotides}} \\
~cytoplasmic & (5) \texttt{\tokentext\tokenbox{~cy}\tokenbox{top}\tokenbox{l}\tokenbox{asm}\tokenbox{ic}} &  (1) \texttt{\tokentext\tokenbox{~cytoplasmic}} \\
~corticosteroids & (4) \texttt{\tokentext\tokenbox{~cort}\tokenbox{ic}\tokenbox{oster}\tokenbox{oids}} &  (1) \texttt{\tokentext\tokenbox{~corticosteroids}} \\
~neurodegenerative & (4) \texttt{\tokentext\tokenbox{~neuro}\tokenbox{deg}\tokenbox{ener}\tokenbox{ative}} &  (1) \texttt{\tokentext\tokenbox{~neurodegenerative}} \\
~asymptotic & (4) \texttt{\tokentext\tokenbox{~as}\tokenbox{ym}\tokenbox{pt}\tokenbox{otic}} &  (1) \texttt{\tokentext\tokenbox{~asymptotic}} \\
~aneurysm & (4) \texttt{\tokentext\tokenbox{~an}\tokenbox{eur}\tokenbox{ys}\tokenbox{m}} &  (1) \texttt{\tokentext\tokenbox{~aneurysm}} \\
    \bottomrule
\end{tabular}
\bigskip
\begin{tabular}{lll}
\multicolumn{3}{c}{\large\textbf{\model{} Worst-case Tokenization}}\\
\toprule
    Word & GPT-2 Tokenization & \model{} Tokenization  \\ 
    \midrule
    ~Schwarzenegger & (1) \texttt{\tokentext\tokenbox{~Schwarzenegger}} &  (5) \texttt{\tokentext\tokenbox{~Sch}\tokenbox{war}\tokenbox{zen}\tokenbox{eg}\tokenbox{ger}} \\
~Bolshevik & (1) \texttt{\tokentext\tokenbox{~Bolshevik}} &  (4) \texttt{\tokentext\tokenbox{~B}\tokenbox{ols}\tokenbox{he}\tokenbox{vik}} \\
~crowdfunding & (1) \texttt{\tokentext\tokenbox{~crowdfunding}} &  (4) \texttt{\tokentext\tokenbox{~crow}\tokenbox{df}\tokenbox{und}\tokenbox{ing}} \\
~misogyny & (1) \texttt{\tokentext\tokenbox{~misogyny}} &  (4) \texttt{\tokentext\tokenbox{~mis}\tokenbox{og}\tokenbox{yn}\tokenbox{y}} \\
~McAuliffe & (1) \texttt{\tokentext\tokenbox{~McAuliffe}} &  (4) \texttt{\tokentext\tokenbox{~Mc}\tokenbox{A}\tokenbox{ul}\tokenbox{iffe}} \\
~unstoppable & (1) \texttt{\tokentext\tokenbox{~unstoppable}} &  (4) \texttt{\tokentext\tokenbox{~un}\tokenbox{st}\tokenbox{opp}\tokenbox{able}} \\
~Timberwolves & (1) \texttt{\tokentext\tokenbox{~Timberwolves}} &  (4) \texttt{\tokentext\tokenbox{~Tim}\tokenbox{ber}\tokenbox{w}\tokenbox{olves}} \\
~excruciating & (1) \texttt{\tokentext\tokenbox{~excruciating}} &  (4) \texttt{\tokentext\tokenbox{~exc}\tokenbox{ru}\tokenbox{ci}\tokenbox{ating}} \\
~Kaepernick & (1) \texttt{\tokentext\tokenbox{~Kaepernick}} &  (4) \texttt{\tokentext\tokenbox{~K}\tokenbox{ae}\tokenbox{per}\tokenbox{nick}} \\
~Valkyrie & (1) \texttt{\tokentext\tokenbox{~Valkyrie}} &  (4) \texttt{\tokentext\tokenbox{~V}\tokenbox{alk}\tokenbox{y}\tokenbox{rie}} \\
    \bottomrule
\end{tabular}
}
\caption{
  Worst case word tokenization with respective tokenizers. 
  We show cases where one tokenizer requires many more tokens to represent a word compared to the other tokenizer.
}
\label{tab:token_worst}
\end{table*}

We consider the words for which there is the greatest discrepancy in the resulting token length between the two tokenizers, where one tokenizer needs many tokens to represent while the other tokenizer uses relatively few tokens. We define a word as a contiguous string delimited by whitespace or punctuation (as defined by \texttt{strings.punctuation} in Python). We perform this analysis at the component level. We only consider words that occur at least 10 times within the given component. We show in \Cref{tab:token_worst} a representative example from the Pile-CC corpus.

\section{Tokenization Examples}\label{app:tokenization}

\begin{figure*}[p]
    \centering
    \input{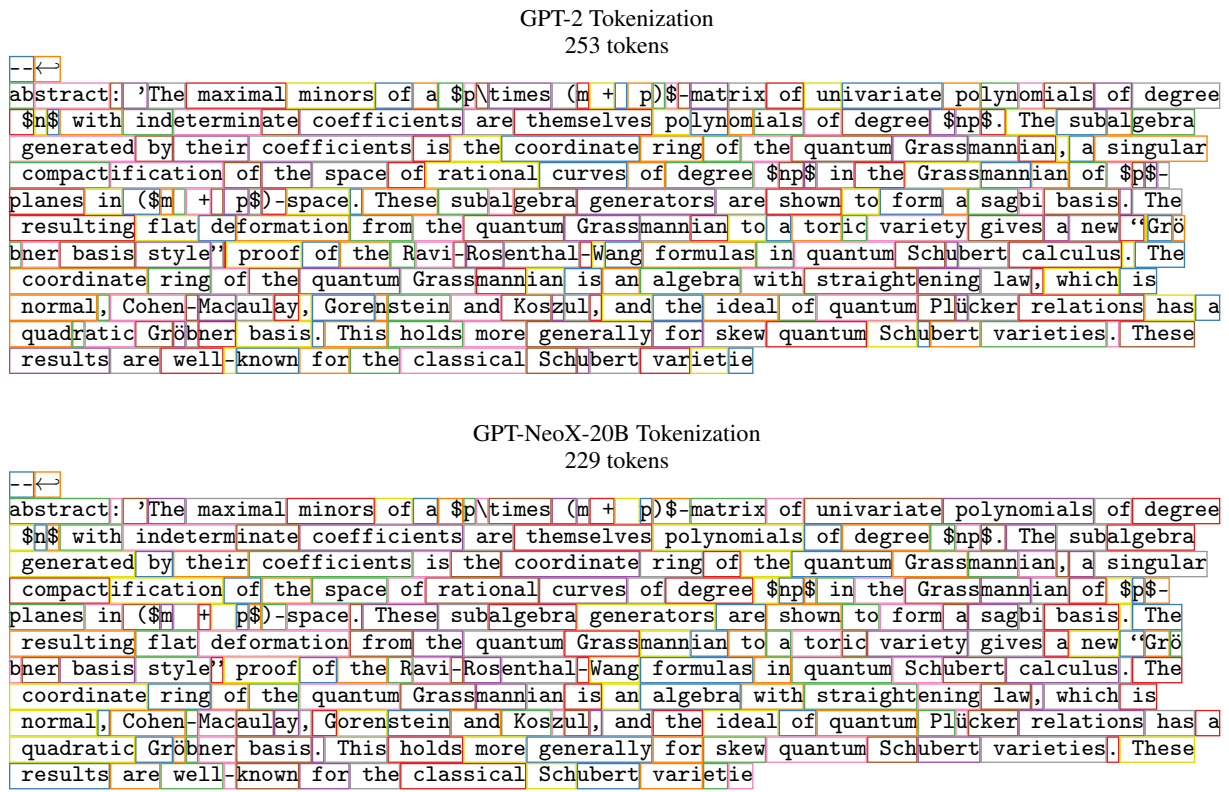}
    \caption{Pile (arXiv) Tokenization Example}
    \label{fig:tok_example_0}
\end{figure*}

In \Cref{fig:tok_example_0,fig:tok_example_15}, we show examples of tokenized documents from the Pile, comparing the GPT-2 tokenizer to ours.

\begin{figure*}[p]
    \centering
    \input{Figures/tokenization/examples_new/BookCorpus2.tex}
    \caption{Pile (BookCorpus2) Tokenization Example}
    \label{fig:tok_example_1}
\end{figure*}

\begin{figure*}[p]
    \centering
    \input{Figures/tokenization/examples_new/DM_Mathematics.tex}
    \caption{Pile (DM Mathematics) Tokenization Example}
    \label{fig:tok_example_3}
\end{figure*}

\begin{figure*}[p]
    \centering
    \input{Figures/tokenization/examples_new/Github.tex}
    \caption{Pile (GitHub) Tokenization Example}
    \label{fig:tok_example_7}
\end{figure*}

\begin{figure*}[p]
    \centering
    \input{Figures/tokenization/examples_new/OpenWebText2.tex}
    \caption{Pile (OpenWebText2) Tokenization Example}
    \label{fig:tok_example_12}
\end{figure*}

\begin{figure*}[p]
    \centering
    \input{Figures/tokenization/examples_new/PubMed_Abstracts.tex}
    \caption{Pile (PubMed Abstracts) Tokenization Example}
    \label{fig:tok_example_15}
\end{figure*}

\end{document}